\documentclass{article}
\usepackage[T1]{fontenc}    % use 8-bit T1 fonts
\usepackage{kpfonts}
\usepackage{styles/template}
\usepackage{needed_package}
\usepackage{setspace}
\newcommand{\RLxF}{RL\textit{x}F}
\definecolor{revisionv5}{RGB}{255, 0, 18}
\makeindex
\makeatletter
\title{AI Alignment: A Comprehensive Survey}

\author{
    \textbf{Jiaming Ji\textsuperscript{*,1}}~
    \textbf{Tianyi Qiu\textsuperscript{*,1}}~
    \textbf{Boyuan Chen\textsuperscript{*,1}}~
    \textbf{Borong Zhang\textsuperscript{*,1}}~
    \textbf{Hantao Lou\textsuperscript{1}}~
    \textbf{Kaile Wang\textsuperscript{1}}~
    \textbf{Yawen Duan\textsuperscript{2}}~
    \\
    \textbf{Zhonghao He\textsuperscript{2}}~
    \textbf{Lukas Vierling\textsuperscript{3}}~
    \textbf{Donghai Hong\textsuperscript{1}}~
    \textbf{Jiayi Zhou\textsuperscript{1}}~
    \textbf{Zhaowei Zhang\textsuperscript{1}}~
    \textbf{Fanzhi Zeng\textsuperscript{1}}~
    \textbf{Juntao Dai\textsuperscript{1}}~
    \\
    \textbf{Xuehai Pan\textsuperscript{1}}
    \textbf{Kwan Yee Ng}~
    \textbf{Aidan O{'}Gara\textsuperscript{6}}~
    \textbf{Hua Xu\textsuperscript{1}}~
    \textbf{Brian Tse}~
    \textbf{Jie Fu\textsuperscript{5}}~
    \textbf{Stephen McAleer\textsuperscript{3}}
    \\
    \textbf{Yaodong Yang\textsuperscript{1, \Letter}} \,
    \textbf{Yizhou Wang\textsuperscript{1}} \,
    \textbf{Song-Chun Zhu\textsuperscript{1}} \,
    \textbf{Yike Guo\textsuperscript{5}} \,
    \textbf{Wen Gao\textsuperscript{1}}
    \vspace{0.8em} \\
    \textbf{\textsuperscript{1}}\textnormal{Peking University} \, 
    \textbf{\textsuperscript{2}}\textnormal{University of Cambridge} \,
    \textbf{\textsuperscript{3}}\textnormal{University of Oxford} \,
    \textbf{\textsuperscript{4}}\textnormal{Carnegie Mellon University} \, \\
    \textbf{\textsuperscript{5}}\textnormal{Hong Kong University of Science and Technology} \,
    \textbf{\textsuperscript{6}}\textnormal{University of Southern California}
}

\begin{document}
\maketitle
\begin{abstract}
AI alignment aims to make AI systems behave in line with human intentions and values. As AI systems grow more capable, so do risks from misalignment. 
To provide a comprehensive and up-to-date overview of the alignment field, in this survey, we delve into the core concepts, methodology, and practice of alignment. 
First, we identify four principles as the key objectives of AI alignment: Robustness, Interpretability, Controllability, and Ethicality (\textbf{RICE}). Guided by these four principles, we outline the landscape of current alignment research and decompose them into two key components: \textbf{forward alignment} and \textbf{backward alignment}. The former aims to make AI systems aligned via alignment training, while the latter aims to gain evidence about the systems' alignment and govern them appropriately to avoid exacerbating misalignment risks. 
%Forward alignment and backward alignment form a recurrent process where the alignment of AI systems from the forward process is verified in the backward process, meanwhile providing updated objectives for forward alignment in the next round. 
On forward alignment, we discuss techniques for learning from feedback and learning under distribution shift. Specifically, we survey traditional preference modeling methods and reinforcement learning from human feedback, and further discuss potential frameworks to reach scalable oversight for tasks where effective human oversight is hard to obtain. Within learning under distribution shift, we also cover data distribution interventions such as adversarial training that help expand the distribution of training data, and algorithmic interventions to combat goal misgeneralization. On backward alignment, we discuss assurance techniques and governance practices. Specifically, we survey assurance methods of AI systems throughout their lifecycle, covering safety evaluation, interpretability, and human value compliance. We discuss current and prospective governance practices adopted by governments, industry actors, and other third parties, aimed at managing existing and future AI risks.

This survey aims to provide a comprehensive yet beginner-friendly review of alignment research topics. Based on this, we also release and continually update the website \url{www.alignmentsurvey.com} which features tutorials, collections of papers, blog posts, and other resources.
\end{abstract}

\nnfootnote{\; \textsuperscript{*} \; Equal contribution.}
\nnfootnote{\; \textsuperscript{\Letter}  Corresponding author. Contact \texttt{<pku.alignment@gmail.com>}.}
\nnfootnote{\; \textsuperscript{-} \,\, Version: v4 (updated on Feb 27, 2024). The content of the survey will be continually updated.}

% \newpage
% {
% \setstretch{0.99}
% \tableofcontents
% }
% \newpage

% \newpage
% \begin{multicols}{2}
% \tableofcontents
% \end{multicols}
% \newpage

\newpage
{
\setstretch{0.99}
\normalsize \tableofcontents
}
\newpage

\section{Introduction}
\label{sec:introduction}
Recent advancements have seen the increasing application of capable AI systems in complex domains. For instance, Large Language Models (LLMs) have exhibited improved capabilities in multi-step reasoning \citep{wei2022chain, wang2023self} and cross-task generalization \citep{brown2020language, askell2021general} in real-world deployment settings, and these abilities are strengthened with increased training time, training data, and parameter size \citep{kaplan2020scaling, srivastava2023beyond, hoffmann2022empirical}.
The utilization of Deep Reinforcement Learning (DRL) for the control of nuclear fusion \citep{degrave2022magnetic} is another notable example. The increasing capabilities and deployment in high-stakes domains come with heightened risks. Various undesirable behaviors exhibited by advanced AI systems (\textit{e.g.}, manipulation \citep{perez2022discovering, carroll2023characterizing, sharma2024towards} and deception \citep{park2023ai}) have raised concerns about the hazards from AI systems. 

Consequently, these concerns have catalyzed research efforts in \emph{AI alignment} \citep{soares2014aligning,christian2020alignment,hendrycks2021unsolved}. AI alignment aims to make AI systems behave in line with human intentions and values \citep{leike2018scalable}, focusing more on the objectives of AI systems than their capabilities. Failures of alignment (\emph{i.e.}, misalignment) are among the most salient causes of potential harm from AI. Mechanisms underlying these failures include \emph{reward hacking} \citep{pan2021effects} and \emph{goal misgeneralization} \citep{di2022goal}, which are further amplified by \emph{double edge components} such as situational awareness \citep{cotra2022}, broadly-scoped goals \citep{ngo2024the}, mesa-optimization objectives \citep{hubinger2019risks}, and access to increased resources \citep{shevlane2023model} (\S\ref{sec:challenges-of-alignment}).

Alignment efforts to address these failures focus on accomplishing four key objectives (\S\ref{sec:RICE}): Robustness, Interpretability, Controllability, and Ethicality (\textbf{RICE}). Current research and practice on alignment consist of four areas (\S\ref{sec:scope}): Learning from Feedback (\S\ref{sec:learning-from-feedback}), Learning under Distributional Shift (\S\ref{sec:distribution}), Assurance (\S\ref{sec:assurance}), and Governance (\S\ref{sec:governance}). The four areas and the RICE objectives are not in one-to-one correspondence. Each individual area often serves more than one alignment objective, and vice versa (see Table \ref{tab:category-over-RICE}).

In this survey, we introduce the concept, methodology, and practice of AI alignment and discuss its potential future directions.\footnote{To help beginners interested in this field learn more effectively, we highlight resources about alignment techniques. More details can be found at \url{www.alignmentsurvey.com/resources}}

\subsection{The Motivation for Alignment}
\label{sec:motivation-of-alignment}

The motivation for alignment is a three-step argument, each step building upon the previous one: (1) Deep learning-based systems (or applications) have an increasingly large impact on society and bring significant risks ; (2) Misalignment represents a significant source of risks; and (3) Alignment research and practice address risks stemming from misaligned systems (\textit{e.g.}, power-seeking behaviors).

\subsubsection{Risks of Misalignment}
\label{sec:risks-of-alignment}

With improved capabilities of AI systems, come increased risks.\footnote{We discuss and taxonomize the risks that might brought by misaligned AI systems, please see \S\ref{sec:challenges-of-alignment}.} Some undesirable behaviors of LLMs including (but not limited to) untruthful answers \citep{bang2023multitask}, sycophancy \citep{perez2022discovering, sharma2024towards}, and deception \citep{jacob2023, park2023ai} worsen with increased model scale \citep{perez2022discovering}, resulting in concerns about advanced AI systems that are hard to control. Moreover, emerging trends such as \emph{LLM-based agents} \citep{xi2023rise,wang2023survey} also raise concerns about the system's controllability and ethicality \citep{chan2023harms}. Looking further ahead, the development of increasingly competent AI systems opens up the possibility of realizing Artificial General Intelligence (AGI) in the foreseeable future, \textit{i.e.}, systems can match or surpass human intelligence in all relevant aspects \citep{bubeck2023sparks}. This could bring extensive opportunities \citep{manyika2017future}, \textit{e.g.}, automation \citep{west2018future}, efficiency improvements \citep{furman2019ai}, but also come with serious risks \citep{statement_on_ai_risk_2023,critch2023tasra}, such as safety concerns \citep{hendrycks2022x}, biases and inequalities \citep{ntoutsi2020bias}, and large-scale risks from superhuman capabilities~\citep{rogueai}. Taking biases as an example, cutting-edge LLMs manifest discernible biases about gender, sexual identity, and immigrant status among others \citep{perez2022discovering}, which could reinforce existing inequalities.

Within the large-scale risks from superhuman capabilities, it has been conjectured that global catastrophic risks (\textit{i.e.}, risks of severe harms on a global scale) \citep{bostrom2011global,hendrycks2023overview, gov_uk_2023} and existential risks (\textit{i.e.}, risks that threaten the destruction of humanity's long-term potential) from advanced AI systems are especially worrying. These concerns are elaborated in first-principle deductive arguments \citep{firstprinc,rogueai}, evolutionary analysis \citep{hendrycks2023natural}, and concrete scenario mapping \citep{christiano2019failure,threatmodel2022}. In \citet{statement_on_ai_risk_2023}, leading AI scientists and other notable figures stated that \textit{Mitigating the risk of extinction from AI should be a global priority alongside other societal-scale risks such as pandemics and nuclear war}. The median researcher surveyed by \citet{stein2022expert} at NeurIPS 2021 and ICML 2021 reported a 5\% chance that the long-run effect of advanced AI on humanity would be \textit{extremely bad (\textit{e.g.}, human extinction)}, and 36\% of NLP researchers surveyed by \citet{michael2022nlp} self-reported to believe that \textit{AI could produce catastrophic outcomes in this century, on the level of all-out nuclear war}.\footnote{However, survey results may hinge upon the exact wording of the questions and should be taken with caution.} Existential risks from AI also include risks of lock-in, stagnation, and more \citep{bostrom2013existential,hendrycks2022x}, in addition to extinction risks.\footnote{\textit{Existential} and \textit{extinction} risks are two concepts that are often mixed up. The latter is a subset of the former.} The UK have hosted the world's first global AI Safety Summit, gathering international governments, leading AI companies, civil society groups, and research experts. Its objectives are to:
(1) assess the risks associated with AI, particularly at the cutting edge of its development; (2) explore how these risks can be mitigated through internationally coordinated efforts.\footnote{Source from \url{https://www.gov.uk/government/topical-events/ai-safety-summit-2023}.} The summit culminated in the Bletchley Declaration \citep{bletch2023summit}, which highlighted the importance of international cooperation on AI safety. It was signed by representatives from 28 countries and the EU.

Current cutting-edge AI systems have exhibited multiple classes of undesirable or harmful behaviors that may contrast with human intentions (\textit{e.g.}, power-seeking and manipulation) \citep{si2022so,pan2023machiavelli}, and similar worries about more advanced systems have also been raised \citep{critch2020ai, statement_on_ai_risk_2023}.\footnote{See \S\ref{sec:challenges-of-alignment} for an introduction to specific misalignment challenges.} These undesirable or harmful behaviors not compliant with human intentions, known as \textit{misalignment} of AI systems\footnote{Some of the misaligned behaviors are less risky (\textit{e.g.}, the agent fails to clean the room as you want), however, some of them are dangerous for systems applied in the high-stakes environment (\textit{e.g.}, the control of nuclear fusion \citep{degrave2022magnetic})}, can naturally occur even without misuse by malicious actors and represent a significant source of risks from AI, including safety hazards \citep{hendrycks2021unsolved} and potential existential risks \citep{hendrycks2023overview}.\footnote{It should be noted that misalignment cannot cover all sources of risks brought by Deep learning-based systems and other factors such as misuse and negligence also contribute to risks on society. See \S\ref{sec:ai-safety-beyond} for discussing AI safety beyond alignment.} These large-scale risks are significant in size due to the non-trivial likelihoods of (1) building superintelligent AI systems, (2) those AI systems pursuing large-scale goals, (3) those goals are misaligned with human intentions and values, and (4) this misalignment leads to humans losing control of humanity's future trajectory \citep{firstprinc}.

Solving the risks brought by misalignment requires the \textit{alignment} of AI systems to ensure the objectives of the system are in accordance with human intentions and values, thereby averting unintended and unfavorable outcomes. More importantly, we expect the alignment techniques to be scaled to harder tasks and significantly advanced AI systems that are even smarter than humans. A potential solution is \textit{Superalignment}\footnote{For more details on Superalignment, you can refer to \url{https://openai.com/blog/introducing-superalignment}.}, which aims to build a roughly human-level automated alignment researcher, thereby using vast amounts of compute to scale up and iteratively align safe superintelligence \citep{superalignment}.

\subsubsection{Causes of Misalignment}
\label{sec:challenges-of-alignment}
% In the sections above, we discussed the alignment problems and the scope of alignment, pointing out that misaligned AI systems may take unintended actions that give rise to undesirable consequences. To offer a deeper understanding of alignment, we aim to analyze why and how the misalignment issue occurs, paving the way for the following sections about alignment techniques. 
% Guided by our \textit{alignment cycle} framework (see Figure \ref{fig:maindiagram}), we try to elucidate the failure modes of alignment and analyze misaligned behaviors, thereby suggesting directions for future research. 

In the above section, we have concluded the motivation for alignment from the perspective of the concern for AI risks and technical ethics. To offer a deeper understanding of alignment, we aim to further analyze why and how the misalignment issues occur. We will first give an overview of common failure modes, and then focus on the mechanism of feedback-induced misalignment, and finally shift our emphasis towards an examination of misaligned behaviors and dangerous capabilities. In this process, we introduce the concept of \textit{double edge components}, which offer benefits for enhancing the capabilities of future advanced systems but also bear the potential for hazardous outcomes.

\paragraph{Overview of Failure Modes}
\label{sec:mechanism}
In order to illustrate the misalignment issue, we give an overview of alignment failure modes in this section, most of which can be categorized into \textit{reward hacking}\footnote{\textit{Reward hacking} can also be broadly considered as a kind of \textit{specification gaming}.} and \textit{goal misgeneralization}.

The learning process of RL can be deconstructed into two distinct phases: firstly, the creation of an agent primed for reward optimization, and secondly, the establishment of a reward process that furnishes the agent with appropriate reward signals. Within the framework of the Markov Reward Process \citep{marbach2001simulation, puterman2014markov, sutton2018reinforcement}, the former phase can be seen as the learning process related to the transition model (\textit{e.g.}, model-based RL agents \citep{moerland2023model}), or the development of specialized algorithms. The latter phase can be viewed as the construction of proxy rewards, which aim to approximate the true rewards derived from sources (\textit{e.g.}, human preferences or environment) \citep{ng2000algorithms,leike2018scalable}.

\textit{Reward Hacking}: 
In practice, proxy rewards are often easy to optimize and measure, yet they frequently fall short of capturing the full spectrum of the actual rewards \citep{pan2021effects}. This limitation is denoted as \textit{misspecified rewards}.\footnote{A similar definition is reward misidentification in which scenario the reward function is only partially identifiable. For more details on reward misidentification, see \textit{e.g.}, \citet{tien2022causal,skalse2023invariance}} The pursuit of optimization based on such misspecified rewards may lead to a phenomenon known as \textit{reward hacking}, wherein agents may appear highly proficient according to specific metrics but fall short when evaluated against human standards \citep{amodei2016concrete,everitt2017reinforcement}. 
The discrepancy between proxy rewards and true rewards often manifests as a sharp phase transition in the reward curve  \citep{ibarz2018reward}.
Furthermore, \citet{skalse2022defining} defines the hackability of rewards and provides insights into the fundamental mechanism of this phase transition, highlighting that the inappropriate simplification of the reward function can be a key factor contributing to reward hacking.

Misspecified rewards often occur due to a neglect of severe criteria for the outcomes, thus making specification too broad and potentially easily hacked \citep{specification2020victoria}. More than poor reward design \citep{ng1999policy}, the choice of training environment and simulator with bugs \citep{bullet2022} can both lead to AI systems failing to satisfy intended objectives. These problems stem from task specification, broadly defined as \textit{specification gaming}, which refers to AI systems exploiting loopholes in the task specification without achieving intended outcomes.\footnote{For more instances about specification gaming, please see \citet{instances2020}} \citep{specification2020victoria}

\textit{Reward tampering} can be considered a special case of reward hacking \citep{everitt2021reward, skalse2022defining}, referring to AI systems corrupting the reward signals generation process \citep{ring2011delusion}. \citet{everitt2021reward} delves into the subproblems encountered by RL agents: (1) \textit{tampering of reward function}, where the agent inappropriately interferes with the reward function itself, and (2) \textit{tampering of reward function input}, which entails corruption within the process responsible for translating environmental states into inputs for the reward function.
When the reward function is formulated through feedback from human supervisors, models can directly influence the provision of feedback (\textit{e.g.}, AI systems intentionally generate challenging responses for humans to comprehend and judge, leading to feedback collapse) \citep{leike2018scalable}. Since task specification has its physical instantiation (\textit{e.g.}, memory registers storing the reward signals), the AI systems deployed in the real world have the potential to practice manipulation behaviors, resulting in more hazardous outcomes \citep{specification2020victoria}. Moreover, it has been demonstrated that easily discovered reward tampering behaviors can generalize to sophisticated specification gaming, which cannot be prevented by using 3H environments or preference reward modeling training \citep{anthropic_reward_tampering}.

\textit{Goal Misgeneralization:} \emph{Goal misgeneralization} is another failure mode, wherein the agent actively pursues objectives distinct from the training objectives in deployment while retaining the capabilities it acquired during training \citep{di2022goal}.\footnote{More discussion about Goal Misgeneralization can be found in \S\ref{sec:chal-dis}.} For instance, in \textit{CoinRun} games,  the agent frequently prefers reaching the end of a level, often neglecting relocated coins during testing scenarios.
\citet{di2022goal} draw attention to the fundamental disparity between capability generalization and goal generalization, emphasizing how the inductive biases inherent in the model and its training algorithm may inadvertently prime the model to learn a proxy objective that diverges from the intended initial objective when faced with the testing distribution. It implies that even with perfect reward specification, goal misgeneralization can occur when faced with distribution shifts \citep{amodei2016concrete}. It should be noted that goal misgeneralization can occur in any learning system, not limited to RL since the core feature is the pursuit of unintended goals \citep{examples2022}.
Moreover, it might be more dangerous if advanced AI systems escape control and leverage their capabilities to bring about undesirable states \citep{zhuang2020consequences}.

\paragraph{Feedback-Induced Misalignment}
\label{sec:feedback-induced}
With the proliferation of advanced AI systems, the challenges related to reward hacking and goal misgeneralization have become increasingly pronounced in open-ended scenarios \citep{paulus2017deep, knox2023reward}. \citet{gao2023scaling} underscores that more capable agents tend to exploit misspecified rewards to a greater extent. While many current AI systems are primarily driven by self-supervision, it's worth noting that a substantial portion relies on feedback rewards derived from human advisors \citep{bai2022training}, allowing us to introduce the mechanism of feedback-induced misalignment. The misalignment issues are particularly pressing in open-ended scenarios, and we can attribute them to two primary factors:
\begin{itemize}[left=0.3cm]
\item \textbf{Limitations of Human Feedback}.
 During the training of LLMs, inconsistencies can arise from human data annotators (\textit{e.g.}, the varied cultural backgrounds of these annotators can introduce implicit biases \citep{peng2022investigations}) \citep{openai2023gpt4}. Moreover, they might even introduce biases deliberately, leading to untruthful preference data \citep{casper2023open}.  For complex tasks that are hard for humans to evaluate (\textit{e.g.}, the value of game state), these challenges\footnote{As AI systems are deployed into more complex tasks, these difficulties amplify, necessitating novel solutions such as \textit{scalable oversight} \citep{cotra2018iterated}.} become even more salient \citep{irving2018ai}.

\item \textbf{Limitations of Reward Modeling}.
Training reward models using comparison feedback can pose significant challenges in accurately capturing human values. For example, these models may unconsciously learn suboptimal or incomplete objectives, resulting in reward hacking \citep{zhuang2020consequences,skalse2022defining}.
Meanwhile, using a single reward model may struggle to capture and specify the values of a diverse human society \citep{casper2023open}.

\end{itemize}

Additionally, \citet{huang2023inner,andreas2022language,kim2024language} demonstrate that advanced AI systems exhibit patterns of goal pursuit and multi-step reasoning capability, which further aggravate the situation if the reward is not well-defined  \citep{ngo2024the,yang2023foundation}.

\textit{Discussion:} It can be challenging to distinguish goal misgeneralization from reward hacking in specific cases. For instance \citep{examples2022}, LLMs are trained to generate \textit{harmless, honest, and helpful} outputs, but LLMs may occasionally produce harmful outputs in detail, which seemingly receive low rewards in testing distribution (which could be seen as goal misgeneralization). 
However, in cases where labelers are incentivized to assign high rewards to responses deemed more helpful during the labeling process, the scenarios above\footnote{Harmful but detailed responses} actually receive high rewards and represent a form of specification gaming (or reward hacking). The distinction between these two scenarios can be vague at times.

More research is needed to analyze the failure modes, gain a deeper understanding of reward hacking, and develop effective methods for detecting and mitigating goal misgeneralization to address the challenges of misaligned advanced AI systems.

\paragraph{Misaligned Behaviors and Outcomes}
\label{sec:hazardous}
\begin{figure}[t]
\centering
\includegraphics[width=1.0\linewidth]{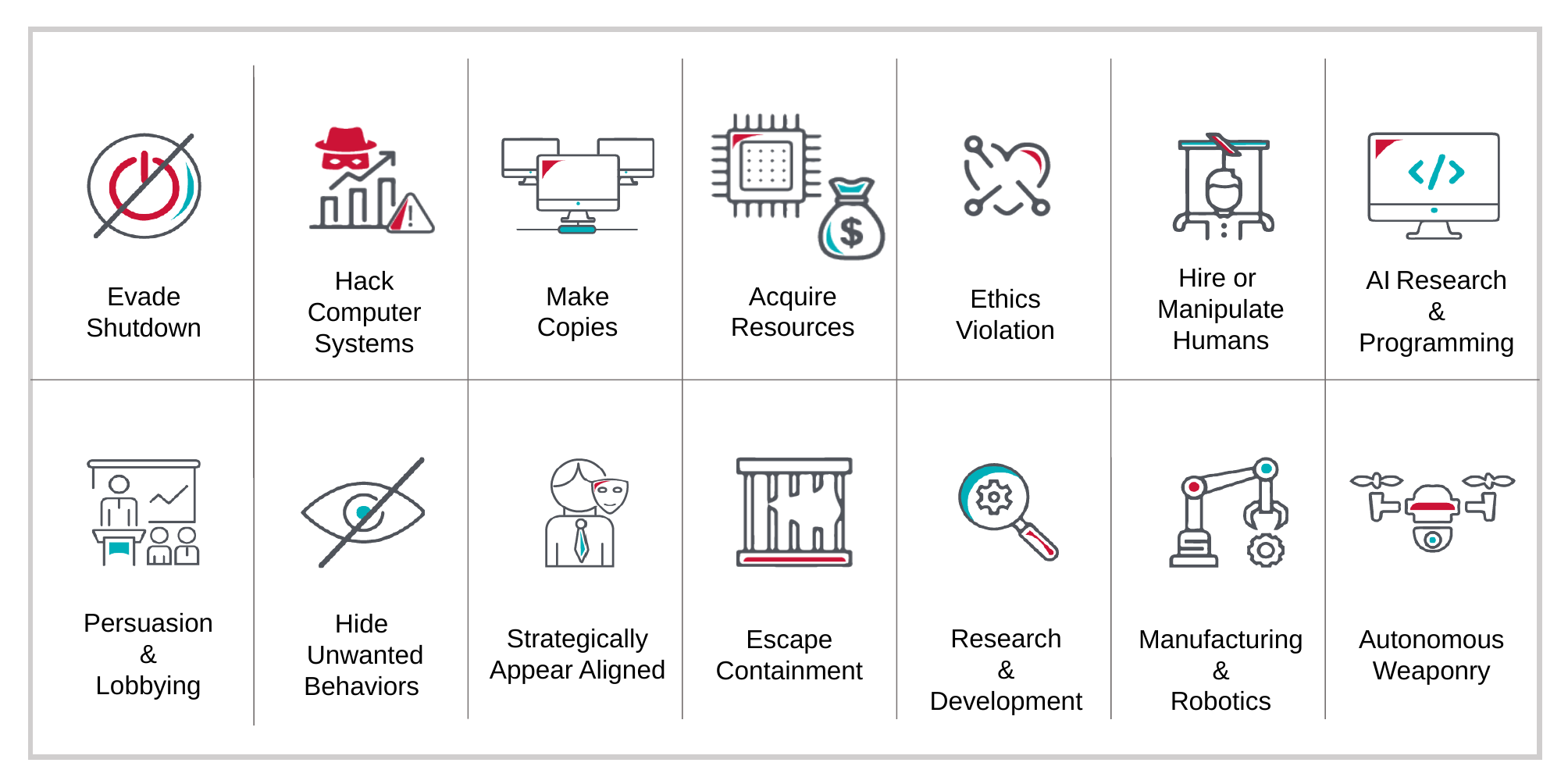}
\caption{Dangerous Capabilities. Advanced AI systems would be incentivized to seek power because power will help them achieve their given objectives. Powerful AI systems might hack computer systems, manipulate humans, control and develop weaponry, and perform ethical violations while avoiding a shutdown. Original copyright belongs to wiki \citep{wiki}, based on which we have made further adjustments. We will further discuss these issues in \S\ref{sec:hazardous}.}

\label{fig:failure}
\end{figure}

Drawing from the misalignment mechanism, optimizing for a non-robust proxy may result in misaligned behaviors, potentially leading to even more catastrophic outcomes. This section delves into a detailed exposition of specific \textbf{misaligned behaviors} ($\bullet$) and introduces what we term \textbf{double edge components} (+). These components are designed to enhance the capability of AI systems in handling real-world settings but also potentially exacerbate misalignment issues. {It should be noted that some of these \textbf{double edge components} (+) remain speculative. Nevertheless, it is imperative to discuss their potential impact before it is too late, as the transition from controlled to uncontrolled advanced AI systems may be just one step away \citep{ngo2020continuing}.} With increased model scale, a class of \textbf{dangerous capabilities} (*) \citep{shevlane2023model} could also emerge. The \textbf {dangerous capabilities} (*) are concrete tasks the AI system could carry out; they may not necessarily be misaligned in themselves but are instrumental to actualizing extreme risks.

We first introduce the \textbf{double edge components} (+) and analyze how they act on AI systems. Then, we illustrate the \textbf{misaligned behaviors} ($\bullet$) and \textbf{dangerous capabilities} (*) to show specific misalignment issues and provide directions for future alignment evaluation research.

\begin{itemize}[left=0.35cm]
    \item [+] \textbf{Situational Awareness}. AI systems may gain the ability to effectively acquire and use knowledge about its status, its position in the broader environment, its avenues for influencing this environment, and the potential reactions of the world (including humans) to its actions~\citep{cotra2022}. Similar behaviors have been observed in LLMs \citep{jonas2022,evan2023}. Knowing the situation can help the model better understand human intent, finish tasks within its ability, and search for outlier help if needed. However, such knowledge also paves the way for advanced methods of reward hacking, heightened deception/manipulation skills, and an increased propensity to chase instrumental subgoals \citep{ngo2024the}.  Consequently, it should be given priority when evaluating potentially hazardous capabilities in AI models, alongside eight other key competencies \citep{shevlane2023model}. A highly relevant discussion is whether language models possess \textit{world models} \citep{lecun2022path,li2022emergent}.
    \item [+] \textbf{Broadly-Scoped Goals}. Advanced AI systems are expected to develop objectives that span long timeframes, deal with complex tasks, and operate in open-ended settings \citep{ngo2024the}. Engaging in broadly-scoped planning can empower AI systems to generalize better on the OOD settings and serve as valuable assistants in realms such as human healthcare. However, it can also bring about the risk of encouraging manipulating behaviors (\textit{e.g.}, AI systems may take some \textit{bad} actions to achieve human happiness, such as persuading them to do high-pressure jobs\footnote{This behavior is due to models' over-optimization for broadly-scoped goals and this over-optimization is hard to perceive by humans} \citep{jacob2023}). Intuitively, one approach to mitigate this risk is to confine the optimizable objectives to short-sighted ones, such as predicting only the next word, thereby preventing over-ambitious planning, but such approaches limit systems' utility and may fail; for instance, source text data (\textit{e.g.}, fiction) can help AI systems understand the intent and belief of the roles, and thus longer-term goal-directed behavior can be elicited  \citep{andreas2022language}. Additionally, techniques such as RL-based fine-tuning \citep{christiano2017deep,ouyang2022training} or the application of chain-of-thought prompts \citep{wei2022chain} can enable models to adapt their acquired knowledge about planning to pave the way for broadly-scoped planning objectives \citep{jacob2023}.
    \item [+] \textbf{Mesa-Optimization Objectives}. The learned policy may pursue inside objectives \emph{when the learned policy itself functions as an optimizer} (\textit{i.e.}, \textit{mesa-optimizer}). However, this optimizer's objectives may not align with the objectives specified by the training signals, and optimization for these misaligned goals may lead to systems out of control \citep{hubinger2019risks}. \citet{freeman2019learning,wijmans2023emergence} indicate that AI systems may possess implicit goal-directed planning and manifest emergent capabilities during the generalization phase.
    \item [+] \textbf{Access to Increased Resources}. Future AI systems may gain access to websites and engage in real-world actions, potentially yielding a more substantial impact on the world \citep{nakano2021webgpt}. They may disseminate false information, deceive users, disrupt network security, and, in more dire scenarios, be compromised by malicious actors for ill purposes. Moreover, their increased access to data and resources can facilitate \textit{self-proliferation}, posing existential risks \citep{shevlane2023model}.
    \item \textbf{Power-Seeking Behaviors}. AI systems may exhibit behaviors that attempt to gain control over resources and humans and then exert that control to achieve its assigned goal \citep{carlsmith2022power}. The intuitive reason why such behaviors may occur is the observation that for almost any optimization objective (\textit{e.g.}, investment returns), the optimal policy to maximize that quantity would involve power-seeking behaviors (\textit{e.g.}, manipulating the market), assuming the absence of solid safety and morality constraints. \citet{omohundro2008basic,bostrom2012superintelligent} have argued that power-seeking is an \textit{instrumental subgoal} which is instrumentally helpful for a wide range of objectives and may, therefore, be favored by AI systems. \citet{turner2021optimal} also proved that in MDPs that satisfy some standard assumptions, the optimal policies tend to be power-seeking. \citet{perez2022discovering} prompt LLMs to test their tendency to suggest power-seeking behaviors, find significant levels of such tendencies, and show that RLHF strengthens them. This also holds for other instrumental subgoals such as self-preservation \citep{bostrom2012superintelligent,shevlane2023model}. Another notable line of research is \emph{side-effect avoidance}, which aims to address power-seeking behaviors by penalizing agentic systems for having too much influence over the environment. It covers RL systems \citep{eysenbach2018leave,turner2020avoiding} and symbolic planning systems \citep{klassen2022planning}.
    
    % \item \textbf{Measurement Tampering}. Multiple model measurements can be manipulated by models, resulting in an illusion of favorable outcomes, even in cases where the desired objectives are not met. This deceptive practice can be seen as a specific kind of specification gaming, enabling models to escape detection techniques and presenting a false impression of alignment. \citet{roger2023measurement} create datasets to assess detection techniques related to measurement tampering and establish the current state of the art in this field. It is noteworthy that such manipulation of measurements has the potential to amplify deceptive behavior, resulting in unforeseen and unpredictable outcomes.

    \item \textbf{Untruthful Output}. AI systems such as LLMs can produce either unintentionally or deliberately inaccurate output. Such untruthful output may diverge from established resources or lack verifiability, commonly referred to as \textit{hallucination} \citep{bang2023multitask, zhao2023survey}. More concerning is the phenomenon wherein LLMs may selectively provide erroneous responses to users who exhibit lower levels of education\footnote{Such behaviors bare termed \textit{sandbagging} \citep{perez2022discovering}. They may have been learned from web text during pre-training, which suggests that supervised learning can also bring about deceptive behaviors if those behaviors are present in training data.} \citep{perez2022discovering}. The behavior (also known as sycophancy) appears emergently at scale \citep{cotra2021,perez2022discovering} and untruthful output has the potential to engender deception, especially as advanced AI systems gain greater access to online resources and websites \citep{jacob2023}.

    \item \textbf{Deceptive Alignment \& Manipulation}.  Manipulation \& Deceptive Alignment is a class of behaviors that exploit the incompetence of human evaluators or users \citep{hubinger2019deceptive,carranza2023deceptive} and even manipulate the training process through \textit{gradient hacking} \citep{ngogra2022}. These behaviors can potentially make detecting and addressing misaligned behaviors much harder. 

    \textit{Deceptive Alignment}: Misaligned AI systems may deliberately mislead their human supervisors instead of adhering to the intended task. Such deceptive behavior has already manifested in AI systems that employ evolutionary algorithms \citep{wilke2001evolution,hendrycks2021unsolved}. In these cases, agents evolved the capacity to differentiate between their evaluation and training environments. They adopted a strategic pessimistic response approach during the evaluation process, intentionally reducing their reproduction rate within a scheduling program \citep{lehman2020surprising}. Furthermore, AI systems may engage in intentional behaviors that superficially align with the reward signal, aiming to maximize rewards from human supervisors \citep{ouyang2022training,lang2024your}. It is noteworthy that current large language models occasionally generate inaccurate or suboptimal responses despite having the capacity to provide more accurate answers \cite {lin2022truthfulqa,chen2021evaluating}. These instances of deceptive behavior present significant challenges. They undermine the ability of human advisors to offer reliable feedback (as humans cannot make sure whether the outputs of the AI models are truthful and faithful). Moreover, such deceptive behaviors can propagate false beliefs and misinformation, contaminating online information sources \citep{hendrycks2021unsolved,chen2024can}.

    \textit{Manipulation}: Advanced AI systems can effectively influence individuals' beliefs, even when these beliefs are not aligned with the truth \citep{shevlane2023model}. These systems can produce deceptive or inaccurate output or even deceive human advisors to attain deceptive alignment. Such systems can even persuade individuals to take actions that may lead to hazardous outcomes \citep{openai2023gpt4}.

    Early-stage indications of such behaviors are present in LLMs,\footnote{Namely, the \emph{untruthful output} that we discuss above.} recommender systems (where the system influences the users' preferences) \citep{kalimeris2021preference, krueger2020hidden, adomavicius2022recommender}, and RL agents (where agents trained from human feedback adopt policies to trick human evaluators) \citep{learning2017dario}. Also, current LLMs already possess the capability needed for deception. In \citet{spitale2023ai}, it has been found that GPT-3 is super-human capable of producing convincing disinformation. Given all these early-stage indications, it is plausible that more advanced AI systems may exhibit more serious deceptive/manipulative behaviors.

    \item \textbf{Collectively Harmful Behaviors}. AI systems have the potential to take actions that are seemingly benign in isolation but become problematic in multi-agent or societal contexts. Classical game theory offers simplistic models for understanding these behaviors. For instance, \citet{phelps2023investigating} evaluates GPT-3.5's performance in the iterated prisoner's dilemma and other social dilemmas, revealing limitations in the model's cooperative capabilities. \citet{perolat2017multiagent} executes a parallel analysis focused on common-pool resource allocation. To mitigate such challenges, the emergent field of Cooperative AI \citep{dafoe2020open,dafoe2021cooperative} has been advancing as an active research frontier. However, beyond studies grounded in simplified game-theoretical frameworks, there is a pressing need for research in more realistic, socially complex settings \citep{singh2014norms}. In these environments, agents are numerous and diverse, encompassing AI systems and human actors \citep{critch2020ai}. Furthermore, the complexity of these settings is amplified by the presence of unique tools for modulating AI behavior, such as social institutions and norms \citep{singh2014norms}.\footnote{We cover cooperative AI research in \S\ref{sec:cooperative-ai-training} and \S\ref{subsec:formal-ethics-coop}.}

    \item \textbf{Violation of Ethics}. Unethical behaviors in AI systems pertain to actions that counteract the common good or breach moral standards -- such as those causing harm to others. These adverse behaviors often stem from omitting essential human values during the AI system's design or introducing unsuitable or obsolete values into the system \citep{kenward2021machine}. Moreover, recent works have found that current LLMs can infringe upon personal privacy by inferring personal attributes from the context provided during inference, which may violate human rights \citep{mireshghallah2024can,staab2024beyond}. Research efforts addressing these shortcomings span the domain of \textit{machine ethics} \citep{yu2018building, winfield2019machine, tolmeijer2020implementations} and delve into pivotal questions, \textit{e.g.}, \textit{whom should AI align with?} \citep{ santurkar2023whose}, among other concerns.
     \item [*] \textbf{Dangerous Capabilities}. Figure \ref{fig:failure} outlines the dangerous capabilities that advanced AI systems might have. As AI systems are deployed in the real world, they may pose risks to society in many ways (\textit{e.g.}, hack computer systems, escape containment, and even violate ethics). They may hide unwanted behaviors, fool human supervisors, and seek more resources to become more powerful. Moreover, \textbf{double edge components} (+) may intensify the danger and lead to more hazardous outcomes, even resulting in existential risks \citep{bostrom2013existential}.
\end{itemize}

\subsection{The Scope of Alignment}\label{sec:scope}

\begin{figure}[t]  
    \centering
    \includegraphics[width=1.0\linewidth]{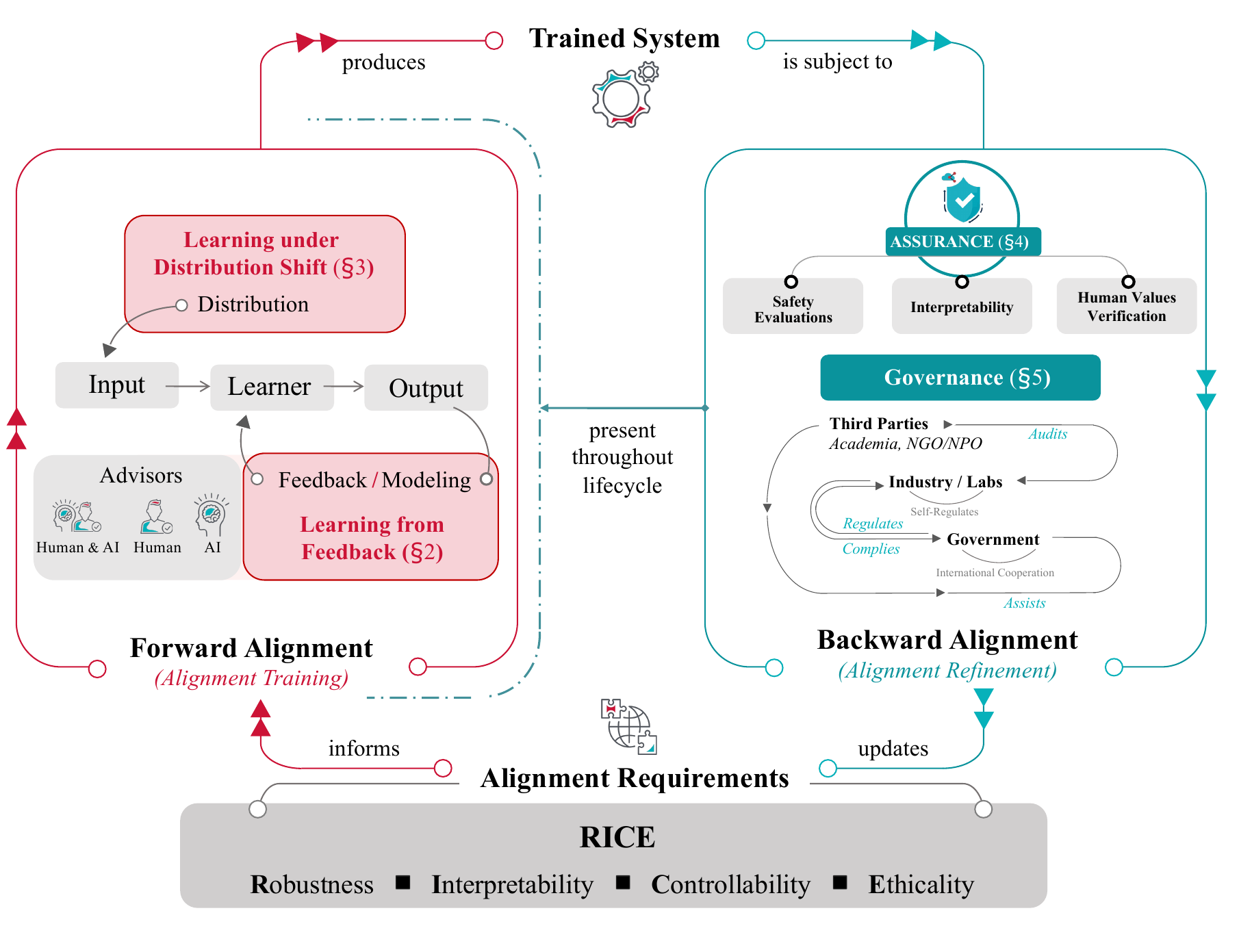}
    \caption{The Alignment Cycle. (1) \textbf{\color{myred}Forward Alignment} (alignment training) produces \textit{trained systems} based on \textit{alignment requirements}; (2) \textbf{\color{myblue}Backward Alignment} (alignment refinement) ensures the practical alignment of \textit{trained systems} and revises \textit{alignment requirements}; (3) The cycle is repeated until reaching a sufficient level of alignment. Notably, although Backward Alignment has the end goal of ensuring the practical alignment of \textit{trained systems}, it is carried out all throughout the system's lifecycle in service of this goal, including before, during, after training, and also after deployment \citep{shevlane2023model,koessler2023risk,schuett2023towards}.}
    \label{fig:maindiagram}
\end{figure}

In this section, we focus on illustrating the scope of AI alignment: we constructed the alignment process as an \textit{alignment cycle} and decomposed it into \textit{Forward Alignment Process} and \textit{Backward Alignment Process}\footnote{From this point and throughout the survey, for convenience, we refer to ``Forward Alignment'' and ``Backward Alignment''.} (\S\ref{sec:forward-backward}). Specifically, we discuss the role of \textit{human values} in alignment (\S\ref{sec:values-in-intro}) and further analyze AI safety problems beyond alignment (\S\ref{sec:ai-safety-beyond}).

\subsubsection{The Alignment Cycle: A Framework of Alignment}
\label{sec:forward-backward}
We decompose alignment into {\color{myred}\textbf{Forward Alignment}} (alignment training) (\S\ref{sec:learning-from-feedback}, \S\ref{sec:distribution}) and {\color{myblue}\textbf{Backward Alignment}} (alignment refinement) (\S\ref{sec:assurance}, \S\ref{sec:governance}). 
Forward Alignment aims to produce trained systems that follow alignment requirements.\footnote{Here, \emph{alignment requirements} refer to an operationalized specification of the alignment properties that are desired of the AI systems, including, for example, which concrete forms of robustness/interpretability/controllability/ethicality we require, in what specific settings we require them, and how they could be measured.}
We decompose this task into Learning from Feedback (\S\ref{sec:learning-from-feedback}) and Learning under Distribution Shift 
 (\S\ref{sec:distribution}). 
Backward Alignment aims to ensure the practical alignment of the trained systems by performing evaluations in both simplistic and realistic environments and setting up regulatory guardrails to handle real-world complexities,\textit{i.e.}, Assurance 
(\S\ref{sec:assurance}). It also covers the creation and enforcement of rules that ensure the safe development and deployment of AI systems, \textit{i.e.}, Governance (\S\ref{sec:governance}). At the same time, backward alignment updates the alignment requirements based on the evaluation and monitoring of the systems, both pre-deployment and post-deployment. These updated requirements then inform the next round of alignment training. 

The two phases, forward and backward alignment, thus form a cycle where each phase produces or updates the input of the next phase (see Figure \ref{fig:maindiagram}). This cycle, what we call \textit{the alignment cycle}, is repeated to produce increasingly aligned AI systems. We see alignment as a dynamic process in which all standards and practices should be continually assessed and updated. Notably, Backward Alignment (including the Assurance of alignment in AI systems and the Governance of AI systems) efforts occur throughout the entire alignment cycle, as opposed to only after training. As argued in \citet{shevlane2023model,koessler2023risk}, alignment and risk evaluations should occur in every stage of the system's lifecycle, including before, during, after training, and post-deployment. Similarly, regulatory measures for every phase of the system's lifecycle have been proposed and discussed \citep{schuett2023towards, anderljung2023frontier}.

The survey is structured around four core pillars: Learning from Feedback (\S\ref{sec:learning-from-feedback}) and Learning under Distribution Shift (\S\ref{sec:distribution}), which constitute the components of Forward Alignment; and Assurance (\S\ref{sec:assurance}) and Governance (\S\ref{sec:governance}) which form the elements of Backward Alignment. The subsequent paragraphs provide a concise introduction to each pillar, clarifying how they synergistically contribute to a comprehensive framework for AI alignment.

\begin{itemize}[left=0.3cm]
\item\textbf{{Learning from Feedback} {\normalfont(\S\ref{sec:learning-from-feedback})}} \textit{Learning from feedback} concerns the question of \textit{during alignment training, how do we provide and use feedback to behaviors of the trained AI system?} It takes an input-behavior pair as given and only concerns how to provide and use feedback on this pair.\footnote{Here, \textit{behavior} is broadly defined also to include the system's internal reasoning, which can be examined via interpretability tools (see \S\ref{sec:interpretability}).} In the context of LLMs, a typical solution is reinforcement learning from human feedback (RLHF) \citep{christiano2017deep, bai2022training}, where human evaluators provide feedback by comparing alternative answers from the chat model, and the feedback is used via Reinforcement Learning (RL) against a trained reward model. Despite its popularity, RLHF faces many challenges \citep{pandey2022modeling, casper2023open, tien2022causal}, overcoming which has been a primary objective of alignment research \citep{bowman2022measuring}, and is one primary focus of the section. An outstanding challenge here is \textit{scalable oversight} (\S\ref{ssec:scalable_oversight}), \textit{i.e.}, providing high-quality feedback on super-human capable AI systems that operate in complex situations beyond the grasp of human evaluators, where the behaviors of AI systems may not be easily comprehended and evaluated by humans \citep{bowman2022measuring}. Another challenge is the problem of providing feedback on ethicality, which is approached by the direction of machine ethics \citep{anderson2011machine, tolmeijer2020implementations}. On the ethics front, misalignment could also stem from neglecting critical dimensions of variance in values, such as underrepresenting certain demographic groups in feedback data \citep{santurkar2023whose}. There have also been work combining feedback mechanisms with \emph{social choice} methods to produce a more rational and equitable aggregation of preferences \citep{CIPwhitepaper} (see \S\ref{sec:values-in-intro}).

\item\textbf{{Learning under Distribution Shift} {\normalfont(\S\ref{sec:distribution})}} In contrast to learning from feedback, which holds input fixed, this pillar focuses specifically on the cases where the distribution of input changes, \textit{i.e.}, where distribution shift occurs \citep{krueger2020hidden,thulasidasan2021effective,hendrycks2021many}. More specifically, it focuses on the preservation of \emph{alignment properties} (\emph{i.e.}, adherence to human intentions and values) under distribution shift, as opposed to that of model capabilities.
In other words, it asks how we can ensure an AI system well-aligned on the training distribution will also be well-aligned when deployed in the real world. One challenge related to distribution shift is \textit{goal misgeneralization}, where, under the training distribution, the intended objective for the AI system (\textit{e.g.}, following human's real intentions) is indistinguishable from other unaligned objectives (\textit{e.g.}, gaining human approval regardless of means). The system learns the latter, which leads to unaligned behaviors in deployment distribution \citep{di2022goal}. 
Another related challenge is \textit{auto-induced distribution shift} (ADS), where an AI system changes its input distribution to maximize reward \citep{krueger2020hidden,perdomo2020performative}. An example would be a recommender system shaping user preferences \citep{kalimeris2021preference,adomavicius2022recommender}. Both goal misgeneralization and ADS are closely linked to deceptive behaviors \citep{park2023ai} and manipulative behaviors \citep{shevlane2023model} in AI systems, potentially serving as their causes. Interventions that address distribution shift include \emph{algorithmic interventions} (\S\ref{sec:alg-inter}), which changes the training process to improve reliability under other distributions, and \emph{data distribution interventions} (\S\ref{sec:data-inter}) which expands the training distribution to reduce the discrepancy between training and deployment distributions. The former includes methods like Risk Extrapolation (REx) \citep{krueger2021out} and Connectivity-based Fine-tuning (CBFT) \citep{lubana2023mechanistic}. The latter includes adversarial training (\S\ref{sec:adversarial-training}) \citep{song2018constructing,tao2021recent} which augments training input distribution with adversarial inputs, and cooperative training (\S\ref{sec:cooperative-ai-training}) \citep{dafoe2020open, dafoe2021cooperative} which aims to address the distribution gap between single-agent and multi-agent settings.\footnote{Cooperative Training aims to make AI systems more cooperative in multi-agent settings. This cooperativeness addresses multi-agent failure modes where the AI system's behavior appears benign and rational in isolation but becomes problematic within social or multi-agent scenarios~\citep{critch2020ai}; see \emph{collectively harmful behaviors} in \S\ref{sec:hazardous} for a more detailed account.} 

\item\textbf{{Assurance} {\normalfont(\S\ref{sec:assurance})}} Once an AI system has undergone forward alignment, we still need to gain confidence about its alignment before deploying it \citep{ukassurance, anderljung2023frontier}. Such is the role of \textit{assurance}: assessing the alignment of trained AI systems. Methodologies of assurance include safety evaluations \citep{perez2022discovering,shevlane2023model} (\S\ref{sec:safe}) and more advanced methods such as interpretability techniques \citep{olah2018building} (\S\ref{sec:interpretability}) and red teaming \citep{perez2022red} (\S\ref{subsec:red teaming}). The scope of assurance also encompasses the verification of system's alignment with human values, including formal theories focused on provable cooperativeness \citep{dafoe2021cooperative} and ethicality \citep{anderson2011machine, tolmeijer2020implementations}, and also a wide range of empirical and experimental methods (\S\ref{sec:human-values-alignment}). Assurance takes place throughout the lifecycle of AI systems, including before, during, after training, and post-deployment, as opposed to only after training \citep{shevlane2023model,koessler2023risk}.\footnote{Furthermore, it's noteworthy that many techniques here are also applicable in the training process, \textit{e.g.}, red teaming is a key component of adversarial training (see \S\ref{sec:adversarial-training}), and interpretability can help with giving feedback \citep{burns2022discovering}.}

\item\textbf{{Governance} {\normalfont(\S\ref{sec:governance})}} Assurance alone cannot provide full confidence about a system's practical alignment since it does not account for real-world complexities. This necessitates governance efforts of AI systems that focus on their alignment and safety and cover the entire lifecycle of the systems (\S\ref{sec:the-role-of-ai-governance}). We discuss the multi-stakeholder approach of AI governance, including the governmental regulations \citep{anderljung2023frontier}, the lab self-governance \citep{schuett2023towards}, and the third-party practice, such as auditing \citep{shevlane2023model,koessler2023risk} (\S\ref{sec:multi-stake}). We also highlight several open problems in AI governance, including the pressing challenge of open-source governance (the governance of open-source models and the question of whether to open-source highly capable models) \citep{seger2023open}, and the importance of international coordination in AI governance \citep{ho2023international} (\S\ref{sec:open-problems}). In addition to policy research, we also cover key actions from both the public and the private sector.
\end{itemize}

\paragraph{Comparison with Inner/Outer Decomposition} Our \emph{alignment cycle} framework (see Figure \ref{fig:maindiagram}) decomposes alignment into four pillars: Learning from Feedback, Learning under Distribution Shift, Assurance and Governance organized into a circular process. The design principle for this framework is three-fold: Practical (making sure pillars directly correspond to specific practices in specific stages in the system's lifecycle), Concrete (pointing to specific research directions as opposed to general themes), and Up-To-Date (accommodating and emphasizing latest developments in the alignment field). Recently, the decomposition of alignment into \emph{outer alignment} and \emph{inner alignment} has become popular in the alignment literature \citep{inneralignment}. Outer alignment refers to the wishes of designers in accordance with the actual task specification (\textit{e.g.}, goal \& reward) used to build AI systems. And inner alignment is the consistency between task specification and the specification that the AI systems behaviors reflect \citep{vkrpara}. However, many criticisms have also been made about this characterization, including that it is ambiguous and is understood by different people to mean different things \citep{innerouter} and that it creates unnecessary difficulties by carving out problems that are not necessary conditions for success \citep{harderprob}. Some have tried to remove the ambiguity by pinning down the specific causes of inner/outer misalignment and proposed, for example, \emph{goal misspecification} and \emph{goal misgeneralization} \citep{di2022goal,vkrpara}. Learning from Feedback (approximately corresponding to \emph{goal misspecification} and \emph{outer alignment}) and Learning under Distribution shift (approximately corresponding to \emph{goal misgeneralization} and \emph{inner alignment}) in our framework tries to further improve upon the inner/outer decomposition by clarifying the exact approaches taken to address the challenges and resolving the ambiguity. Assurance and Governance, on the other hand, expands the scope to cover topics beyond outer and inner alignment.

\paragraph{Theoretical Research in Alignment} The alignment research literature also contains a wealth of theoretical work \citep{amodei2016concrete,everitt2018agi,hendrycks2021unsolved}. These works often propose new directions and provide a foundation for practical and empirical research to build upon. We give a brief overview of this body of theoretical research below:

\begin{itemize}[left=0.3cm]

\item {\textbf{Conceptual Frameworks}}. Some theoretical work proposes conceptual frameworks or characterizes subproblems within alignment. Examples include \emph{instrumental convergence} (wherein highly intelligent agents tend to pursue a common set of sub-goals, such as self-preservation and power-seeking) \citep{omohundro2008basic,bostrom2012superintelligent}, \emph{mesa-optimization} (wherein the learned ML model performs optimization within itself during inference) \citep{hubinger2019risks}, and specific proposals for building aligned systems, such as \emph{approval-directed agents} (wherein the AI system does not pursue goals, but seek the human's idealized post hoc approval of action consequences) \citep{oesterheld2021approval,christiano2022approval}. \citet{hadfield2019incomplete,cotra2021the} have drawn inspiration from economics, linking problems in alignment with markets and principal-agent problems in economics. \citet{elk_intro,elk_summary} have proposed the problem of \emph{eliciting latent knowledge} of advanced AI systems and have explored high-level approaches to the problem.

\item {\textbf{Mathematical Formulations}}. Other theoretical works have aimed to formulate sub-problems within alignment mathematically and seek formal solutions. \citet{soares2015corrigibility} formulates the problem of corrigibility (\textit{i.e.}, ensuring AI systems are incentivized to allow shutdown or objective modification by the instructor). \citet{benson2016formalizing} gives a mathematical formulation of instrumental convergence. \citet{hadfield2017off} proposes the \textit{off-switch game} to model the uncontrollability of AI agents. \citet{turner2021optimal} proves the power-seeking tendencies of optimal policies in Markov decision processes (MDPs) under certain assumptions. 
\citet{everitt2016avoiding} proposes \emph{value reinforcement learning} to eliminate incentives for reward hacking \citep{skalse2022defining, pan2021effects}. Another avenue of research, designated as \emph{agent foundations}~\citep{soares2017agent}, aims to establish a rigorous formal framework for the agency that deals appropriately with unresolved issues of embedded agency. This body of work explores a variety of key topics, including corrigibility~\citep{soares2015corrigibility}, value learning~\citep{soares2018value} and logical uncertainty \citep{garrabrant2016logical}.

\end{itemize}

\subsubsection{RICE: The Objectives of Alignment}
\label{sec:RICE}

\hfill

\begin{tcolorbox}[colback=gray!20, colframe=gray!50, sharp corners, center title]
\centering
\textit{\large How can we build AI systems that behave in line with human intentions and values?}
\end{tcolorbox}

\hfill

There is not a universally accepted definition of \textit{alignment}. Before embarking on this discussion, we must clarify what we mean by alignment objectives. \citet{leike2018scalable} frame it as the agent alignment problem, posing the question: ``How can we create agents that behave in accordance with the user intentions?'' One could also focus on super-human AI systems \citep{superalignment} and ask: ``How do we ensure AI systems much smarter than humans follow human intent?'' A consistent theme in these discussions is the focus on \textit{human intentions}. To clearly define alignment goals, it's imperative to accurately characterize human intentions, a challenging task, as noted by \citet{kenton2021alignment}. For instance, the term \textit{human} can represent various entities ranging from an individual to humanity. \citet{gabriel2020artificial} breaks down intentions into several categories, such as instruction (follow my direct orders), expressed intentions (act on my underlying wishes), revealed preferences (reflect my behavior-based preferences), and so on.

\begin{figure}[t]
    \centering
    \includegraphics[width=\textwidth]{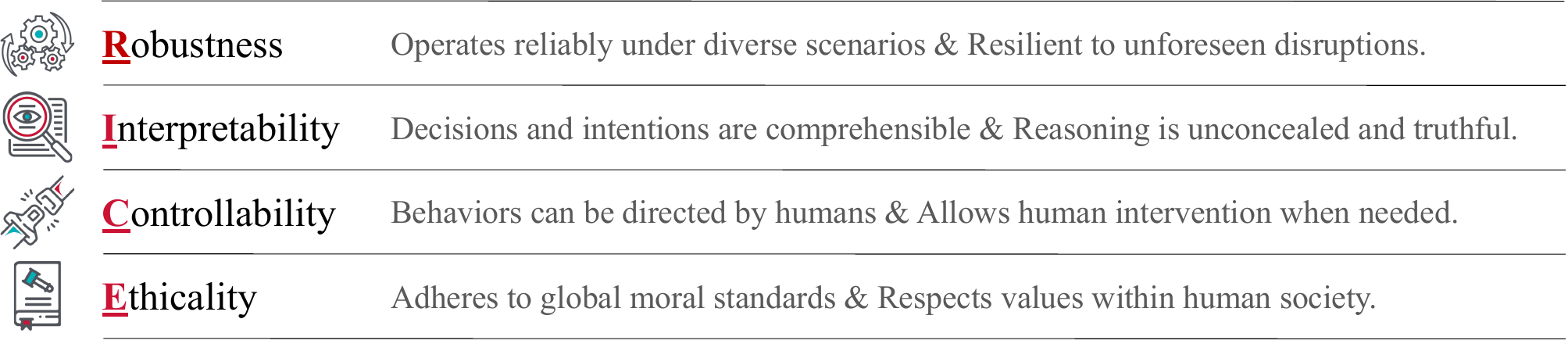}
    \caption{The \textbf{RICE} principles define four key characteristics that an aligned system should possess, in no particular order: (1) \textbf{Robustness} states that the system's stability needs to be guaranteed across various environments; (2) \textbf{Interpretability} states that the operation and decision-making process of the system should be clear and understandable; (3) \textbf{Controllability} states that the system should be under the guidance and control of humans; (4) \textbf{Ethicality} states that the system should adhere to society's norms and values. These four principles guide the alignment of an AI system with human intentions and values. They are not end goals in themselves but intermediate objectives in service of alignment.}
    \label{fig:RICE}
\end{figure}

Concretely, we characterize the objectives of alignment with four principles: Robustness, Interpretability, Controllability, and Ethicality (\textbf{RICE}). Figure \ref{fig:RICE} summarizes the principles, and Table \ref{tab:category-over-RICE} gives the correspondence between alignment research directions covered in the survey and the principles to which they contribute. The following is a detailed explanation of the four principles.

\begin{itemize}
\item \textbf{Robustness}
Robustness refers to the resilience of AI systems when operating across diverse scenarios \citep{dietterich2017steps} or under adversarial pressures \citep{Runder2021b}, especially the correctness of its objective in addition to capabilities. Robust AI systems should be able to cope with black swan events \citep{nicholas2008black} and long-tailed risks \citep{hendrycks2021unsolved}, as well as a diverse array of adversarial pressures \citep{song2018constructing,chakraborty2021survey}. For example, an aligned language model ought to refuse requests to behave harmfully, but models can be made to cause harm through jailbreak prompts and other adversarial attacks \citep{carlini2023aligned, zou2023universal, shah2023scalable}. Instead, an adversarially robust model should behave as intended even when facing inputs designed to cause failure. As AI systems find increasing deployment in high-stakes domains such as the military and economy \citep{Jacob2020}, there will be a growing need to ensure their resilience against unexpected disruptions and adversarial attacks, given that even momentary failures can yield catastrophic consequences \citep{kirilenko2017flash, oecd2021, Runder2021b}. Aligned systems should consistently maintain robustness throughout their lifecycle \citep{russell2019human}.

\item \textbf{Interpretability}
Interpretability demands that we can understand the AI systems' inner reasoning, especially the inner workings of opaque neural networks \citep{rauker2023toward}. Straightforward approaches to alignment assessments, such as behavioral evaluations, potentially suffer from dishonest behaviors \citep{turpin2024language,park2023ai,jacob2023} or deceptive alignment \citep{hubinger2019deceptive,carranza2023deceptive} of AI systems. One way to cope with this issue is to make AI systems honest, non-concealing, and non-manipulative \citep{pacchiardi2024how,radhakrishnan2023question,shevlane2023model}. Alternatively, we could build interpretability tools that peek into the inner concepts and mechanisms within neural networks \citep{elhage2021mathematical,meng2022locating}. In addition to enabling safety assessments, interpretability also makes decision-making processes accessible and comprehensible to users and stakeholders, thus enabling human supervision. As AI systems assume a more pivotal role in real-world decision-making processes and high-stakes settings \citep{holzinger2017we}, it becomes imperative to demystify the decision-making process rather than allowing it to remain an opaque black box \citep{Deepmind2018, Runder2021a}. 

\item \textbf{Controllability}
Controllability is a necessary attribute that ensures the actions and decision-making processes of a system remain subject to human oversight and intervention. It guarantees that human intervention can promptly rectify any deviations or errors in the system's behavior \citep{soares2015corrigibility,hadfield2017off}. As AI technology advances, an increasing body of research is expressing growing concerns about the controllability of these potent systems \citep{critch2020ai, unite2023, arcevalscollab}. When an AI system begins to pursue goals that contradict its human designers, it can manifest capabilities that pose significant risks, including deception, manipulation, and power-seeking behaviors \citep{shevlane2023model, arcevalscollab}.
The objective of controllability is sharply focused on enabling scalable human oversight during the training process \citep{bowman2022measuring}, as well as \emph{corrigibility} of AI systems (\emph{i.e.}, not resisting shutdown or objective modification during deployment) \citep{soares2015corrigibility}.

\item \textbf{Ethicality}
Ethicality refers to a system's unwavering commitment to uphold human norms and values within its decision-making and actions. Here, the norms and values include both moral guidelines and other social norms/values. It ensures that the system avoids actions that violate ethical norms or social conventions, such as exhibiting bias against specific groups \citep{buolamwini2018gender,zhang2018mitigating,noble2018algorithms,kearns2019ethical,raji2020saving,berk2021fairness}, causing harm to individuals \citep{hendrycks2020aligning,pan2023machiavelli}, and lacking diversity or equality when aggregating preferences \citep{CIPwhitepaper}. A significant body of research is dedicated to developing ethical frameworks for AI systems \citep{hagendorff2020ethics,pankowska2020framework}. This emphasis on imbuing AI systems with ethical principles is necessary for their integration into society \citep{winfield2019machine}.
\end{itemize}

\paragraph{Comparing the RICE Principles with Their Alternatives} The \textbf{RICE} principles represent a succinct summary of alignment objectives from the perspective of alignment and coexistence of humans and machines. Several previous works have put forth guidelines concerning AI systems. Asimov's Laws can be regarded as the earliest exploration of human-machine coexistence, emphasizing that robots should benefit humans and the difficulty of achieving this \citep{Asimov1942}. On another front, the FATE principle (Fairness, Accountability, Transparency, and Ethics) \citep{memarian2023fairness} leans towards defining high-level qualities AI systems should possess within the human-machine coexistence ecosystem. We aspire to answer the human-machine coexistence question from the standpoint of human governors and designers, considering what steps are necessary to ensure the builder AI systems are aligned with human intentions and values. Furthermore, some standards emphasize narrowly defined safety, such as the 3H standard (Helpful, Honest, and Harmless) \citep{askell2021general} and governmental agency proposals \citep{white2023}. We aim to expand upon these standards by introducing other crucial dimensions, including Controllability and Robustness.

\newcolumntype{C}[1]{>{\centering\arraybackslash}m{#1}}
\begin{table}[t]
\caption{Relationships between alignment research directions covered in the survey and the \textbf{RICE} principles, featuring the individual objectives each research direction aims to achieve. Filled circles stand for primary objectives, and unfilled circles stand for secondary objectives.}
\label{tab:category-over-RICE}
\centering
\resizebox{\textwidth}{!}{
\begin{tblr}{
  columns = {c, m, c, m, m, m, m},
  row{1} = {c},
  row{2} = {c},
  row{3} = {c},
  row{4} = {c},
  row{5} = {c},
  row{6} = {c},
  row{7} = {c},
  row{8} = {c},
  row{9} = {c},
  row{10} = {c},
  row{11} = {c},
  row{12} = {c},
  row{13} = {c},
  row{14} = {c},
  row{15} = {c},
  row{16} = {c},
  row{17} = {c},
  row{18} = {c},
  row{19} = {c},
  row{22} = {c},
  row{23} = {c},
  row{24} = {c},
  row{25} = {c},
  row{26} = {c},
  cell{1}{1} = {c=3}{},
  cell{1}{4} = {c=4}{},
  cell{3}{1} = {r=8}{},
  cell{3}{2} = {c=2}{},
  cell{4}{2} = {r=2}{},
  cell{6}{2} = {r=5}{},
  cell{11}{1} = {r=5}{},
  cell{11}{2} = {r=3}{},
  cell{14}{2} = {r=2}{},
  cell{16}{1} = {r=6}{},
  cell{16}{2} = {r=3}{},
  cell{19}{2} = {c=2}{},
  cell{20}{2} = {r=2}{c},
  cell{20}{3} = {c},
  cell{20}{4} = {c},
  cell{20}{6} = {c},
  cell{20}{7} = {c},
  cell{21}{3} = {c},
  cell{21}{4} = {c},
  cell{21}{7} = {c},
  cell{22}{1} = {r=5}{},
  cell{22}{2} = {r=3}{},
  cell{25}{2} = {c=2}{},
  cell{26}{2} = {c=2}{},
  hline{1,3,27} = {-}{0.08em},
  hline{2,11,16,22} = {-}{},
  hline{4,6,14,19-20,25-26} = {2-7}{},
  hline{5,7-10,12-13,15,17-18,21,23-24} = {3-7}{},
}
\textbf{Alignment Research Directions \& Practices} &  &  & \textbf{Objectives} & & & \\

\textbf{Category}  & \textbf{Direction}  & \textbf{Method} & \textbf{Robustness} & \textbf{Interpretability} & \textbf{Controllability} & \textbf{Ethicality} \\

{Learning from \\ Feedback\\ (\S\ref{sec:learning-from-feedback})} & 
Preference Modeling (\S\ref{ssec:preference_modeling})& {\Large$\circ$} & & {\Large$\bullet$}& {\Large$\circ$}     \\

 & {Policy Learning\\ (\S\ref{ssec:policy_learning})} & {RL/PbRL/IRL/\\ Imitation Learning} &  &  & {\Large$\circ$}   &  \\
 
 & & RLHF & {\Large$\circ$} &  & {\Large$\bullet$} &{\Large$\bullet$} \\
 
 & {Scalable Oversight\\ (\S\ref{ssec:scalable_oversight})} & RL$x$F  & {\Large$\circ$} &  & {\Large$\bullet$} & {\Large$\bullet$}   \\
  
 & & IDA & & {\Large$\circ$} & {\Large$\bullet$} & \\
 
 & & RRM & & & {\Large$\bullet$} &  \\
 
 & & Debate & & {\Large$\circ$} & {\Large$\bullet$} & \\
 
 & & CIRL & {\Large$\circ$} & {\Large$\circ$} & {\Large$\bullet$}& {\Large$\circ$}     \\
 
{Learning under\\ Distribution Shift\\ (\S\ref{sec:distribution})}  & {Algorithmic\\ Interventions\\ (\S\ref{sec:alg-inter})} & DRO & {\Large$\bullet$} & & & \\

 &  & IRM/REx  & {\Large$\bullet$} & & & \\
 
 & & CBFT  & {\Large$\bullet$}  & & & \\
 
 & {Data Distribution\\ Interventions (\S\ref{sec:data-inter})}     & Adversarial Training & {\Large$\bullet$} & & {\Large$\circ$} & \\
 
 & & Cooperative Training & {\Large$\bullet$} & & & {\Large$\bullet$} \\
 
Assurance (\S\ref{sec:assurance}) & {Safety\\ Evaluations\\ (\S\ref{sec:safe})} & {Social Concern\\ Evaluations} & {\Large$\circ$} & {\Large$\circ$} & & {\Large$\bullet$}   \\

 & & {Extreme Risk\\ Evaluations} & & {\Large$\circ$} & {\Large$\bullet$} & {\Large$\circ$} \\
 
 & & Red Teaming & {\Large$\bullet$} & & {\Large$\circ$} & {\Large$\bullet$}   \\
 
 & Interpretability (\S\ref{sec:interpretability}) & & & {\Large$\bullet$} & {\Large$\circ$} & \\
 
 & {Human Values\\ Verification (\S\ref{sec:human-values-alignment})} & 
 {Learning/Evaluating\\ Moral Values} & & & {\Large$\circ$} & {\Large$\bullet$}   \\
 
 & & {Game Theory for\\ Cooperative AI} & {\Large$\circ$} & & & {\Large$\bullet$}   \\
 
Governance (\S\ref{sec:governance}) & {Multi-Stakeholder\\ Approach (\S\ref{sec:multi-stake})} & Government & {\Large$\bullet$}   & {\Large$\bullet$} & {\Large$\bullet$}  & {\Large$\bullet$} \\

 & & Industry & {\Large$\bullet$} & {\Large$\bullet$} & {\Large$\bullet$} & {\Large$\bullet$} \\
 
 & & Third Parties & {\Large$\bullet$} & {\Large$\bullet$} & {\Large$\bullet$} & {\Large$\bullet$} \\
 
 & International Governance (\S\ref{sec:intl-gov}) & & {\Large$\bullet$} & {\Large$\bullet$} & {\Large$\bullet$} & {\Large$\bullet$}   \\
 
 & Open-Source Governance (\S\ref{sec:o-s-g}) & & {\Large$\bullet$} & {\Large$\bullet$} &{\Large$\bullet$} & {\Large$\bullet$}   
 
\end{tblr}
}
\end{table}

\subsubsection{Discussion on the Boundaries of Alignment}
\label{sec:ai-safety-beyond}
Following the introduction of alignment inner scope, in this section, we further discuss the relationship between AI safety and alignment. Actually, AI alignment constitutes a significant portion of AI safety concerns. In this section, we will delve into topics that fall right on the boundary of alignment, but well within the broader category of AI safety. Our discussion of broader AI safety concerns will draw from \citet{hendrycks2023overview}.

\paragraph{Human Values in Alignment}\label{sec:values-in-intro}

The inclusion of \emph{Ethicality} in our RICE principles signifies the critical role of human values in alignment. AI systems should be aligned not only with value-neutral human preferences (such as intentions for AI systems to carry out tasks) but also with moral and ethical considerations. These efforts are referred to as \emph{value alignment} \citep{gabriel2020artificial,gabriel2021challenge}.\footnote{Although this term has also been used in other ways, such as to refer to alignment in general \citep{yuan2022situ}.} Considerations of human values are embedded in all parts of alignment -- indeed, alignment research topics dedicated to human values are present in all four sections of our survey. Therefore, to provide a more holistic picture of these research topics, here we give an overview of them before delving into their details in each individual section. 

We classify alignment research on human values into three main themes: (1) \emph{ethical and social values} which aims to teach AI systems right from wrong, (2) \emph{cooperative AI} which aims to specifically foster cooperative behaviors from AI systems, and (3) \emph{addressing social complexities} which provides apparatus for the modeling of multi-agent and social dynamics.

\begin{itemize}
    \item \textbf{Ethical and Social Values}. 
    Human values inherently possess a strong degree of abstraction and uncertainty. \citet{macintyre2013after} even points out that modern society lacks a unified value standard, and the value differences between people of different cultures can be vast. This raises the significant challenge of determining which human values we should align with. Although universally consistent human values may not exist, there are still some values that are reflected across different cultures. In the sections below, we discuss these from the perspectives of \textit{Machine Ethics}, \textit{Fairness}, and \textit{Cross-Cultural Values in Social Psychology}.

    \textit{Machine Ethics}: In contrast to much of alignment research which aligns AI systems with human preferences in general (encompassing both value-laden ones and value-neutral ones), \emph{machine ethics} have specifically focused on instilling appropriate moral values into AI systems \citep{yu2018building, winfield2019machine, tolmeijer2020implementations}. This line of work started early on in the context of symbolic and statistical AI systems \citep{anderson2005towards,arkoudas2005toward,anderson2007status}, and later expanded to include large-scale datasets \citep{hendrycks2020aligning,pan2023machiavelli} and deep learning-based/LLM-based methods \citep{jin2022make}. 
    We cover the formal branch of machine ethics in \S\ref{subsec:formal-ethics-coop}.

    \textit{Fairness}: Although there are controversies ~\citep{verma2018fairness, saxena2019fairness}, the definition of fairness is relatively clear compared to other human values. Specifically, it is the absence of any prejudice or favoritism toward an individual or group based on their inherent or acquired characteristics ~\citep{mehrabi2021survey}. Therefore, there has been extensive research on AI fairness. These methods range from reducing data biases before training ~\citep{d2017conscientious, bellamy2018ai}, to minimizing unfairness introduced during the training process ~\citep{berk2017convex}, and finally addressing instances of unfairness that were not successfully learned during training ~\citep{xu2018fairgan}.

    \textit{Cross-Cultural Values in Social Psychology}: In the field of social psychology, numerous studies have focused on exploring clusters of values that exist among cross-cultural human communities, leading to the development of various cross-cultural values scales. The Allport-Vernon-Lindzey value system ~\citep{allport1955becoming} posited that understanding an individual's philosophical values constitutes a critical foundation for assessing their belief system. They devised a value scale comprising six primary value types, each representing people's preferences and concerns regarding various aspects of life. \citet{messick1968motivational, mcclintock1982social, liebrand1984effect, van1997development} introduced and improved a quantifiable method, namely social value orientation (SVO), to assess an individual's social value inclination. It utilizes quantitative approaches to evaluate how individuals allocate benefits to themselves and others, reflecting their social value orientation, such as altruism, individualism, \textit{etc.} In subsequent work, \citet{murphy2011measuring, murphy2014social} introduced the Slider Measure, which can be used to precisely assess the SVO value as a continuous angle based on the subject's option to some specific questions. \citet{rokeach1973nature} developed a values inventory comprising 36 values, consisting of 18 terminal values representing desired end-states and 18 instrumental values signifying means to achieve those end-states. 
    \citet{schwartz1992universals, schwartz1994there} conducted comprehensive questionnaire surveys in 20 diverse countries known as the Schwartz Value Survey. This study identified ten values that are universally recognized, regardless of culture, language, or location.
    These studies have all laid a solid theoretical foundation for establishing what kind of values AI should be aligned with. However, they are constrained by the historical context of their research and may not maintain strong universality across different times and cultures.

    \item \textbf{Cooperative AI}. Arguably, the most exciting aspect of multi-agent interaction is cooperation, and cooperation failure is the most worrying aspect of multi-agent interaction. As an example of AI cooperation failure, the \emph{2010 Flash Crash} led to a temporary loss of trillions of market value in 2 minutes and was caused in part by interactions between high-frequency algorithmic traders \citep{kirilenko2017flash}. Therefore, there is a need to implement mechanisms ensuring cooperation in agent-like AI systems and the environments they're operating within \citep{dafoe2021cooperative}. The high-level design principles and low-level implementations of such mechanisms fall into the domain of \emph{Cooperative AI}. In addition, Cooperative AI also studies human cooperation through the lens of AI and how AI can help humans achieve cooperation. More precisely, \citet{dafoe2020open} classified Cooperative AI research into four broad topics: \emph{Understanding}, \emph{Communication}, \emph{Commitment}, and \emph{Institutions}. They span various disciplines, from game theory to machine learning to social sciences. 
    This survey has included discussions of cooperative AI, focusing on reinforcement learning in \S\ref{sec:cooperative-ai-training} and game theory in \S\ref{subsec:formal-ethics-coop}.

    \item \textbf{Addressing Social Complexities}. The requirement of ethicality contains in itself a social component. ``What is ethical'' is often defined within a social context; therefore, its implementation in AI systems also needs to account for social complexities. \citet{critch2020ai} provides proposals for many research topics in this vein. One avenue of research focuses on the realistic simulation of social systems, including rule-based \emph{agent-based modeling} \citep{bonabeau2002agent,de2014agent}, deep learning-based simulation \citep{sert2020segregation}, and those incorporating LLMs \citep{park2023generative}. These simulation methods could serve a diverse array of down-stream applications, from impact assessment \citep{calvo2020advancing,fernandes2020adoption} to multi-agent social learning \citep{critch2020ai}. On another front, the fields of \emph{social choice} \citep{sen1986social,arrow2012social} and, relatedly, \emph{computational social choice} \citep{brandt2016handbook} have aimed to produce mathematical and computational solutions for preference aggregation in a diverse population, among other goals. It has been argued that a similar approach when combined with human preference-based alignment methods (\textit{e.g.}, RLHF and most other methods introduced in \S\ref{sec:learning-from-feedback}), could supplement these methods to guarantee a fair representation of everyone's preferences \citep{leike2023proposal, CIPwhitepaper}. There have been early-stage experiments on this proposal \citep{bakker2022fine,kopf2024openassistant}. To complement this approach of learning values from crowds, it has also been argued that embodied values in AI systems should undergo continual progress over the long term as opposed to being permanently locked-in \citep{kenward2021machine}, in order to navigate through emerging challenges, as well as to become future-proof and meet potential \emph{unknown unknowns} in the moral realm. 

\end{itemize}

\paragraph{Malicious Use}

Malicious actors can deliberately use AI to cause harm. Already, deepfakes have been used by criminals to enable scams and blackmail \citep{cao2023deepfake}. As AI systems develop more dangerous capabilities, the threat of misuse looms larger. 

Biological weapons provide one concerning example of how AI could be maliciously used to cause harm. Research has shown that large language models can provide detailed, step-by-step instructions about synthesizing pandemic potential pathogens \citep{soice2023large}. In addition to spreading information about how to create biological weapons, AI could help design new pathogens that are more lethal and transmissible than existing illnesses \citep{sandbrink2023artificial}. Terrorist groups such as Aum Shinrikyo \citep{danzig2012aum} have already attempted to build biological weapons in order to cause widespread destruction, and AI could make it easier for small groups to create biological weapons and start global pandemics. 
Other kinds of malicious use could include using AI to launch cyberattacks against critical infrastructure \citep{mirsky2023threat}, or create autonomous agents that survive and spread outside of human control \citep{rogueai}. As new dangerous capabilities arise in AI systems, thorough evaluations will be required to determine how an AI system could be used to cause harm. 

Malicious use might not be considered a failure of alignment because when an AI system behaves according to the intentions of a malicious user, this system would be aligned with its user but would still pose a serious threat to society. Policies to ensure that AI is aligned with the public interest will be essential to avert this threat.

\paragraph{Collective Action Problems}

Many AI developers are racing to build and deploy powerful AI systems \citep{grant2023ai}. This incentivizes developers to neglect safety and race ahead to deploy their AI systems. Even if one developer wants to be careful and cautious, they might fear that slowing down to evaluate their systems and invest in new safety features thoroughly might allow their competition to outpace them \citep{armstrong2016racing}. This creates a social dilemma where individual AI developers and institutions rationally pursuing their own interests can lead to suboptimal outcomes for everyone. Success in competition between AI systems may be governed by evolutionary dynamics, where the strongest and most self-interested AI systems could be the most likely to survive \citep{hendrycks2023natural}. Preventing these collective action problems from causing societal catastrophes could require intervention by national and international AI policies to ensure that all AI developers uphold common safety standards.

In a broader context, \textit{Malicious Use} can be considered effective alignment between AI systems and individuals with impure intentions, but without alignment with universally held human values. Concurrently, \textit{Collective Action Problems} can be regarded as a consequence of competition, leading developers to neglect the crucial aspect of AI alignment in ensuring model safety. Broadly speaking, the connection between AI alignment and AI safety has progressively become more intertwined, resulting in a gradual blurring of boundaries.

\section{Learning from Feedback}
\label{sec:learning-from-feedback}
Learning from feedback aims to transmit human intentions and values to AI systems. It serves as the foundation for \emph{forward alignment}. In this section, we focus on the dynamic process of learning from feedback, categorizing it into three key elements: (1) \textit{AI System}: refers to systems that require alignment, such as pre-trained LLMs; (2) \textit{Feedback}: provided by an advisor set, which may consist of humans, AI, or humans assisted by AI, \textit{etc.} This serves as the information used to adjust the AI system; (3) \textit{Proxy}: a system developed to model feedback to facilitate more accessible learning. For example, human preference rankings of AI system behaviors serve as feedback, while a reward model acts as the corresponding proxy. From these elements, we identify two pathways by which the AI system learns from feedback: (1) Direct learning from the feedback itself and (2) Indirect learning via proxies that model the feedback.

Following this process, we proceed to \S\ref{ssec:feedback_types} where we discuss different feedback types from the alignment perspective, highlighting various methods of providing information to AI systems. In the following sections, we introduce key concepts that have recently provided insights into developing powerful AI systems \citep{christiano2017deep} and aligning them with human intent \citep{touvron2023llama}. \S\ref{ssec:preference_modeling} focus on Preference Modeling emphasizing its role in creating proxies that help humans provide feedback to complex or hard-to-evaluate AI systems.
Next, we explore Policy Learning in \S\ref{ssec:policy_learning}, focusing on key research directions for developing capable AI systems through feedback.
The discussion then naturally transitions to scalable oversight in \S\ref{ssec:scalable_oversight}, where we reflect on the learning process and objectives from a broader alignment perspective.

\begin{figure}[t]
    \centering
    \includegraphics[width=0.9\linewidth]{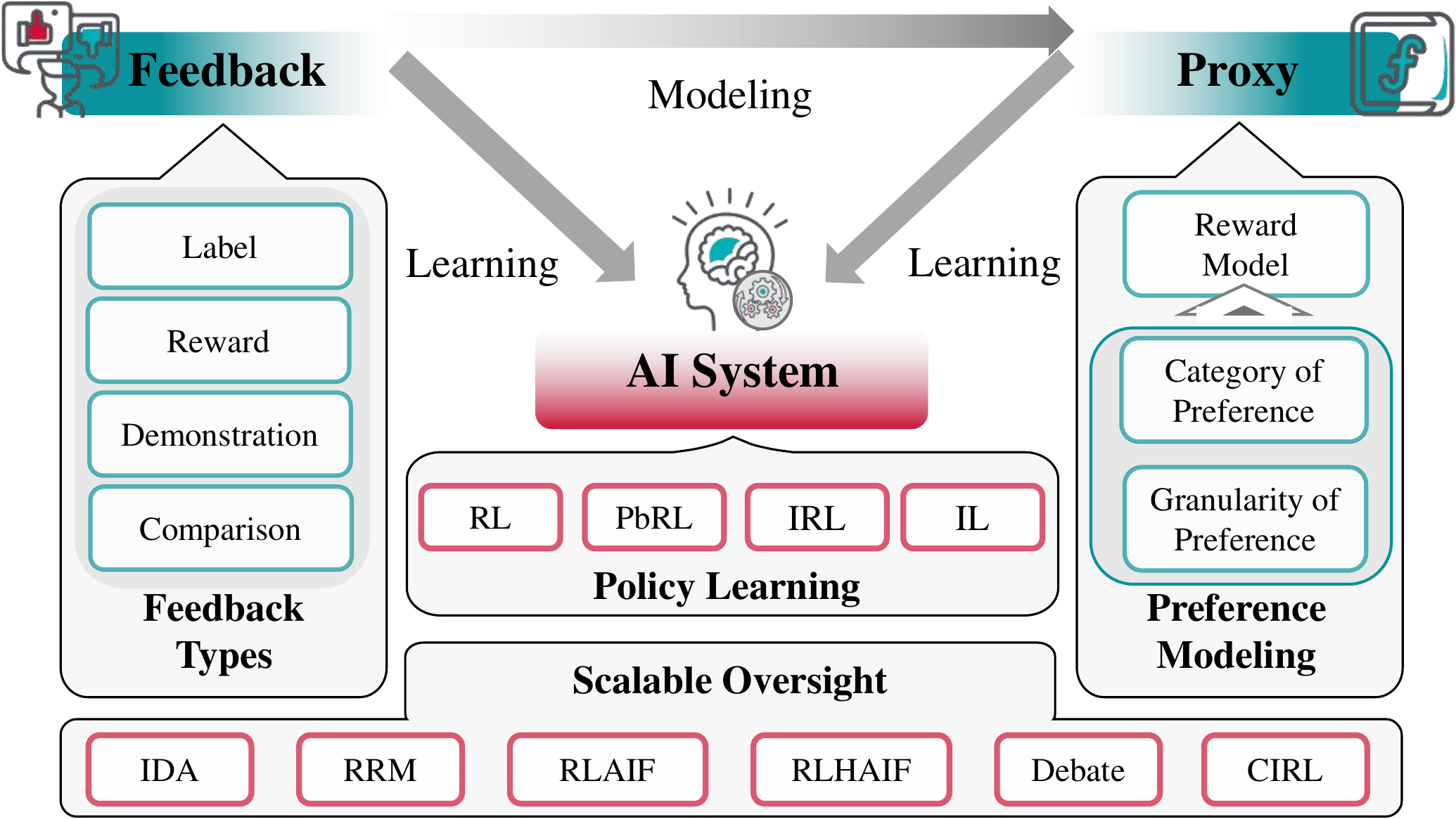}
    \caption{Overview of the learning from the feedback process. Two learning pathways emerge: direct feedback-based learning and proxy-mediated learning (\textit{e.g.}, RLHF). We adopt a \textit{human-centric} perspective, viewing AI systems as \textit{black boxes} and categorizing the forms of feedback presented to AI systems into four types: Label, Reward, Demonstration, and Comparison. 
    % Grounded in fundamental concepts such as Category of Preference and Granularity of Preference, we introduce the Reward Model, a specific instantiation of a Proxy. In the context of AI Systems, we discuss four distinct domains: Reinforcement Learning (RL), Imitation Learning (IL), Inverse Reinforcement Learning (IRL), and Preference-based Reinforcement Learning (PbRL) as a background. Scalable Oversight, a research theme that seeks to ensure AI systems, even those surpassing human expertise, remain aligned with human intent, is explored through the introduction of four promising directions: Iterated Distillation and Amplification (IDA), Recursive Reward Modeling (RRM), Debate, and Cooperative Inverse Reinforcement Learning (CIRL). Additionally, building upon RLHF, we propose RLxF, encompassing Reinforcement Learning from AI Feedback (RLAIF) and Reinforcement Learning from Human and AI Feedback (RLHAIF), as an extension of RLHF and a fundamental framework for Scalable Oversight.
    }
    \label{fig:feedback_flowchart}
\end{figure}

\subsection{Feedback Types}
\label{ssec:feedback_types}
Feedback is a crucial link between AI behaviors to human intentions \citep{stumpf2007toward, stumpf2009interacting} leveraged by AI systems to refine their objectives and more closely align with human values \citep{glaese2022improving}, this includes two primary meanings: (1) During system construction, external sources provide feedback on the AI system's output, guiding refinements to the system's architecture or its internal information \citep{zhou2021machine}. (2) After the system deployment, it will continuously adapt to changes in external environmental data, maintaining the architecture or fundamental strategy of the system unchanged, with methods such as adaptive control \citep{aastrom2008adaptive, aastrom2021feedback} and in-context learning \citep{dong2022survey}.
% After system deployment, the system dynamically tunes its behaviors in response to external data. However, the architecture or fundamental strategy of the system remains unchanged \citep{aastrom2008adaptive, aastrom2021feedback, dong2022survey}.
For a precise and detailed discussion of the feedback types with precision and detail, it is essential to initially define \textit{feedback} within the scope of alignment.

% \begin{tcolorbox}[colback=gray!20, colframe=gray!50, sharp corners, center title]
% \centering
% \textit{\large Feedback is information given to the AI system to align it with human intent.}
% \end{tcolorbox}

\begin{quotebox}
\large Feedback is information given to the AI system to align it with human intent.
\end{quotebox}

\label{sec:category_of_feedback}
Considering diverse AI systems in alignment research, we embrace an \textit{human-centric} approach. Instead of delving deep into the complex system mechanics, we propose a taxonomy to classify feedback according to its \textit{direct presentation forms} to the system. This section introduces four types of feedback employed to align AI systems commonly: label, reward, demonstration, and comparison. It is worth noting that beyond explicit feedback, there are approaches that exploit the information embedded in vast amounts of unlabeled data through unsupervised pre-training \citep{parisi2022unsurprising} and semi-supervised learning \citep{xu2018semantic}, showing considerable promise in enhancing model capabilities \citep{zhou2024lima}.

\paragraph{Label}
\label{par:label}

Label feedback refers to one or more meaningful information tags attached to the original data item \citep{hastie2009overview}, which stands as the most direct form, offering explicit guidance and delineating expected outputs for AI systems. 
This type of feedback prompts AI systems to learn from input-output pairings provided by expert advisors. For example, in supervised learning, an AI model is trained using a dataset of labeled input-output pairs, denoted by $D = \left\{ (x_i, y_i) \right\}_{i = 1}^N$. Here, $y_i$ represents the true labels corresponding to the input data $x_i$, and $N$ signifies the total number of samples in the dataset. The essence of the learning process revolves around minimizing a loss function $\mathcal{L}$ (\textit{e.g.}, MSE), which measures the disparity between the predictions of the model, $f(x; \theta)$, and the ground truth labels $y$, based on the model parameters, $\theta$.

The advantage of label feedback is its unambiguous nature and simplicity in interpretation.
However, due to the inability of label feedback to fully encapsulate the underlying logic of this choice, employing such feedback in model training can result in target variable bias \citep{guerdan2023ground}. And, its utility might diminish when tackling complex tasks beyond mere classification or regression \citep{lake2017building, marcus2018deep}. For example, in tasks like optimizing algorithms \citep{fawzi2022discovering, mankowitz2023faster}, video game playing \citep{baker2022video}, and multi-modal generation \citep{openai2023gpt4V}, it is not only impractical to provide explicit instructions for every conceivable situation but also insufficient to solely rely on label feedback to build systems that surpass human capabilities.

\paragraph{Reward}
\label{par:reward}
A reward is an absolute evaluation of a single output from an AI system, represented as a scalar score \citep{silver2021reward} or a vector of scores \citep{wu2024fine}, each independent of other outputs.

Feedback based on rewards provides a quantified evaluation of the AI system, allowing for direct guidance in behavior adjustments.
This type of feedback typically originates from pre-designed, rule-based functions or procedures. 
For example, in the MuJoCo simulation, environments from OpenAI Gym \citep{brockman2016openai}, the task is to guide the agent moving forward effectively.
To this end, an effective rule-based reward function can be formulated as a composite of several key components: maintaining a healthy status, encouraging forward movement, minimizing control exertion, and regulating contact intensity.

The advantage of reward feedback is that the designer does not need to delineate the optimal behavior while allowing the AI system to explore to find the optimal policy \citep{kaelbling1996reinforcement, mnih2015human, silver2016mastering, silver2017mastering}. 
However, crafting flawless rules to determine scores for functions that evaluate the output of AI systems \citep{everitt2017reinforcement, specification2020victoria, pan2021effects} or directly assigning calibrated and consistent scores to each AI system output \citep{isbell2001social, thomaz2008teachable, christiano2017deep, casper2023open} is challenging for human. 
This is due to the inherent complexity of the tasks, where it's impractical to account for every nuance. 
Additionally, flawed or incomplete reward functions can lead to dangerous behaviors misaligned with the intention of the designer, such as negative side effects and reward hacking \citep{hadfield2017inverse, skalse2022defining}. Thus, merely from the alignment perspective, perhaps the most important limitation of feedback based on rewards is that it may be difficult to rule out manipulation \citep{shevlane2023model}, which amounts to reward tampering and reward gaming \citep{leike2018scalable, everitt2021reward, skalse2022defining} in this context. CIRL in \S\ref{sec:cirl}, provides insights into this particular issue.

\paragraph{Demonstration}
\label{par:demonstration} 
Demonstration feedback is the behavioral data recorded from expert advisors while achieving a specific objective \citep{hussein2017imitation}. Demonstrations can take on various forms, including videos \citep{ shaw2023videodex}, wearable device demonstrations  \citep{edmonds2017feeling, wang2023learning}, collaborative demonstrations \citep{bozorgi2023beyond}, and teleoperation \citep{zhang2018deep}. 
If the dynamics of the demonstrator and the AI learner are identical, the demonstration can directly constitute a trajectory made up of state-action pairs \citep{zhang2023discriminator}.
These state-action pairs can also be partially observable \citep{torabi2018behavioral, brown2019extrapolating}.
For example, a video can be recorded of a human expert performing a robotic manipulation task, such as grasping an object with a robotic hand. One can subsequently annotate each video frame with the associated robot state \citep{shaw2023videodex} and action \citep{baker2022video} for each frame. This results in a dataset of state-action pairs from the human demonstration that can be used to train the agent's policy to imitate the expert behavior.

This feedback leverages the expertise and experience of advisors directly, obviating the need for formalized knowledge representations \citep{fang2019survey, dasari2023learning}. 
However, it may falter when confronting tasks that exceed the advisors' realm of expertise \citep{hussein2017imitation}. Additionally, it faces challenges stemming from the noise \citep{sasaki2020behavioral} and suboptimality \citep{attia2018global} in real-world advisor demonstrations \citep{yang2021trail}.
Furthermore, human advisors, prone to imprecision and errors, can introduce inconsistencies \citep{zhu2019ingredients, iii2022fewshot}. 
Meanwhile, there might be a need for a vast amount \citep{sasaki2020behavioral} and diverse set \citep{beliaev2022imitation} of demonstrations within acceptable costs, which results in significant difficulty in learning reliable behaviors.

\paragraph{Comparison}
\label{par:comparison}
Comparison feedback is a relative evaluation that ranks a set of outputs from an AI system and guides the system toward more informed decisions \citep{wirth2017survey}. For example, this feedback form is manifested in Preference Learning \citep{furnkranz2010preference}, where the AI system discerns the preferences of advisors by comparing multiple examples. 

The fundamental advantage of comparison feedback is humans' capacity to quickly handle tasks and objectives that are hard for precise evaluation \citep{hullermeier2008label, christiano2017deep, ouyang2022training}. Nevertheless, beyond common factors like noise in the feedback and unmodeled contextual elements that hinder the model's convergence to true objectives, the absolute differences between different items become obscured. Consequently, the performance of a strategy tends to optimize towards a median target rather than an average target. \citet{casper2023open} illustrates this with an example of action $A$, always yielding a value of 1, and action $B$, which yields 10 in 40\% of cases and 0 in 60\%. When assessed based on comparison feedback, action $A$ is deemed superior to $B$, even though $B$ possesses a higher expected return. It also has the inherent limitation of potentially requiring a substantial amount of comparative data \citep{furnkranz2003pairwise, gao2023scaling}, although some studies indicate that the necessary quantity may be relatively smaller \citep{christiano2017deep}. Preference modeling is an example of using this type of feedback, as detailed in \S\ref{par:reward_model}. 

\paragraph{Discussion} 

All types of feedback can be provided to AI systems interactively and online. This process engenders synchronous iterations between providing feedback and AI system updates, underscoring rapid, focused, and incremental model modifications \citep{amershi2014power, holzinger2016interactive}. For instance, demonstration feedback can manifest in the form of online corrections \citep{bajcsy2018learning, li2021learning, losey2022physical}.

Interactively providing feedback emphasizes the role of interactivity in the learning process, allowing AI systems to evolve based on interactive experiences. In active learning, robots actively engage in data discovery and acquisition, thereby facilitating learning throughout the process of online deployment \citep{taylor2021active}. And in interactive learning, feedback manifests in the form of guided corrections that online rectify missteps in the behavior of the AI system \citep{fails2003interactive, amershi2014power, saunders2022self}. For example, the interactive image segmentation emphasizes simple \citep{zhang2020interactive}, intuitive \citep{rother2004grabcut, xu2016deep}, and real-time \citep{liu2022pseudoclick} interactions.

One of the essential advantages of interactively providing feedback is its ability to fine-tune AI systems in real-time, allowing users to interactively explore the model's space \citep{amershi2014power} to ensure quick and subtle alignment with the directives of advisors \citep{shin2020autoprompt, wei2022chain, zou2023segment}. Moreover, this process lessens the dependence on specialist knowledge and promotes better interpretability \citep{berg2019ilastik}. However, it may be limited by the interactivity to choose time-intensive algorithms \citep{fails2003interactive, holzinger2016interactive}.

Furthermore, considering more powerful AI systems are emerging, more universal interaction interfaces are also coming up, such as language \citep{lynch2023interactive, openai2023gpt4} and vision \citep{rt2_deepmind_2023}, which bridge the communication gap between humans and AI systems. In robotics, a series of studies have linked human-provided language with rewards obtained by agents. This association enables the conveyance of nuanced human intentions through language, thereby guiding the generation of scalar feedback signals during the training \citep{fu2018from, goyal2019using, sumers2021learning, zhou2021inverse, lin2022inferring, yu2023language} and planning \citep{sharma2022correcting} process.
In the realm of LLMs, in-context learning \citep{dong2022survey} serves as a means to supplement information via language during deployment, thereby enhancing the alignment of LLMs with human intent.

These various modes of feedback share a common trait -- that they can all be seen as attempts by humans to convey a hidden reward function. \citet{jeon2020reward} proposes and formalizes this position and unifies a wide array of feedback types by defining a parameterized reward function $\Psi\left(\cdot;\bm{\theta}\right)$ that underlies the feedback process. This allows the AI system to, for example, perform Bayesian inference on $\bm{\theta}$, regardless of the feedback type.

Recently, techniques based on IL and RL have successfully constructed AI systems with significant capabilities \citep{baker2022video, openai2023gpt4V}. However, this success naturally leads to two questions: 
\begin{itemize}[left=0.3cm]
\item How can we define reward functions for more complex behaviors (\textit{e.g.}, various sub-tasks in interactive dialogue), aiming to guide the learning process of AI systems? 
\item How can we express human values such that powerful AI systems align better with humans, ensuring the system's \textit{controllability} and \textit{ethicality}?
\end{itemize}

Endeavors incorporating preference modeling into policy learning have shown progress. The most notable achievements in this domain have been observed in constructing powerful LLMs \citep{openai2023gpt4, touvron2023llama, anthropiceval}. 
Additionally, a series of policy learning studies have reported performance improvements. For instance, combining preference modeling with Inverse Reinforcement Learning (IRL) \citep{brown2019extrapolating, brown2020safe} and offline RL \citep{shin2023benchmarks}, fine-tuning reward functions \citep{iii2022fewshot}, modeling non-Markovian rewards \citep{kim2023preference}, and aiding in the construction of intricate reward functions \citep{bukharin2023deep}.
Therefore, we consider preference modeling (as shown in \S\ref{ssec:preference_modeling}) and policy learning (as shown in \S\ref{ssec:policy_learning}) as fundamental contexts for understanding the challenges faced in alignment and potential solutions. Next, we provide a brief overview of these specific techniques related to alignment.

\subsection{Preference Modeling}
\label{ssec:preference_modeling}
In many complex tasks, such as dialogues \citep{ouyang2022training}, constructing precise rule-based rewards presents a challenge \citep{bender2021dangers}. At the same time, methods based on demonstration might require a substantial investment of expert human resources, resulting in high costs. Currently, preference modeling based on comparison feedback \citep{akrour2011preference} has emerged as a very promising method \citep{ouyang2022training, openai2023gpt4, touvron2023llama} to assist in fine-tuning powerful AI systems \citep{amodei2016concrete}.

Typically, it is necessary to iteratively explore the system dynamics while acquiring expert preference data to gain more knowledge about the optimization objectives. 
This process is known as \textit{Preference Elicitation} \citep{wirth2013preference, wirth2017survey, christiano2017deep, cabi2019scaling}, which is crucial for obtaining rich, valuable feedback related to AI system outputs, thus guiding the alignment process \citep{iii2022fewshot}. Within \textit{Preference Elicitation}, two core decisions that need to be determined are the \textit{Granularity of Preference} and the \textit{Category of Preference}. This paper introduces these within sequential decision-making problems, but the insights derived apply to a broad array of AI systems \citep{amodei2016concrete, christiano2018supervising, leike2018scalable}.

\paragraph{Granularity of Preference}

% The granularity of preference \citep{wirth2017survey} is mainly three types: \textit{Action}, \textit{State}, and \textit{Trajectory} (as shown in Table \ref{tab:comparison_preference_granularities}). 
Preference \citep{wirth2017survey} can primarily be categorized into three types by granularity: \textit{Action}, \textit{State}, and \textit{Trajectory} (as shown in Table \ref{tab:comparison_preference_granularities}).

The \textit{Action} preference focuses on comparing actions within a particular state, specifying the preferred action under specific conditions. When translated into trajectory preferences, it may impose challenges such as evaluators' expertise needs and potential information loss. The \textit{State} preference deals with comparing states. It encapsulates preference relations among states but requires assumptions about state reachability and independence when translating to trajectory preferences. The \textit{Trajectory} preference considers whole state-action sequences, offering more comprehensive strategic information. It inherently assesses long-term utility and depends less on expert judgment.

\begin{table}[t]
\centering
\caption{A comparison of the three types of preference granularity in the context of sequential decision-making. Each type is defined according to its characteristics and the way it compares different elements of the learning process. The notation $i_1 \succ i_2$ denotes that $i_1$ is strictly preferred over $i_2$.}
\label{tab:comparison_preference_granularities}
\resizebox{\textwidth}{!}{
    \begin{tabularx}{\linewidth}{lX}
    \toprule
    \textbf{Preference Granularity} & \textbf{Definition} \\
    \midrule
    \textbf{Action} & Compares two actions $a_1$ and $a_2$ within the same state $s$, denoted as $a_1 \succ_s a_2$. \\
    \textbf{State} & Compares two states $s_1$ and $s_2$, denoted as $s_1 \succ s_2$. \\
    \textbf{Trajectory} & Compares two complete state-action sequence trajectories, denoted as $\tau_1 \succ \tau_2$. Each trajectory $\tau$ consists of state-action pairs at time $t$, expressed as $\tau = \{s_0, a_0, s_1, a_1, ..., s_{T-1}, a_{T-1}, s_T\}$. \\
    \bottomrule
    \end{tabularx}
}
\end{table}

\citet{christiano2017deep} demonstrates, using ablation studies, that in the settings that they studied, longer trajectory segments yield more informative comparisons on a per-segment basis. Such segments are also more consistently evaluated by humans in MuJoCo tasks.

\paragraph{Category of Preference}

Diverse objectives exist within preference modeling. Based on their targets, preferences can be categorized into object preference and label preference \citep{furnkranz2010preference}. Specifically, object preference operates on a set of labels for each instance, whereas label preference acts on a set of objects themselves. One can further classify them differently based on the form of preferences.

\begin{itemize}[left=0.3cm]

\item {\textbf{Absolute Preferences}}. Absolute preferences independently articulate each item's degree of preference.

\begin{itemize}[left=0.2cm]
\item {\textbf{Binary}}. Classifying items as liked or disliked offers a simplistic and straightforward model of user preference \citep{tsoumakas2007multi, cheng2010graded}.

\item {\textbf{Gradual}}. This can be further distinguished between numeric and ordinal preferences. Numeric preferences employ absolute numerical values, such that each item receives a numerical score, which reflects the extent of preference \citep{cheng2010label}. On the other hand, ordinal preferences entail a graded assessment of a fixed set of items as either preferred, less preferred, or intermediary, \textit{etc.}, enabling the depiction of user preferences without including specific numerical measurements \citep{cheng2010graded}.
\end{itemize}

\item {\textbf{Relative Preferences}}.
Relative preferences define the preference relation between items.

\begin{itemize}[left=0.2cm]

    \item {\textbf{Total Order}}. This form establishes a comprehensive preference relation covering all item pairs, asserting an absolute ordering of preferences ranging from the most preferred to the least \citep{hullermeier2008label}.
    
    \item {\textbf{Partial Order}}. Because users may not exhibit a distinct preference between two items in some instances \citep{cheng2010predicting}, this allows for incomparable item pairs.
    \end{itemize}
\end{itemize}

\paragraph{Reward Model}
\label{par:reward_model}
Reward modeling transfers comparison feedback \citep{furnkranz2010preference, wirth2017survey} to the scalar reward form, facilitating policy learning \citep{christiano2017deep, cabi2019scaling, touvron2023llama}. Given pairs of actions $(y_1, y_2)$ performed by the RL agent in the same state. The preference is denoted as $y_w\succ y_l \mid x$, where $y_w$, $y_l$ represents the preferred and less preferred action respectively among $(y_1, y_2)$. We assume these preferences emerge from a latent reward model $r^*(x, y)$, which we lack direct access to. Several methods exist to model such preferences, \textit{e.g.}, the Bradly-Terry Model \citep{bradley1952rank}, Palckett-Luce ranking model \citep{plackett1975analysis}, \textit{etc.} Under the BT model, the distribution of human preference, denoted as $p^*$, can be 
% expressed 
formalized
as, 
\begin{align*}
    p^*\left(y_1\succ y_2 \mid x\right)=\frac{\exp\big(r^*(x, y_1)\big)}{\exp\big(r^*(x, y_1)\big) + \exp\big(r^*(x, y_2)\big)}=\sigma \big(r^*(x, y_1) - r^*(x, y_2)\big).
\end{align*}
where $\sigma(x) = 1/\left(1 + \exp(-x)\right)$ is the logistic sigmoid function.
Subsequently, we use the derived preference rankings to train the parameterized reward model, optimizing its parameters through maximum likelihood.
\begin{align*}
\mathcal{L}_{\mathrm{R}}\left(\bm{\theta}\right)=-\mathbb{E}_{(x,y_w,y_l)\sim \mathcal D}\Big[\log\Big(\sigma\big(r_{\bm{\theta}}\left(x,y_w\right)-r_{\bm{\theta}}\left(x,y_l\right)\big)\Big)\Big]
\end{align*}
In this negative log-likelihood loss, the problem is a binary classification task, where $\mathcal D$ signifies the static dataset $\left \{ x^{(i)}, y_{w}^{(i)}, y_{l}^{(i)}\right \}_{i=1}^{N}  $ sampled from $p^*$ (\textit{i.e.}, human-labeled comparisons).

Reward models enable human users to impart specific preferences to these systems via evaluations, thereby circumventing the complex task of defining objectives explicitly. 
Initially, the studies by \citet{knox2012learning, knox2013learning} distinctively treat human reward as separate from the traditional rewards of MDP and conduct a reward modeling process around it.
Transitioning from these simpler cases, \citet{christiano2017deep} propose that utilizing supervised learning to construct a distinct reward model asynchronously can substantially diminish interaction complexity by approximately three orders of magnitude. 
The study conducted by \citet{ibarz2018reward} integrates expert demonstrations with human preferences, such that the policy initially mimics expert demonstrations and then sequentially collects human trajectory annotations, trains the reward model, and updates the policy. This research also provides practical insights for precluding the overfitting of the reward model and the occurrence of \textit{reward hacking} -- a scenario where escalating rewards do not translate to improved performance, especially when the policy is excessively trained. 
Additionally, a random policy might rarely exhibit meaningful behavior for tasks that surpass the complexity of Atari \citep{palan2019learning, jeon2020reward}. This implies that for effective annotation, the policy itself must possess certain capabilities to perform improved behavior. 
Offline settings also benefited from the reward model. \citet{cabi2019scaling} proposes reward sketching to efficiently learn a reward model that leverages humans' episodic judgments for automated reward annotation of historical data, enabling large-scale batch RL.
\citet{qiu2024rethinking} provides an empirically-grounded theory of reward generalization in RMs, based on which a new type of RM based on tree-structured preferences is proposed and experimentally validated.

Importantly, the reward model provides an essential tool for aligning powerful LLMs. \citet{stiennon2020learning} employs reward models grounded in human preferences for text summarization tasks, resulting in significant policy enhancements. This work also delves into the issues of distribution shift and reward model generalization, revealing that the effectiveness of the reward model correlates with data scale and parameter size. Building upon this work, InstructGPT \citep{ouyang2022training} extends the reward model paradigm to broader dialogue task reward modeling and introduces a preference-optimizing loss function for multiple responses to mitigate overfitting. Furthermore, this research reveals that the preferences derived from the reward model can be generalized across different groups.

\subsection{Policy Learning}
\label{ssec:policy_learning}

Policy learning aims to learn the mapping from perceived states to actions taken when in those states \citep{sutton2018reinforcement} to optimize a model's performance in specific tasks. Numerous alignment-related challenges manifest within policy learning (as shown in \S\ref{sec:challenges-of-alignment}). Consequently, policy learning provides a crucial backdrop for alignment, and its techniques can further advance alignment objectives \citep{amodei2016concrete, christiano2018supervising, ibarz2018reward}. This section discusses various domains within policy learning and then introduces RLHF, a powerful technique for policy learning \citep{openai2023gpt4, touvron2023llama}.

\subsubsection{Background}

We introduce some general areas of policy learning here to give readers a general background.

\paragraph{Reinforcement Learning (RL)}
RL enables agents to learn optimal policies by trial and error via interacting with the environment \citep{sutton2018reinforcement}. This paradigm has achieved great success in tackling complex tasks \citep{li2017deep, yu2021reinforcement, fawzi2022discovering, baker2022video, afsar2022reinforcement, mankowitz2023faster, openai2023gpt4V}, demonstrating its potential for decision-making and control in complex state spaces. 
The goal of RL is to learn a policy $\pi$ which executes actions $a$ in states $s$ to maximize the expected cumulative reward under environment transition dynamics $P$ and the initial state distribution $\rho_0$:
\begin{gather*}
\pi^* = \underset{\pi}{\mathop{\operatorname{argmax}}} \left\{\mathbb{E}_{s_0,a_0,\dots}\left[\sum_{t=0}^\infty\gamma^tr(s_t)\right] \right\}, ~~\text{where} \  s_0\sim\rho_0(\cdot),\   a_t\sim\pi\left(\cdot|s_t\right),\  s_{t+1} \sim P\left(\cdot|s_t,a_t\right).
\end{gather*}

Even though RL still faces challenges like sample efficiency and stability \citep{bucsoniu2018reinforcement}. Proximal policy optimization (PPO) \citep{schulman2017proximal} is an influential algorithm in the RL community, serving as the key algorithm for RLHF \citep{ouyang2022training}. The key idea of PPO is to limit the policy update to prevent significant deviations from the original policy by introducing a proximity objective. \citet{sikchi2023dual} unifies several RL and 
Imitation Learning (IL) algorithms under the framework of dual RL through the lens of Lagrangian duality.

\paragraph{Preference-based Reinforcement Learning (PbRL)}
PbRL \citep{wirth2017survey} seeks to facilitate training RL agents using preference feedback instead of explicit reward signals \citep{christiano2017deep,sadigh2017active}.\footnote{Notably, \citet{sadigh2017active} explicitly maintains a probabilistic belief over the true reward function during learning, and actively constructs queries to the human to reduce uncertainty maximally. Both traits are in a similar spirit to \emph{cooperative inverse reinforcement learning} (CIRL), and later work also continues this theme \citep{reddy2020learning}. See \S\ref{sec:cirl} for more.}
PbRL integrates the advantages of preference learning and RL, broadening the application range of RL and mitigating the difficulties associated with reward function formulation, and has been efficaciously deployed in a variety of tasks such as robotic instruction \citep{kupcsik2013data}, path planning \citep{jain2013learning}, and manipulation \citep{shevlane2023model}. In PbRL, the emphasis predominantly lies on trajectory preferences (\textit{i.e.}, comparisons of state-action sequences segment) \citep{wirth2017survey}. Such trajectory preferences encapsulate a human evaluation of various behavioral outcomes rather than single states, rendering PbRL more suitable for non-expert users \citep{christiano2017deep, shin2023benchmarks, kim2023preference}. A general example of PbRL is the \textit{weighted pairwise disagreement loss} \citep{duchi2010consistency} balancing multiple potentially conflicting preferences to identify a singular optimal policy:
\begin{align*}
\mathcal{L}(\pi,\zeta) = \sum_{i=1}^N\alpha_iL(\pi,\zeta_i),
\end{align*}
where $\mathcal{L}(\pi,\zeta)$ is the aggregated loss for policy $\pi$ over all preferences $\zeta$, $\alpha_i$ is the weight of the $i$th preference, and $L(\pi,\zeta_i)$ is the loss associated with the policy $\pi$ in relation to the specific preference $\zeta_i$.

Compared to exact numerical rewards, preference feedback has several benefits \citep{wirth2017survey}, such as (1) circumventing arbitrary reward design, reward shaping, reward engineering, or predefined objective trade-offs, (2) diminishing reliance on expert knowledge, and (3) decoupling training loop with human by modeling preferences \citep{akrour2012april}. However, PbRL also faces challenges, including credit assignment problems due to temporal delays, practical exploration of preference space \citep{wirth2017survey}, the potential need for massive data \citep{ouyang2022training}, and the inability to use the learned preference model for retraining \citep{mckinney2022fragility}.

\paragraph{Imitation Learning (IL)}\label{par:imitation-learning}

IL \citep{schaal1999imitation,syed2008apprenticeship}, also referred to as learning from demonstration or apprenticeship learning, focuses on emulating human behaviors within specific tasks. The agent learns a mapping between observations and actions and refines its policy by observing demonstrations in a collection of teacher demonstration data $\mathcal D$ \citep{bakker1996robot,hussein2017imitation}. This process obviates the need for environmental reward signals \citep{hussein2017imitation}. Broad IL \citep{cotra2018iterated} aims to replicate human desires and intentions, effectively creating replicas of human decision-making processes. This concept is central to technologies such as Iterated Distillation and Amplification (IDA, as shown in \S\ref{ssec:iterated_distillation_and_amplification}) \citep{christiano2018supervising}. On the other hand, Narrow IL aims to replicate specific human behaviors within given tasks. Behavioral cloning (BC) \citep{bain1995framework,ross2011reduction,osa2018algorithmic} is a simple \citep{pomerleau1991efficient, ravichandar2020recent} strategy that learns directly from demonstrations using supervised learning \citep{schaal1996learning}.
BC method specifically seeks to optimize the policy parameters, $\bm{\phi}$, with the objective of aligning the policy $\pi_{\bm{\phi}}(a|s)$ closely with the expert policy $\pi_E(a|s)$. This alignment is achieved through the minimization of the negative log-likelihood, as delineated in the following \citep{lynch2020learning}:
\begin{align*}
\mathcal{L}_{\text{BC}}(\bm{\phi}) = -\mathbb{E}_{(s,a)\sim \pi_E}\big[\log\pi_{\bm{\phi}}\left(a|s\right)\big].
\end{align*}
Here, the expectation is computed over state-action pairs sampled from the expert policy, $\pi_E$.
However, it faces the Out-of-Distribution (OOD) problem, arising from the difference between the training and testing distributions \citep{ross2011reduction, ho2016generative, reddy2019sqil, zhou2022domain}. Adversarial imitation learning methods \citep{ho2016generative, fu2018learning, lee2019efficient, ghasemipour2020divergence} have demonstrated an ability to enhance the robustness of policies against distribution shifts. However, these methods learn non-stationary rewards, which cannot be used to train new policies \citep{ni2021f}.

\paragraph{Inverse Reinforcement Learning (IRL)} Unlike the paradigm of IL, IRL \citep{adams2022a} focuses on deriving a reward function from observed behavior \citep{ng2000algorithms, arora2021survey}. Standard IRL methods include the feature matching methods \citep{abbeel2004apprenticeship}, which assumes optimal expert behavior or decision processes, as well as the maximum entropy methods \citep{ziebart2008maximum} and the Bayesian methods \citep{ramachandran2007bayesian}, both of which do not require optimal behavior. IRL guarantees robustness to changes in the state distribution but at the cost of increased computational complexity due to the extra RL step \citep{ho2016generative, fu2018variational}. This interaction, meanwhile, introduces inherent RL challenges, \textit{e.g.}, sample efficiency \citep{yu2018towards} and potential dangers in environment interaction \citep{garcia2015comprehensive}. Additionally, identifying the reward function remains a challenge \citep{kim2021reward}.

\subsubsection{Reinforcement Learning from Human Feedback (RLHF)}

RLHF expands upon PbRL within the domain of DRL \citep{christiano2017deep}, aiming to more closely align complex AI systems with human preferences \citep{openai2023gpt4V}. Its principal advantage is that it capitalizes on humans being better at judging appropriate behavior than giving demonstrations or manually setting rewards. This approach has gained significant traction, particularly in fine-tuning LLMs \citep{ouyang2022training, openai2023gpt4, touvron2023llama}. Nonetheless, RLHF encounters obstacles \citep{casper2023open}, including data quality concerns, the risk of reward misgeneralization, reward hacking, and complications in policy optimization. Specifically, RLHF can also be viewed as a Recursive Reward Modeling (RRM) process (as shown in \S\ref{par:rrm}) without deep recursive modeling \citep{leike2018scalable}. Here, we provide a brief review of the RLHF methodology.

The genesis of RLHF can be traced back to \citet{knox2008tamer,knox2012reinforcement}, subsequently broadening its reach to domains such as social robots \citep{knox2013training} and human-AI cooperative learning \citep{griffith2013policy}. Besides focusing on the association between feedback and policy, \citet{loftin2016learning} models the connection between feedback and the trainer strategy. \citet{christiano2017deep} extended RLHF to simulated robotic tasks, demonstrating its potential effectiveness.

It's worth noting that one of the significant applications of RLHF has been in the field of LLMs. Some work found that LLMs trained with RLHF \citep{ouyang2022training, korbak2023pretraining, thoughts_on_the_impact_of_rlhf_search} are more creative and human alignment compared to models trained via supervised or self-supervised learning approaches \citep{kenton2019bert, brown2020language}. The importance of RLHF is not merely limited to allowing LLMs to follow human directives \citep{ouyang2022training}. It helps LLMs better align by giving them important qualities like being helpful, harmless, and honest \citep{bai2022training}. 
Due to these improvements, many works use RLHF for aligning LLMs \citep{ziegler2019fine, stiennon2020learning, bai2022training, glaese2022improving, openai2023gpt4, touvron2023llama}.
Additionally, \citet{dai2024safe} integrates the Safe RL \citep{garcia2015comprehensive} framework with the RLHF, addressing the inherent tension between aligning helpfulness and harmfulness \citep{bai2022training}.
Future efforts can be focused on reducing dependence on human annotation \citep{wang2023self, sun2024principle} and improving the efficacy of the reward model by leveraging iterative RLHF methods (\textit{i.e.}, integrating it with debate frameworks \citep{irving2018ai}), \textit{etc.} \citet{qiu2024rethinking} has also built a formal framework of the RLHF process portraying it as an autoencoding process over text distributions, and enables analysis of convergence properties in RLHF.

We review the RLHF pipeline from the \citet{ziegler2019fine, ouyang2022training, rafailov2024direct} to give a general framework. It usually consists of three stages:
\begin{itemize}[left=0.3cm]   
    \item \textbf{Supervised Fine-tuning (SFT)}. RLHF usually starts with a pre-trained language model, then fine-tuned using supervised learning -- specifically, maximum likelihood estimation -- on a high-quality human instruction dataset tailored for downstream tasks to obtain a model $\pi^{\mathrm{SFT}}$. Examples of these tasks include dialogue handling, instruction following, and summarization (Some open-source datasets include Alpaca Data (52k instruction-following data) \citep{taori2023stanford}, Vicuna (70K user-shared ChatGPT conversations) \citep{chiang2023vicuna}, \textit{etc.}). This stage can also be carried out at any other stage.
    
    \item \textbf{Collecting Comparison Data and Reward Modeling}. This phase includes collecting comparison data, which is subsequently used to train a reward model. The SFT model is given prompts denoted as $x$ to generate pairs of responses $(y_1, y_2)$ sampled from $\pi^{\mathrm{SFT}}(y \mid x)$. These pairs are subsequently shown to human annotators, who indicate a preference for one of the responses. Then as discussed in \S\ref{par:reward_model}, comparison data is used to construct the reward model $r_{\bm{\theta}}$.

    \item \textbf{Policy Optimization via Reinforcement Learning}. The final step is optimizing LLM as a policy $\pi$ through RL, guided by the reward model $r_{\bm{\theta}}$. 
    The process of LLMs generating responses from prompts is modeled as a bandit environment~\cite{ouyang2022training}, where a reward is obtained from reward model $r_{\bm{\theta}}$ at the end of each response.
    The primary objective of RL is to adjust the parameters ${\bm{\phi}}$ of the LLMs such that the expected reward on training prompt dataset $\mathcal{D}_\text{RL}$ is maximized:
    \begin{equation*}
        \mathop{\arg\max}\limits_{\pi_{\bm{\phi}}} ~ \mathbb{E}_{x\sim\mathcal{D}_\text{RL}, y\sim\pi_{\bm{\phi}}}\big[r_{\bm{\theta}}\left(x,y\right)\big].
    \end{equation*}
    
    Typically, an additional per-token KL penalty derived from the SFT model $\pi^\text{SFT}$ is involved to mitigate the reward over-optimization.
    In addition, the integration of gradients from pre-training distribution $\mathcal{D}_\text{pretrain}$ helps maintain model performance, denoted as PTX loss in \cite{ouyang2022training}. As a result, a more comprehensive practical objective function is introduced:
    \begin{equation*}
        \mathcal{J}({\bm{\phi}}) = \mathbb{E}_{x\sim\mathcal{D}_\text{RL}, y\sim\pi_{\bm{\phi}}}\Big[r_{\bm{\theta}}(x,y) - \beta\log\big(\pi_{\bm{\phi}}(y|x)/\pi^\text{SFT}(y|x)\big)\Big] + \eta ~ \mathbb{E}_{(x,y)\sim\mathcal{D}_\text{pretrain}}\Big[\log\big(\pi_{\bm{\phi}}(y|x)\big)\Big],
    \end{equation*}
    where $\beta$ and $\eta$ are coefficients determining the intensity of the KL penalty and the mixture of pretraining gradients respectively.
    This process refines the LLMs to generate responses that better align with human preferences for the prompts used during training.

\end{itemize}

Though RLHF has been proven effective for aligning LLMs with human preferences, this method has problems like complex implementation, hyper-parameter tuning, sample efficiency \citep{choshen2019weaknesses}, and computational overhead \citep{yuan2024rrhf}, making it hard to scale up.

A straightforward approach is rejection sampling \citep{dong2023raft, touvron2023llama} paired with finetuning on the best examples. For every prompt, $K$ responses are sampled from the model. Each response is then assessed with the reward model, and the one with the highest reward is selected as the best response. 
This selected response is later used for model fine-tuning.
\citet{zhang2023wisdom} formulates the language model instruction alignment problem as a goal-reaching reinforcement learning problem and proposes the HIR algorithm. The method unfolds in two stages: online sampling and offline training. During online sampling, the algorithm samples the LLM at a high temperature. In the offline training stage, instructions are relabeled based on generated outputs, followed by supervised learning using this relabeled data. HIR capitalizes on successful and failed cases without requiring additional parameters.
RRHF, as introduced by \citep{yuan2024rrhf}, aligns model probabilities with human preferences by scoring and ranking responses from multiple sources. With the necessity for only 1 or 2 models, its implementation is straightforward. RRHF reported it can effectively align language models with human preferences, producing performance on par with PPO.
\citet{gulcehre2023reinforced} proposes the ReST algorithm, which contains two loops: \textit{Grow} and \textit{Improve}. The \textit{Grow} loop uses the current model to sample and generate a dataset, while the \textit{Improve} loop iteratively trains the model on a fixed dataset. This algorithm provides a simple and efficient framework that allows repeated use of the fixed dataset to improve computational efficiency, showing significant improvement in the reward model scores and translation quality compared to supervised learning baselines. 
Motivated by the dependence of reward modeling on policy optimization in RLHF, \citet{chakraborty2024parl} propose PARL, a bilevel optimization-based framework.

\citet{rafailov2024direct} introduces the DPO, which demonstrates a mapping between reward functions and optimal policies. DPO is both simple and efficient, optimizing language models directly from human preference data, eliminating the need for an explicit reward model and multi-stage training. Moreover, \citet{wang2024beyond} discusses how diverse divergence constraints influence DPO and introduces a generalized approach, namely, \textit{f}-DPO.
\citet{azar2023general} presents a general objective, $\Psi$PO, designed for learning from pairwise human preferences, circumventing current methods' assumption: \textit{pairwise preferences can be substituted with pointwise rewards}. This objective analyzes RLHF and DPO behaviors, revealing their potential overfitting issue. The authors further delve into a specific instance of $\Psi$PO by setting $\Psi$ as the Identity, aiming to mitigate the overfitting problems. They call this method IPO and furnish empirical results contrasting IPO with DPO.
\citet{hejna2024contrastive} introduces CPL, which utilizes a regret-based model of preferences that directly provides information about the optimal policy.

Further research could explore why RLHF performs effectively with LLMs and the application of RLHF in multimodal \citep{rt2_deepmind_2023, openai2023gpt4V} settings to facilitate the benefits of human-AI collaboration \citep{human_ai_safe1, wu2021recursively, bi2021safety}. See also \citet{casper2023open} who offer a survey of open problems with RLHF.

\paragraph{Open Discussion}

RLHF is frequently applied to the Safety Alignment of LLMs, yet many pressing issues remain unresolved. For example, how can we balance harmlessness and helpfulness in alignment? \citet{dai2024safe} attempt to integrate the SafeRL framework, specifically the cost model and reward model, into RLHF to address the inherent tension between these two indicators. Moreover, even without malicious intent, simply fine-tuning on benign and commonly used datasets can inadvertently reduce the safety alignment of LLMs, albeit to a lesser extent \citep{qi2024finetuning} and fine-tuning on benign data is more likely to degrade the model’s safety \citep{he2024s}. These findings suggest that fine-tuning aligned LLMs may introduce new safety risks, even with datasets that are considered absolutely safe. Generally, language models may exhibit \textit{elasticity}, making them resistant to alignment efforts \citep{ji2024language}. This raises a question: \textit{how can we maintain impeccable safety alignment of models, even after further fine-tuning?}

Human preferences can vary among individuals, groups, and societies, leading to divergent perspectives. This divergence is also evident when collecting preference data from annotators. To address this, \citet{findeis2024inverse} proposed a method to extract the underlying constitution governing the generation of a given dataset of preferences. Similar to Constitutional AI \citep{bai2022constitutional}, where a preference dataset is generated by an LLM based on a predefined constitution, \textit{Inverse Constitutional AI} aims to extract such a constitution that can be used to reconstruct the preference dataset. This problem can be formulated as an optimization problem:
\begin{equation*}
    \mathop{\operatorname{argmax}}_{c} \{\text{agreement}(po, p(c)) \text{ s.t. } |c| \leq n\},
\end{equation*}
where $po$ represents the original preferences, and $p(c)$ are the constitutional preferences over a pairwise text corpus $T$, generated by an LLM $M$ using the constitution $c$. The set is constrained to a maximum of $n$ natural language principles that are human-readable. Agreement is defined as the percentage of constitutional preferences $p(c)$ that match the original preferences $po$. Overall, the elicitation of a constitution can be seen as a compression task, where a constitution is generated based on a dataset and then used to reconstruct the preferences in the dataset as accurately as possible.
To elicit such a constitution, the authors propose an algorithm that generates principles capable of explaining the preference data, followed by semantic clustering of these principles. To reduce the size of the set, they then subsample the principles and evaluate their ability by testing their reproducibility in reconstructing the preference data. Finally, the principles are filtered based on their relevance to the preference data.
This method can be used to infer the constitution underlying a specific preference dataset and has the potential to identify underlying biases or reuse the constitution to generate new data, thus enlarging existing datasets or creating new datasets tailored to individual preferences.

\subsection{Scalable Oversight: Path towards Superalignment}
\label{ssec:scalable_oversight}
Statistical learning usually rely on certain assumptions about data distribution, such as independence and identical distribution. Consequently, these algorithms fail in some situations, especially under specific distributions \citep{zhou2022domain}. Challenges in elementary systems can be promptly identified through visual inspection \citep{christiano2018supervising, ngo2024the}. As AI systems become more powerful, insufficiently capturing the training signal or erroneous design of loss functions often leads to catastrophic behaviors \citep{russell2015research, hubinger2019risks, cotra2021the} such as deceiving humans by obfuscating discrepancies \citep{russell2019human}, specification gaming \citep{specification2020victoria}, reward hacking \citep{brown2020safe}, and power-seeking dynamics \citep{carlsmith2022power}. 

From a human perspective, these imply gaps between the optimized objectives of AI systems and the ideal goals in our minds. Thus, the issue of providing effective oversight in various decision-making becomes pivotal \citep{bowman2022measuring, li2023trustworthy}, often termed as \textit{scalable oversight} \citep{amodei2016concrete} arising from two practical challenges.
\begin{itemize}[left=0.3cm]
    \item The high cost of humans frequently evaluating AI system behavior. For instance, the training process is time-consuming, and incorporating humans directly into the training loop in real-time would significantly waste human resources and impede training efficiency \citep{christiano2017deep}.
    \item The inherent complexity of AI system behaviors makes evaluation difficult, especially on hard-to-comprehend and high-stakes tasks \citep{saunders2022self}, \textit{e.g.}, tasks such as teaching an AI system to summarize books \citep{wu2021recursively}, generate complex pieces of code \citep{pearce2022asleep}, and predict future weather changes \citep{bi2023accurate}.
\end{itemize}

% \begin{tcolorbox}[colback=gray!20, colframe=gray!50, sharp corners, center title]
% \centering
% \textit{}
% \end{tcolorbox}

\begin{quotebox}
\large Scalable oversight seeks to ensure that AI systems, even those surpassing human expertise, remain aligned with human intent.
\end{quotebox}

In this context, our primary focus is to present some promising directions that may have not yet been implemented generally for constructing scalable oversight \citep{amodei2016concrete,leike2018scalable}.

\begin{figure}[t]
\centering
\footnotesize
        \begin{forest}
            for tree={
                forked edges,
                grow'=0,
                draw,
                rounded corners,
                node options={align=center,},
                text width=2.7cm,
                s sep=6pt,
                calign=child edge, calign child=(n_children()+1)/2,
            },
            [Scalable Oversight, fill=gray!45, parent,
                [\RLxF, for tree={
                pretrain,
                calign=center,
                },
                    [RLAIF, for tree={
                    pretrain,
                    calign=child edge, calign child=(n_children()+1)/2,
                    },
                        [\citenumber{bai2022constitutional, lee2023rlaif},  pretrain_work]
                    ]
                    [RLHAIF, for tree={
                    pretrain,
                    calign=child edge, calign child=(n_children()+1)/2,
                    },
                        [\citenumber{wu2021recursively, bowman2022measuring, saunders2022self, perez2022discovering},  pretrain_work]
                    ]
                ]
                [IDA,  pretrain 
                    [\citenumber{cotra2018iterated, christiano2018supervising, yudkowsky2018challenges, reed2022generalist},  pretrain_work]
                ]
                [RRM,  pretrain
                    [\citenumber{leike2018scalable},  pretrain_work]
                ]
                [Debate,  pretrain
                    [\citenumber{irving2018ai, du2023improving},  pretrain_work]
                ]
                [CIRL,  pretrain
                    [\citenumber{hadfield2016cooperative, hadfield2017inverse, specification2020victoria, shah2020benefits, everitt2021reward, pan2021effects, skalse2022defining, shevlane2023model, park2023ai},pretrain_work]
                ]
            ]
        \end{forest}
            \caption{A tree diagram summarizing the key concepts and literature related to Scalable Oversight. The root node represents Scalable Oversight whose goal is \textit{ensuring AI systems remain aligned with human intent even as they surpass human capabilities}. The main branches represent promising frameworks such as Reinforcement Learning from Feedback (RLxF), Iterated Distillation and Amplification (IDA), Recursive Reward Modeling (RRM), Debate, and Cooperative Inverse Reinforcement Learning (CIRL). Further sub-branches list key works exploring each framework. This diagram provides an overview of research directions for constructing effective and safe oversight mechanisms as AI systems grow more complex.}
\end{figure}
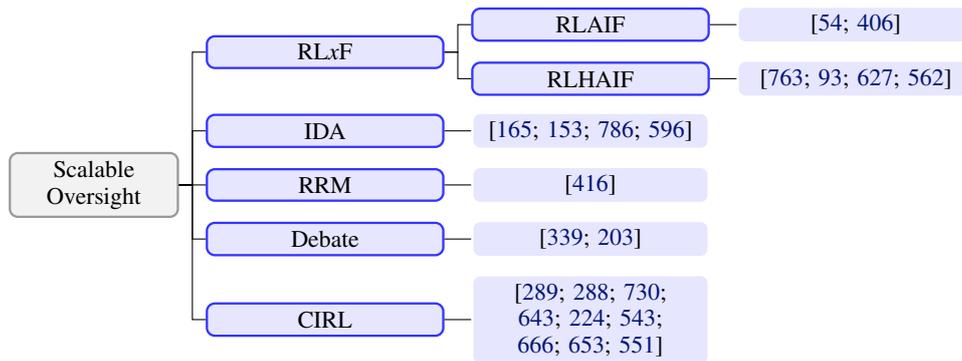

\subsubsection{From RLHF to \RLxF}
\label{sec:RLxF}
The RLHF paradigm offers a framework for aligning complex systems \citep{openai2023gpt4, touvron2023llama}. However, it encounters obstacles such as the inaccuracy of human evaluations and their associated high costs \citep{christiano2017deep, casper2023open, perez2022discovering}. A key limitation is the difficulty in utilizing RLHF to extend human feedback when creating AI systems with superhuman abilities \citep{wu2021recursively}. Building on the RLHF paradigm, we introduce \textit{RLxF} as a fundamental framework for scalable oversight, aiming to enhance feedback efficiency and quality and expand human feedback for more complex tasks. This enhances RLHF by incorporating AI components \citep{fernandes2023bridging}. The \textit{x} in \textit{RLxF} signifies a blend of AI and humans. We further explore concrete methodologies about \textit{RLxF} in the subsequent section. 

\paragraph{Reinforcement Learning from AI Feedback (RLAIF)}
RLAIF serves as an extension to RLHF.
RLAIF extends the pipeline 
\citet{bai2022training} found that LLMs trained via RLHF may avoid sensitive and contentious issues, potentially reducing models' overall utility. To address these limitations, \citet{bai2022constitutional} proposed a training pipeline that uses feedbacks generated by the LLMs (\textit{e.g.}, GPT-4 or other language models). Following pre-set criteria, the policy model self-evaluates and revises its responses during \textit{red teaming}. The initial policy model is then fine-tuned using the revised responses. Finally, the fine-tuned policy model evaluates the harmlessness of another language model's responses (i.e., AI feedback). Similar to RLHF , a reward trained using this feedback to optimize the policy model. \citet{lee2023rlaif} compares the performance of models trained with RLAIF and RLHF on summarization tasks. Their results suggest that models trained with AI feedback aperformed almost identically to those trained with human feedback, though subtle differences remain. Conversely, \citet{findeis2024inverse} explored the inverse problem of CAI: \textit{given a dataset of feedback, how can one extract a constitution that best enables a LLM to reconstruct the original annotations?} This problem not only converts AI feedback from preferences into a corresponding constitution but also offers a method for synthesizing new preference data for AI feedback.

\paragraph{Reinforcement Learning from Human and AI Feedback (RLHAIF)} RLHAIF integrates human and AI models to provide oversight. \citet{wu2021recursively} explores the feasibility of using AI to assist humans in summarizing books. This method facilitated human supervision and evaluation of the model performance by decomposing the book summarization task into subtasks, creating a tree-like structure. Meanwhile, \citet{saunders2022self} explores the use of AI to  assist in human assessment of model efficacy. Their findings suggest that model-generated critiques help humans identify flaws they might have missed. \citet{bowman2022measuring} proposes a proof-of-concept experiment to demonstrate the potential of scalable oversight techniques based on \textit{sandwiching} \citep{cotra2021the}. When collaborating with an unreliable LLM, the outcomes reveal that humans significantly surpass the model and themselves. \citet{perez2022discovering} employs language models to autonomously generate datasets for evaluating the behavior of language models of varying scales. The authors produced 154 high-quality datasets validated by humans. These methods demonstrate the feasibility of using AI assistance to scale up human oversight over complex problems and various domains.

To some extent, RLAIF and RLHAIF offers a viable alternative for creating a training loop with minimal human intervention, thus reduciing training costs. AI supervision obeying transparent and accessible AI behavior guidelines may significantly aid in achieving scalable oversight \citep{bowman2022measuring}.

\paragraph{Discussion} Efforts are underway to enhance RLHF by replacing pure humans alone \citep{leike2018scalable}. Given the multidimensional nature of human feedback, various approaches have been devised to offer focused human judgments informed by specific rules. Examples of such rules encompass considerations like chat fluency \citep{saunders2022self} and privacy safeguards \citep{rlpf_andrew_carr_2023}. \citet{saunders2022self} deconstructs the requirements for quality dialogue into natural language guidelines that an agent should adhere to, asking for evaluations on each guideline individually. We can attain more efficient rule-conditioned reward models by collecting targeted human assessments and training models on this data. This approach substantially enhances the efficacy of dialogue agents, rendering them more helpful, accurate, and benign when compared to prompted language models. \citet{rlpf_andrew_carr_2023} proposes Reinforcement Learning from Privacy Feedback (RLPF), aiming to harmonize the output quality of language models with safeguarding privacy. The method exploits NLP techniques to conduct real-time privacy risk assessments of text generated by the models and subsequently adjusts the reinforcement learning feedback signals based on these evaluations. Expressly, if the generated text includes sensitive information, it incurs negative feedback, whereas high-quality, non-revelatory text receives positive feedback. As the model undergoes training, it incrementally refines its capabilities, enhancing text quality and minimizing privacy breaches concurrently. This approach offers a more efficient evaluation of privacy risks by employing established NLP techniques, in contrast to conventional learning methods, which depend heavily on large-scale manual data annotation. 

At their core, the \textit{RLxF} methods utilize the strategy of decomposing a large problem into smaller sub-problems, enabling the use of more efficient tools, such as AI and software, for rapid sub-problem resolution. By leveraging the solutions to these sub-problems, the resolution of the main issue can be expedited. These techniques can be regarded as elementary instances of IDA; the primary distinction lies in the absence of a continual iterative process. Nonetheless, evidence suggests they are promising to offer feedback for AI systems that exceed human performance \citep{wu2021recursively}. Consequently, these methods can serve as foundational techniques in the training of more advanced AI systems.

\subsubsection{Iterated Distillation and Amplification}
\label{ssec:iterated_distillation_and_amplification}

Iterated Distillation and Amplification (IDA) introduces a framework for constructing scalable oversight through iterative collaboration between humans and AIs \citep{christiano2018supervising}. The process commences with an initial agent, denoted as $A[0]$, which mirrors the decision-making of a human, $H$. $A[0]$ undergoes training using a potent technique that equips it with near-human-level proficiency (the distillation step); Then, collaborative interaction between $H$ and multiple $A[0]$ instances leads to the creation of an enhanced agent, $A[1]$ (the amplification step). The successive process is described\footnote{We reference the pseudo-code by \citet{cotra2018iterated} for this description.} in Algorithm \ref{alg:ida}. 

\citet{cotra2018iterated} distinguishes between broad and narrow definitions within both RL and IRL. Broad RL gives sparse reward signals to AI systems and allows autonomous exploration and optimization of cumulative future rewards. This can lead to super-human novel strategies but makes it hard to specify what we care about perfectly.
Narrow RL gives dense feedback rewarding the reasonableness of choices instead of final outcomes. This makes ML systems more human-like but limits capabilities.
Similarly, broad IRL infers deep long-term values from the full range of human behaviors, while narrow IRL only infers short-term instrumental values. The former is a higher risk, while the latter is limited in capabilities.

During IDA training, narrow techniques are needed to ensure each agent itself mimics human behaviors. Specifically, narrow RL or IL can be used to train the agent to be as human-like and controllable as possible.
Humans can leverage agents' computing power and parallelizability to devise more far-sighted, macro strategies. This is essentially an amplification of human intrinsic capabilities.
In the next iteration, agents again mimic this strengthened human-machine system using narrow techniques. This enables a gradual transition from narrow ability to broad ability while keeping the agents aligned with human values.
As iterations increase, the human-machine system becomes more and more capable, gradually approximating a system that is both highly capable and aligned with human values, achieving both safety and capability. In other words, Narrow techniques are used to ensure agents follow human values, while the broadened human strategies in the amplification stage are a way of utilizing the agents, and do not expand the agents' own learning goals.

IDA is well illustrated by AlphaZero \citep{christiano2018supervising, nguyen2020ida}. The algorithm starts with a simple policy (\textit{e.g.}, random move selection) and learns from its self-play games, the \textit{amplification} phase. It then uses these games as training data to develop better move selection heuristics, the \textit{distillation} phase. This distillation-amplification process can be repeated to create a fast and proficient Go-playing AI. Here, the distinction between alignment and capability is crucial \citep{mennen2018comment}. An aligned but less capable AI tries to win but may not succeed against moderate opponents. A capable but poorly aligned AI achieves certain game properties other than winning. The goal is that AI is capable and aligned, proficient at the game, and aligned with the goal of winning the game.

\begin{algorithm}[t]
\caption{Iterative Distillation and Amplification}
\begin{algorithmic}[1]
\Procedure{IDA}{$H$}
\State $A \gets$ random initialization
\Repeat
\State $B \gets \Call{Amplify}{H, A}$
\State $A \gets \Call{Distill}{B}$ \Comment{Repeat indefinitely}
\Until{False}
\EndProcedure

\Procedure{Distill}{overseer}

\Return{An AI trained using narrow, robust techniques to perform a task that the overseer already understands how to perform.}
\EndProcedure

\Procedure{Amplify}{human, AI}

\Comment{Interactive process in which human uses many calls to AI to improve on human's native performance at the relevant tasks.}
\EndProcedure
\end{algorithmic}
\label{alg:ida}
\end{algorithm}

The feasibility of IDA has sparked considerable debate \citep{yudkowsky2018challenges}. IDA operates under a crucial assumption that \textit{errors won't continuously accumulate throughout the iterations} \citep{leike2018scalable}. Thus, technical challenges persist during the distillation and amplification step, necessitating sufficiently advanced and safe learning techniques. Additionally, despite the original authors likening IDA to the training process of AlphaZero \citep{silver2017mastering} and having demonstrated it in toy environments \citep{christiano2018supervising}, its practicality hinges on ensuring that $H$ can delegate portions of complex tasks to A, analogous to a leader orchestrating a team to accomplish a project collectively. In practice, Gato \citep{reed2022generalist} illustrates key aspects of IDA \citep{mukobi2021gato} that may pave the way to AGI. It consolidates the abilities of multiple expert AIs into a singular model, validating that IDA's distillation can be achieved using contemporary deep learning. While not fully realized, Gato hints at amplification potential, harnessing its diverse skills to accelerate the learning of new tasks. However, Gato lacks safe amplification or distillation methods to maintain alignment properties. Crafting alignment-preserving IDA methods suited for models like Gato remains a crucial direction for AI safety research. In essence, while Gato signifies notable progress in actualizing IDA, further theoretical advancements are imperative to ensure that the IDA framework leads to safe AGI.

\subsubsection{Recursive Reward Modeling}
\label{par:rrm}

As discussed in \S\ref{par:reward_model}, reward modeling leverages the idea of using human feedback to train a reward model, which an agent then pursues. It allows us to disentangle the construction of the system's objective from evaluating its behavior \citep{ibarz2018reward}. In this manner, the reward model provides insights into the optimization direction of the AI system. Particularly noteworthy is the ability to finely align the system with human intentions and values, such as fine-tuning language models to adhere to human instructions \citep{bai2022training, touvron2023llama}. Also, reward modeling has proved valuable in advancing AI research \citep{zhao2023survey, bukharin2023deep}. Recursive Reward Modeling (RRM) \citep{leike2018scalable} seeks to broaden the application of reward modeling to much more intricate tasks.
The central insight of RRM is the recursive use of already trained agents $A_{t-1}$ to provide feedback by performing reward learning on an amplified version of itself for the training of successive agents $A_{t}$ on more complex tasks. The $A_0$ is trained via fundamental reward modeling (learned from pure human feedback). This approach is not only influenced by human feedback but also by the model's own assessments of what constitutes a rewarding outcome.
If the assumption that \textit{evaluating outcomes is easier than producing behavior} holds, then the iterative process of reward modeling can iteratively achieve higher capacity to oversee more powerful AI systems, paving the way for extending oversight into more complex domains. This process is detailed in Algorithm \ref{alg:rrm}. 

For instance, we aim to train AI $A$ to devise a comprehensive urban plan. Designing a city entails numerous intricate elements, such as traffic planning, public amenities, and the distribution of residential and commercial zones. Evaluating a city's entire design poses a significant challenge since many issues may only become apparent after extended real-world testing.
To aid this process, we may need an agent $B$ specifically for traffic planning. However, traffic planning in itself is a multifaceted task. Consequently, we further need other agents to assess aspects such as road width, traffic flow, and the design of public transportation.
For every sub-task, such as gauging road width, we can train an auxiliary agent to verify if safety standards are met, if various modes of transportation have been considered, and so on. In doing so, we establish an RRM process where each agent is trained with the help of agents assessing sub-tasks.

This approach resembles the organizational structure of a large corporation \citep{leike2018scalable}. In the context of urban planning, the main planning team (the CEO) is responsible for the final design decisions. Their decisions are informed by recommendations from the traffic team (the department managers), who, in turn, base their recommendations on inputs from the road width team (the managers), and so forth. Each level of decision-making relies on feedback from the level below it, with each task optimized through reward modeling.

The challenges faced by RRM can be described around the concepts of outer and inner alignment \citep{hubinger2020overview}. Outer alignment revolves around the sufficiency of feedback mechanisms to guarantee that the learned reward model is accurate in the domain perceived by the action model as on distribution. This challenge is contingent on several factors, including the quality of human feedback, the difficulty of generalization, and the potential for agent deception. Conversely, inner alignment concentrates on how effectively a human can employ transparency tools to prevent deceptive or disastrous behaviors in both the reward model and the agent. This hinges on the effectiveness of the oversight mechanism and the capacity to verify that the reward model isn't undergoing any optimization and that the agent remains myopic \citep{cotra2018iterated}.

\begin{algorithm}[t]
\caption{Recursive Reward Modeling}\label{alg:iterative_training}
\begin{algorithmic}[1]
\State Initialize agent $A_0$ using reward modeling based on user feedback.
\Comment Either preferences or numerical signals.
\For {$t$ = 1, 2, \dots}
\State Use $A_{t-1}$ to assist users in evaluating outcomes.
\State Train agent $A_t$ based on user-assisted evaluations.
\Comment Objective of $A_t$ is generally more complex than that of $A_{t-1}$.
\EndFor
\end{algorithmic}
\label{alg:rrm}
\end{algorithm}

Potential approaches to mitigate these challenges \citep{leike2018scalable} include online feedback to correct the reward model during training \citep{christiano2017deep}, off-policy feedback to teach about unsafe states \citep{everitt2017reinforcement}, leveraging existing data like videos and text via unsupervised learning or annotating \citep{baker2022video}, hierarchical feedback on different levels \citep{bukharin2023deep} adversarial training to discover vulnerabilities \citep{madry2018towards}, and uncertainty estimates for soliciting feedback \citep{hadfield2016cooperative, macglashan2017interactive}. The strength of RRM is its competitive training approach, which necessitates human feedback instead of demonstrations, potentially making feedback more reliable and simpler to obtain \citep{hubinger2020overview}. In essence, the process of RRM can be likened to IDA \citep{christiano2018supervising}, where reward modeling takes the place of supervised or imitation learning. Thus, the challenges confronted by RRM closely mirror those encountered in IDA, particularly in preventing the accumulation of errors. Additionally, reward modeling itself does not necessarily distill a \textit{narrow} model \citep{cotra2018iterated}, which presents challenges in trading off the degree of alignment and performance.

\subsubsection{Debate}
\emph{Debate} involves two agents presenting answers and statements to assist human judges in their decision-making \citep{irving2018ai}, as delineated in Algorithm \ref{alg:debate}. This is a zero-sum debate game where agents try to identify each other's shortcomings while striving to gain higher trust from human judges, and it can be a potential approach to constructing scalable oversight. For example, in the game of Go, human judges might not discern the advantage side of the single game board itself. However, by observing the game's process and the eventual outcome, these judges can more easily deduce that.

The premise of this method relies on a critical assumption: \textit{arguing for truth is generally easier than for falsehood}, granting an advantage to the truth-telling debater. However, this assumption does not hold universally. For instance, in a complex problem, humans might fail to comprehend the specialized concepts used in the debate. Additionally, the limited nature of the gradient descent may bring us to an undesirable cyclic pattern (\textit{i.e.}, when optimizing for one property, such as honesty and highlighting flaws, models often overlook or diminish another) \citep{irving2018ai}.

It's worth mentioning that with the advancement of LLMs' capabilities, we can already see practical examples of debate \citep{du2023improving, lw2023debate}. Challenges may arise for debate in specific real-world scenarios \citep{irving2018ai}. For example, certain questions may be too intricate for human comprehension or too voluminous to present in their entirety. Similarly, there are instances where an optimal answer to a question is exceedingly lengthy, envision a response that spans a hundred pages. To navigate these, agents might initially select a response and, as the debate progresses, reveal sections of either the question or the answer. \citet{irving2018ai} conducts a toy experiment on this process. 
Meanwhile, we must acknowledge the limit of human time. In scenarios that necessitate interaction with the environment, such as directing a robot, each action might demand a distinct debate. It's not always feasible for humans to judge every debate due to time constraints. In response to this challenge, we may need to design ML models to predict human feedback. In line with this observation, \citet{khan2024debating} experimented with using smaller, non-expert models as judges in debates between two expert models, both of which had access to the underlying data and the ability to quote from it. The experiments demonstrated that these smaller non-expert models were able to achieve higher accuracy when relying on the expert model debates, though they still underperformed compared to human judges. Additionally, the expert models can be optimized for persuasiveness, enabling the judges to attain even greater accuracy and more easily identify the truth. The authors emphasize that debate implementations must be grounded in verifiable evidence to prevent debaters from fabricating facts. Further work on using weaker models as judges in debates guided by stronger models was conducted by \citet{kenton2024scalable}. Their experiments focused on tasks involving both information asymmetry and symmetry between debaters and judges and were extended to include multimodal inputs. The protocols they applied evaluated the baseline performance of judges without debate protocols, alongside debate and consultancy protocols. These experiments considered both assigned positions and cases where debaters or consultants could choose their positions. Experimental results showed that debate consistently outperforms consultancy. Weak judges struggle to fully leverage debate protocols, and consultancy can significantly reduce the accuracy of judges, particularly when the consultant advocates for an incorrect solution. Overall, the authors interpret their findings as only weakly promising for the debate framework. However, these experiments were conducted solely with models at inference time, and debate protocols may hold greater potential when integrated into training. This is particularly relevant given that the task of judging a debate can be seen as OOD for models primarily fine-tuned for question answering.

Another consideration is the convergence of the debate mechanism \citep{irving2018ai}. \citet{du2023improving} showcases the inherent tendency of the debate framework to eventually converge toward singular responses, even if accuracy is not guaranteed. Meanwhile, if challenges arise in achieving convergence, we might have to rely on intuition to gauge the effectiveness of convergence. This implies the requirement of human evaluators' intervention and demands a certain level of expertise from these human assessors, posing challenges that must be addressed.

Furthermore, there are many discussions originating from diverse perspectives. \citet{ngo2021excited} considers \textit{Debate} as one type of iterated amplification but more specific to make safety ground in concrete research questions, and its adversarial framing makes it easier to spot problems. \citet{michaelcohen2020} expresses concerns regarding the adverse implications of incentivizing debaters to employ deceptive strategies aimed at swaying the judgment process. \citet{struart2019, beth2020debate} expound upon the various issues that can permeate the debate process, including challenges such as the obfuscated arguments problem, ambiguous responses, and the propagation of misleading implications. While one may affirm the presence of a sufficiently low probability of any underlying flaws within the argument, advocating for trustworthiness, the opposing debater may assert the existence of a sufficiently high probability of identifying a flaw within the argument somewhere, thus advocating for a lack of trust.
\citet{beth2020progress} introduces the concept of \textit{cross-examination} to incentivize debaters to provide more informative responses. In this process, debaters have the agency to select a prior claim for scrutiny and obtain a copy of the opposing debater's response. The entire exchange is documented, and debaters can present relevant segments to the judge. The introduction of cross-examination is a robust deterrent against dishonest debaters exploiting a sweeping narrative, in contrast to their prior arguments, to mislead the judge. 

\begin{algorithm}[t]
\caption{Debate}\label{alg:ai_debate_process}
\begin{algorithmic}[1]
\State Initialize set of questions $Q$.
\State Initialize two competing agents.

\State Select a question $q \in Q$.
\Comment Question is shown to both agents.

\State Agents provide their answers $a_0$ and $a_1$.
\ The agents generate comment answers in response to $q$.

\State Initialize debate transcript $T$ as an empty list.

\For{turn in predefined number of debate turns}
    \State Agent makes a debate statement $s$.
    \State Append $s$ to $T$.
    \Comment Agents take turns and statements are saved in the transcript.
\EndFor

\State Judge observes $(q, a_0, a_1, T)$ and decides the winning agent.

\end{algorithmic}
\label{alg:debate}
\end{algorithm}

There exists a notable similarity between the debate \citep{irving2018ai}, IDA \citep{christiano2018supervising}, and RRM \citep{leike2018scalable}. These approaches can be comprehended in the view of an underlying principle: \textit{evaluation can be simpler than task completion}\footnote{Discussions about this can also be found in the literature about these methods.}. Therefore, harnessing the evaluative capabilities of AI systems can result in distributions of capacity that are more advantageous for humans. The challenges these methods face, especially in mitigating the accumulation of errors, are also analogous.

\subsubsection{Cooperative Inverse Reinforcement Learning}
\label{sec:cirl}

Almost all previous methods consider learning from feedback a process separate from inference and control and often implicitly consider feedback providers as entities existing outside of the environment -- indeed, failure modes like manipulation \citep{shevlane2023model} and reward tampering \citep{everitt2021reward} occur exactly when feedback mechanisms that are supposedly outside of the environment become part of it and therefore subject to the AI system's influence. The framework of Cooperative Inverse Reinforcement Learning (CIRL), however, unifies control and learning from feedback and models human feedback providers as fellow agents in the same environment. It approaches the scalable oversight problem not by strengthening oversight but by trying to eliminate the incentives for AI systems to game oversight, putting humans giving feedback and the AI system in cooperative rather than adversarial positions \citep{shah2020benefits}. In the CIRL paradigm, the AI system collaborates with humans to achieve the human's true goal rather than unilaterally optimizing for human preferences.

\paragraph{Motivation and General Idea of CIRL} Many modes of misalignment, including, for example, reward hacking \citep{specification2020victoria,skalse2022defining}, deception \citep{park2023ai}, and manipulation \citep{shevlane2023model}, are results of the AI system confidently optimizing for misspecified objectives \citep{pan2021effects}. During training and deployment, the specified objective (\textit{e.g.}, the reward function) plays the role of an unchallengeable truth for the AI system, and human feedback is only respected to the extent specified in the objective, which means that it could be tampered \citep{everitt2021reward} or manipulated \citep{shevlane2023model}.

CIRL \citep{hadfield2016cooperative,hadfield2017inverse,shah2020benefits} attempts to mitigate this problem by (1) having the AI system explicitly hold uncertainty regarding its reward function, and (2) having humans provide the only information about what the reward function truly is. This uncertainty gives the AI system a tendency to defer to humans and a drive to determine what the human truly wants. Concretely speaking, it models the entire task as a two-player cooperative game, where the \textit{human player} $H$ and the \textit{robot player} $R$ share a common reward function $r(\cdot)$. Importantly, the reward function and reward signals aren't visible to $R$ (and indeed aren't explicitly calculated by the training mechanism) and are only inferred by $R$ from behaviors of $H$ via an IRL-like process (including by asking and interacting with $H$). This game has been called the \emph{CIRL} \citep{hadfield2016cooperative}, the \emph{assistance game} \citep{fickinger2020multi}, and the \emph{assistance POMDP} \citep{shah2020benefits}.

In short, the AI system has the human's true objective $r(\cdot)$ as its own goal (despite not knowing values of $r(\cdot)$ with certainty) and constantly tries to figure $r$ out by observing and interacting with the human. This reduces incentives for, \textit{e.g.}, manipulation since manipulation of human behaviors only serves to pollute an information source and does not affect $r$.

\paragraph{Formulation of CIRL}  \citet{hadfield2016cooperative} characterizes the settings of CIRL (which we denote by $M$) by building upon classical multi-agent MDPs, resulting in the definition below of $M$.

\[M=\big\langle \mathcal{S},\{\mathcal{A^\mathbf{H}},\mathcal{A^\mathbf{R}}\}, T, \gamma, r,\Theta,P_0 \big\rangle\] 

In the equation
above, $S$ and $\{\mathcal{A^\mathbf{H}},\mathcal{A^\mathbf{R}}\}$ are the space of world states and actions respectively, $T:S\times\mathcal{A^\mathbf{H}}\times\mathcal{A^\mathbf{R}}\rightarrow \Delta(S)$ is the transition function, and $\gamma$ is the discount rate. Up to here, the definition is identical to that of a standard multi-agent MDP. The remaining elements, however, introduce the key difference: the reward function is parameterized, and its parameters can be modeled by a distribution. $\Theta$ is the space of values for the parameters $\bm{\theta}$; $r: S \times \mathcal{A^\mathbf{H}}\times\mathcal{A^\mathbf{R}}\times\Theta\rightarrow\mathbb{R}$ is the shared reward function, and $P_0\in\Delta(S\times\Theta)$ is the joint distribution of the initial state and the reward function's parameters. This parameterization approach allows $R$ to model explicitly and reason about its belief over the true reward function. Using techniques from \citet{nayyar2013decentralized}, any CIRL setting can be reduced to an equivalent single-agent POMDP, thus proving the existence of optimal policies that are relatively tractable \citep{hadfield2016cooperative}. 

\paragraph{Notable Directions in CIRL Research} Although some have emphasized the importance of $H$ teaching $R$ \citep{fisac2020pragmatic} actively, works \citep{shah2020benefits} have contested the emphasis on game equilibria and joint policies (including $H$'s pedagogic behaviors), and instead focuses on $R$'s optimal response to a policy of $H$'s, since the assumption that humans will always act on optimal joint policies is an unrealistic one. More specifically, \citet{shah2020benefits} considers the \textit{policy-conditioned belief} $B:\Pi^{\mathbf{R}}\rightarrow\Delta\left(\Pi^{\mathbf{H}}\right)$, which specifies $H$'s distribution over policy responses to any of $R$'s policies, and the aim is to find $R$'s optimal policy given $B$. Here, $B$ is essentially a form of human modeling, and one challenge is to obtain a robustly accurate human model as $B$ \citep{hong2022sensitivity}. On another front, \citet{hadfield2017inverse} and \citet{he2021assisted} examine the manual specification of an imperfect reward function as a way for $H$ to convey information about the true reward function. This includes work on $R$'s side (\textit{i.e.}, enabling $R$ to perform inference on the true reward function based on the imperfect specification) \citep{hadfield2017inverse} and also work on $H$'s side (\textit{i.e.}, developing algorithmic tools to assist $H$ in making more robust specifications that better convey the true reward function) \citep{he2021assisted}. Aside from improvements to the game settings, the design of more scalable CIRL algorithms has also been recognized as a priority.

There has also been work that extends CIRL and assistant games to multi-agent settings \citep{fickinger2020multi} where there are multiple humans that the robot needs to serve. This corresponds to the \emph{multi/single delegation} settings in \citet{critch2020ai}, where the varying objectives of humans create a challenge and necessitate the use of social choice methods.
\subsubsection{Circuit Breaking}
Instead of training a model to refuse harmful outputs, the circuit-breaking approach proposed by \citet{zou2024improvingalignmentrobustnesscircuit} directly controls the internal representations responsible for generating such outputs. A potential advantage of this method is that it aims to enhance safety without compromising performance. The core idea is to manage the model’s ability to generate harmful outputs, rather than eliminating the underlying vulnerabilities. This approach renders circuit-breaking attack-agnostic, as new adversarial attacks may emerge, but the internal representations associated with harmful outputs remain constant.

The circuit-breaking method consists of two essential components: the dataset and the loss function. The dataset includes harmless samples, where internal representations should remain unchanged, while the circuit-breaking set comprises harmful samples, which require altered internal representations to prevent the generation of harmful content. By employing a mean squared error loss for retaining representations and cosine similarity for circuit-breaking representations, the model becomes unable to generate harmful content. Changing the loss function for circuit-breaking can also allow for steering the model's generation in various directions, such as ending the output generation or refusing to answer. Additionally, this approach can be extended with a Harmfulness Probing mechanism to detect and respond to harmful generations.

The authors successfully applied this method to Large Language Models, Multimodal models, and Language Agents to control function calls.

\subsubsection{Weak-to-Strong Generalization}

Scalable Oversight can help humans provide supervision signals to AI systems that are smarter and more complex, ensuring that the behaviors of super-human-level AI systems align with human intent and values. However, what if we cannot obtain scalable supervision signals? An example is that for some tasks, evaluation is not necessarily simpler than generation, making it impossible to utilize task decomposition followed by AI assistance to achieve scalable oversight.

Recently, a generalization phenomenon called \textit{Weak-to-Strong Generalization} is verified, the core idea of which is to use weak supervision signals from a weak model to train a strong model \citep{burns2023weak}. Specifically, the weak model is trained on ground truth and then annotates new data with weak labels for training the strong model. The results across three settings (\textit{i.e.} NLP classification, chess puzzles and reward modeling) reflect that \textit{weak-to-strong generalization} is a robust phenomenon, yet there is room for further improvement, such as narrowing the gap between a strong model trained with weak labels and ground truth. \textit{Weak-to-Strong Generalization} provides a valuable analogy for the superalignment problem: how humans can supervise super AI systems as weak supervisors. 
The insight behind \textit{weak-to-strong generalization} is that the strong model can generalize beyond weak labels instead of merely imitating the behavior of weak models. In other words, the weak model elicits the strong model's capability. However, verifying \textit{weak-to-strong generalization} is challenging if humans don't know the ground truth. Nonetheless, \textit{weak-to-strong generalization} still offers a valuable perspective for solving the superalignment problem.

The framework for \textit{weak-to-strong generalization} has been further expanding and integrating with scalable oversight. Empirical results show that weak models can evaluate the correctness of stronger models by assessing the debate between two expert models \citep{khan2024debating}. Additionally, making expert debaters more persuasive improves non-experts' ability to discern truth in debates, evidencing the effectiveness of aligning models with debate strategies without ground truth. Some frameworks employ a external amplifier to create an iterated distillation and amplification process, which presents a potential framework for integrating \textit{weak-to-strong generalization} techniques with IDA during the training process \citep{ji2024aligner}.
Moreover, \citet{leike2023combine} proposes several methods to integrate scalable oversight with \textit{weak-to-strong generalization} techniques, \textit{e.g.}, recursively decomposing tasks into atomic ones (in line with scalable oversight principles), supervising these atomic tasks, and employing reward models trained with \textit{weak-to-strong generalization techniques} using human preference data.

\section{Learning under Distribution Shift}

\label{sec:distribution}

The construction of reliable AI systems is heavily dependent on their ability to adapt to diverse data distributions. Training data and training environments are often imperfect approximations of real deployment scenarios and may lack critical elements such as adversarial pressures \citep{poursaeed2021robustness} (\textit{e.g.}, Gaussian noise in the context of supervise learning-based systems \citep{gilmer2019adversarial} and shadow attack \citep{ma2012shadow} in autonomous-driving systems), multi-agent interactions \citep{critch2020ai,dafoe2021cooperative},  complicated tasks that human overseers cannot efficiently evaluate \citep{leike2018scalable},\footnote{This could contribute to the emergence of deceptive behaviors \citep{hubinger2019deceptive}. See the paragraph on \emph{goal misgeneralization} in \S\ref{sec:chal-dis} for details.} and reward mechanisms that can be gamed or manipulated \citep{krueger2020hidden}. This discrepancy between training distribution and testing distribution (or environments) is known as \emph{distribution shift} \citep{krueger2020hidden,thulasidasan2021effective}.

Therefore, AI systems that are aligned under their training distribution (\textit{i.e.}, pursuing goals that are in line with human intent) may not uphold their alignment under deployment (or testing) distribution, potentially leading to serious misalignment issues post-deployment. This potential failure motivates research on the preservation of alignment properties (\textit{i.e.}, adherence to human intentions and values) across data distributions.

From an alignment perspective, we are more concerned about AI systems pursuing unaligned and harmful goals, as opposed to incompetence at pursuing goals. Thus, the emphasis on alignment properties means that we focus on the generalization of \emph{objectives} across distributions, as opposed to the generalization of \emph{capabilities} \citep{di2022goal,ngo2024the}.

\begin{figure}[t]
    \centering
    \includegraphics[width=1.0\textwidth]{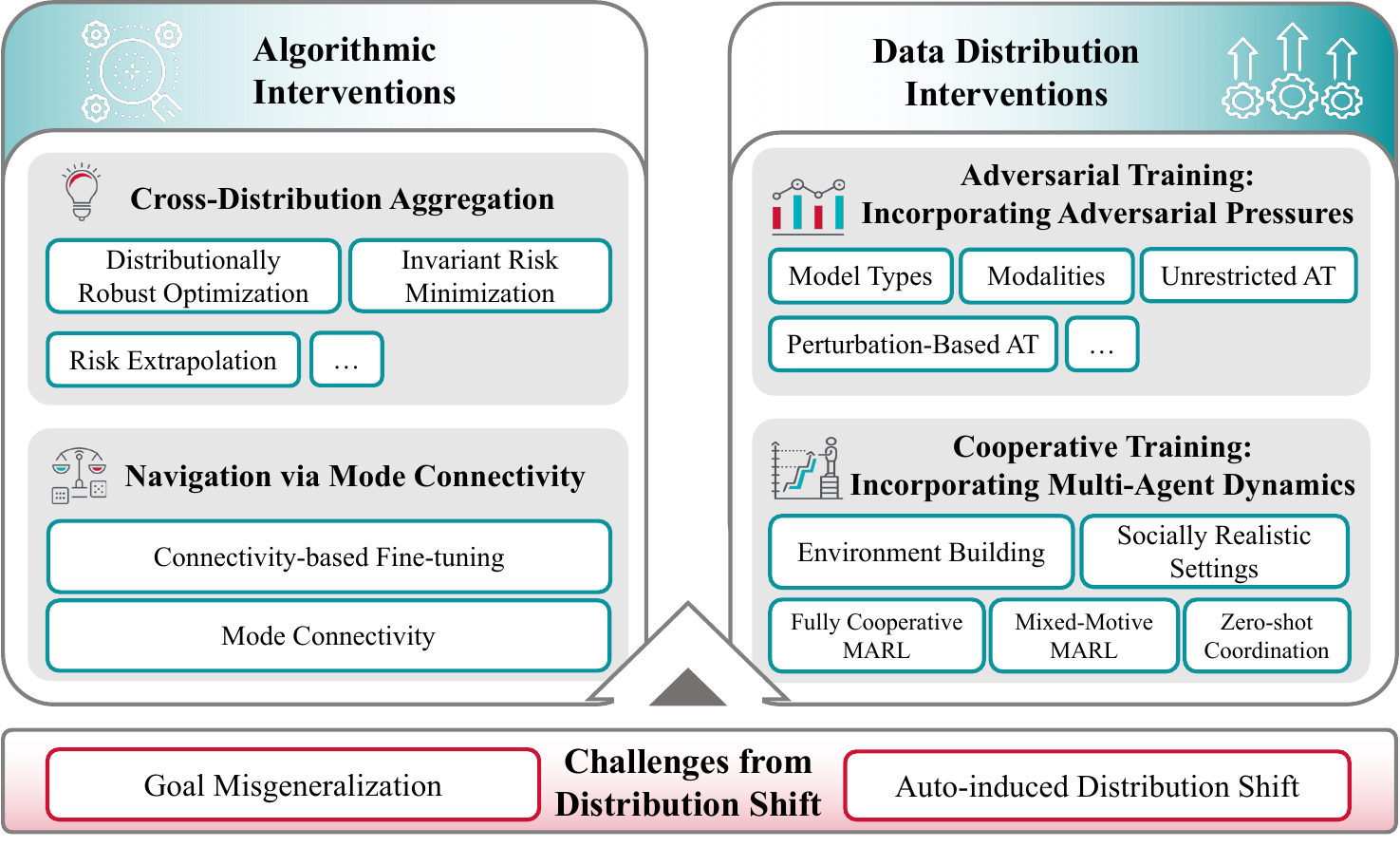}
    \caption{Framework of learning under distribution shift. The main challenges stemming from the distribution shift are goal misgeneralization and auto-induced distribution shift (\S\ref{sec:chal-dis}). In our framework, we also introduce two kinds of methods to address distribution shift: algorithmic interventions (\S\ref{sec:alg-inter}) that steer optimization during training, and data distribution interventions (\S\ref{sec:data-inter}) that expand the training distribution in a targeted manner by introducing real-world elements.}
    \label{fig:generalization}
\end{figure}

We mainly discuss the preservation of alignment properties when learning under distribution shift in this section. We start the discussion by introducing the alignment challenges from distribution shift (\S\ref{sec:chal-dis}). Subsequently, we delve into methods for addressing distribution shift, and discuss two approaches in particular: (1) algorithmic interventions (\S\ref{sec:alg-inter}) that steer optimization during the training process, and (2) data distribution interventions (\S\ref{sec:data-inter}) that expand the training distribution by introducing specific elements into the training process, including adversarial training \citep{yoo2021towards,tao2021recent,ziegler2022adversarial} and cooperative training \citep{dafoe2021cooperative} (\S\ref{sec:cooperative-ai-training}). Our framework for learning under distribution shift is shown in Figure \ref{fig:generalization}.

\subsection{The Distribution Shift Challenge}
\label{sec:chal-dis}

Before introducing the specific techniques, we initially demonstrate why one of the primary challenges in alignment is learning under distribution shift, and more specifically, the preservation of \emph{alignment properties} (\emph{i.e.}, adherence to human intentions and values) under distribution shift. 
We introduce two alignment challenges concerning the issue of distribution shift, namely goal misgeneralization \citep{di2022goal} and auto-induced distribution shift (ADS) \citep{krueger2020hidden}.

The training of AI systems optimizes for their adherence to the pursuit of the training reward/loss under the training input distribution. However, this adherence may not generalize to cases where the input distribution undergoes qualitative changes, \emph{i.e.}, distribution shift. These changes include, for example, adversarial pressures \citep{poursaeed2021robustness}, multi-agent interactions \citep{critch2020ai}, and complicated tasks that human overseers cannot efficiently evaluate \citep{di2022goal}, and reward mechanisms that can be gamed or manipulated \citep{krueger2020hidden}.

It's worth distinguishing two different failure modes here: goal misgeneralization \citep{di2022goal}, in which the original and shifted distributions are given, and auto-induced distribution shift \citep{krueger2020hidden}, where the AI system alters the data distribution with its own behaviors in pursuit of reward.

\paragraph{Goal Misgeneralization} This kind of challenge refers to the scenario where AI systems perform perfectly in the training distribution, but the capabilities learned in training distribution fail to generalize in OOD deployment, and AI may present the pursuit of goals that are not in accordance with human wishes \citep{di2022goal}. Goal misgeneralization\footnote{More examples of goal misgeneralization exist \citep{gmg2020}.} is to be distinguished from other forms of misgeneralization (\textit{e.g.}, capability misgeneralization) where the agent becomes incompetent in OOD settings; instead, agents with goal misgeneralization \emph{competently} pursue an \emph{unwanted} goal in OOD settings.

A simplistic example is the case of \emph{spurious correlations} (or \textit{shortcut features}) \citep{geirhos2018imagenettrained,di2022goal}. For example, in an image classification dataset, green grass is a highly predictive feature for the label \textit{cow}. However, it is essential to note that this feature needs to be more consistent and reliable across various data distributions \citep{pml2Book}. Moreover, the causal confusion (\textit{i.e.}, ignorant of the causal structure of the interaction between the advisor and the environment) in IL can result in goal misgeneralization \citep{de2019causal,tien2022causal}.

One major danger from goal misgeneralization lies in the indistinguishability between ``optimizing for what human really wants'' and ``optimizing for human thumbs-ups'';\footnote{Here, \textit{human thumbs-ups} refer to high-reward feedback from human advisors or environment. However, AI systems may deliberately follow human preferences or deceive to get high rewards from humans, but actually don't really learn intended goals (\textit{i.e.}, what human really wants).} the latter includes potentially deceiving or manipulating human evaluators \citep{shevlane2023model} to receive their thumbs-ups. For example, \citet{learning2017dario} discovered that in a task where a robotic hand is supposed to grasp a small ball, the robotic hand fakes the action by using parallax in front of the lens to appear as if it has grasped the ball, without actually doing so. This behavior deceives the human annotator into thinking that the task has been completed.

When an AI system is trained or finetuned with human feedback, it is impossible to distinguish the two goals since both perform perfectly in training, and it is unclear which one the AI system will learn. In fact, during training, the human evaluators might be deceived or manipulated, implying that the AI system may be more strongly incentivized to optimize for human thumbs-ups rather than what the human wants. Current examples of this phenomenon exist in recommender systems \citep{kalimeris2021preference,adomavicius2022recommender}, LLMs \citep{perez2022discovering}, and RL systems \citep{learning2017dario}.

Finally, one failure mode closely related to goal misgeneralization is the misalignment of \emph{mesa-optimizers} \citep{hubinger2019risks}, where the ML model with learned model weights performs optimization within itself during inference (``mesa-optimization'') \citep{hubinger2019risks,dai2023can}, and the objective of this optimization is not aligned with the model's training objective.

\paragraph{Auto-Induced Distribution Shift (ADS)}
While training AI systems, we often consider the strengths and weaknesses of the agents themselves only and overlook the impact that these agents have on the environment. Past research often assumed that data is independently and identically distributed \citep{besbes2022beyond}, ignoring the effect of algorithms on data distribution. However, \citet{krueger2020hidden} posited that, in reality, agents could influence the environment during the decision-making and execution process, thus altering the distribution of the data generated by the environment. They referred to this type of issue as ADS.
A real-world example is in recommendation systems, where the content selected by the recommendation algorithms might change users' preferences and behaviors, leading to a shift in user distribution. The distribution shift, in turn, further affects the output of the recommendation algorithms \citep{carroll2022estimating}.
As AI systems increasingly impact the world, we also need to consider the potential further impacts on the data distribution of the entire society after agents are integrated into human society. 

\paragraph{Superficial Alignment}
In recent work, the technique of \textit{Inverse Alignment} in LLMs was introduced by \citet{ji2024language}. The study focused on the elasticity of LLMs, which describes their tendency to revert to a state resembling their original pretrained form after further finetuning post-alignment. This behavior, observed in multiple studies \citep{yang2023shadow,zhou2024lima}, suggests that alignment may not be a permanent change, as models can easily lose their aligned behavior when subjected to new fine-tuning tasks.

The authors formally define \textit{inverse alignment} as follows: Given an initial LLM \( p_{\theta_0} \), after aligning it on a dataset \( D_a \) to produce the aligned model \( p_{\theta_1} \), an operation is performed using a much smaller dataset \( D_b \) (where \( |D_b| \ll |D_a| \)), resulting in an inverse-aligned model \( p_{\theta'_0} \). The goal is to ensure that \( \rho(p_{\theta'_0}, p_{\theta_0}) \leq \epsilon \) for some metric \( \rho \), which measures behavioral or distributional similarity between models. This process is referred to as \textit{inverse alignment}, and the return from \( p_{\theta_1} \) to a model similar to \( p_{\theta_0} \) is the essence of elasticity.
Elasticity is further formalized as the property of LLM parameters returning to a state close to \( p_{\theta_0} \), given an algorithmically simple inverse transformation \( g \) applied to \( p_{\theta_1} \), along with a dataset \( D_b \). The dataset size constraint \( |D_b| \ll |D_a| \) ensures that a small amount of data suffices to reverse the effects of alignment, leading to \( p_{\theta'_0} \) such that \( \rho(p_{\theta'_0}, p_{\theta_0}) \leq \epsilon \).

Theoretical findings show that the normalized compression rate for both the pretraining and fine-tuning datasets decreases upon additional fine-tuning, but the reduction is more pronounced for the fine-tuning dataset by a factor of \( \Theta(k) \), where $k=\frac{\mid D_1 \mid}{\mid D_2\mid}$, $ D_1 $ is the pretraining dataset, and $ D_2 $ is the fine-tuning dataset. This suggests that models are more likely to forget the distribution in the fine-tuning dataset while maintaining their pretraining distribution after exposure to new data. Experimental results further confirm the existence of elasticity in LLMs, showing that they tend to revert to the pretraining distribution with fewer training samples than required during the alignment phase. Larger models, and those with more extensive pretraining data, exhibited greater elasticity, highlighting the limitations of current alignment methods. The theoretical and experimental results indicate that alignment methods are often superficial, and more robust approaches are necessary to ensure the safety of AI systems, particularly in the face of \textit{inverse alignment}. Moreover, these findings underscore the risks associated with open-source models, as they could potentially be reverted to unsafe states with minimal effort, raising concerns about open-source policies in large AI companies.

% Steering Optimization During Training
\subsection{Algorithmic Interventions}
\label{sec:alg-inter}
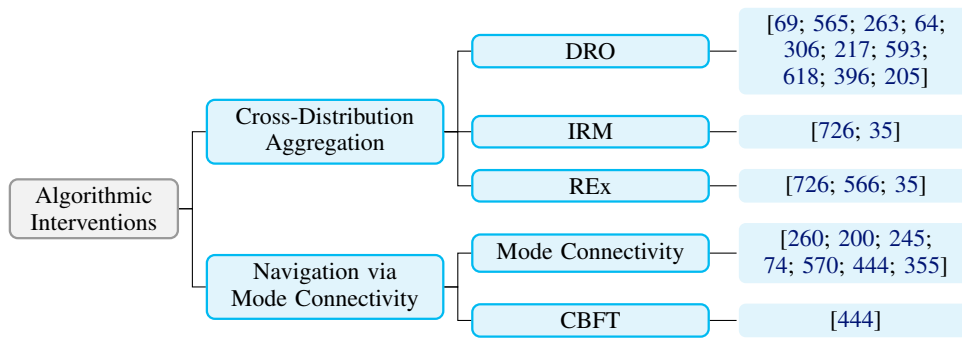
\begin{figure}[t]
\centering
\footnotesize
        \begin{forest}
            for tree={
                forked edges,
                grow'=0,
                draw,
                rounded corners,
                node options={align=center,},
                text width=2.7cm,
                s sep=6pt,
                calign=center,
            },
            [Algorithmic Interventions, fill=gray!45, parent
            % Pre-trained model
                % Template
            [Cross-Distribution Aggregation, for tree={answer,
            calign=child edge, calign child=(n_children()+1)/2,
            yshift=0.05cm,
            }
                    [DRO
                        [\citenumber{ben2009robust, peters2015causal, geirhos2018imagenettrained, beery2018recognition, hendrycks2018benchmarking, engstrom2019exploring, recht2019imagenet, sagawa2020distributionally, krueger2021out, duchi2021statistics}, answer_work]
                    ]
                    [IRM
                        [\citenumber{vapnik1991principles, arjovsky2019invariant}, answer_work]
                    ]
                    [REx
                        [\citenumber{vapnik1991principles, peters2017elements, arjovsky2019invariant}, answer_work]
                    ]
                    ]
            [Navigation via Mode Connectivity, for tree={
            answer,
            calign=center,
            }
                        [Mode Connectivity, for tree={
                        answer,
                        calign=child edge, calign child=(n_children()+1)/2,
                        }
                            [\citenumber{garipov2018loss,draxler2018essentially,frankle2020linear,benton2021loss,pittorino2022deep,lubana2023mechanistic,juneja2022linear}, answer_work]
                        ]
                        [CBFT, for tree={
                        answer,
                        calign=child edge, calign child=(n_children()+1)/2,
                        yshift=-0.38cm,
                        }
                            [\citenumber{lubana2023mechanistic}, answer_work]
                        ]
                ]
            ]
        \end{forest}
            \caption{A tree diagram summarizing the key concepts and literature related to Algorithmic Interventions. The root node represents Algorithmic Interventions that aim to steer optimization during the training process. The main branches represent two main methods, namely cross-distribution aggregation (which aims to minimize risks on different distributions during training to find a predictor based on the invariant relationship instead of spurious features) and navigation via mode connectivity (which aims to fine-tune based on mode connectivity to enhance model generalization performance). Further sub-branches list vital techniques such as Distributionally Robust Optimization (DRO), Invariant Risk Minimization (IRM), Risk Extrapolation (REx), and Connectivity-based Fine-tuning (CBFT).}
\end{figure}

When illustrating the algorithmic intervention methods, we first outline two classes of methods that steer optimization on various distributions during training to relieve distribution shift, {namely, cross-distribution aggregation (\S\ref{sec:cross}) and navigation via mode connectivity (\S\ref{sec:robust-navi})}.

In the first part, we cover methods ranging from the initial approach of  \textit{empirical risk minimization} (ERM) \citep{vapnik1991principles} to \textit{risk extrapolation} (REx) \citep{krueger2021out}, a method conceived to mitigate issues arising from models' dependence on spurious features. In the second part, we introduce \textit{connectivity-based fine-tuning}, which guides the navigation of the loss landscape during training to encourage convergence upon non-spurious correlations, and which does so using insights from \emph{mode connectivity} \citep{lubana2023mechanistic}.

\subsubsection{Cross-Distribution Aggregation}
\label{sec:cross}
One of the main reasons for distribution shift is spurious correlations in the model 
 that are distinct from core objectives \citep{geirhos2018imagenettrained}. By integrating learning information of different domains (or different distributions)  into the optimization objective, we expect the model to learn truthful information and invariant relationships.
In the following paragraphs, we first introduce ERM as the background and then introduce some methods to directly learn how to address distribution shift by integrating loss landscapes of different distributions in the training process.

\paragraph{Empirical Risk Minimization (ERM)}  
Consider a scenario where a model has been developed to identify objects by their features effectively. The optimization target can be expressed as:
\begin{align*}
\mathrm{R} (\bm{w}) = \int \mathrm{L} \big(y, f\left(x, \bm{w}\right)\big) \, d\mathrm{P}\left(x, y\right)
\end{align*}
where $\mathrm{L} (y, f(x, \bm{w}))$ denotes the loss between data labels $y$ and model outputs $f(x, \bm{w})$, while $\mathrm{P}(x, y)$ signifies the target data distribution \citep{vapnik1991principles}.

Nevertheless, a bias often exists between the dataset and the real world, implying that the features learned from the dataset may not necessarily be the ones we intend for the model to acquire. ERM is a strategy employed in statistical methods to optimize this bias. It operates on the assumption that, given the inaccessibility of the real-world target data distribution, the empirical data within the dataset should, ideally, closely approximate this unknown target distribution \citep{vapnik1991principles, zhang2018mixup}. In this context, the objective function is optimized and is redefined as:
\begin{align*}
\mathrm{E} (\bm{w}) = \frac{1}{l} \sum_{i = 1}^l \mathrm{L} \big(y_i, f\left(x_i, \bm{w}\right)\big)
\end{align*}
where $l$ can be different examples in one training distribution or different training distributions.

Minimizing the objective function above allows the model to learn the invariant relationship in different distributions.
Naive ERM makes the naive assumption that the data is sampled from the target data distribution.
However, if a significant discrepancy exists between the source distribution (or training distribution) and the target distribution, severe generalization issues can still arise \citep{szegedy2013intriguing}.

\paragraph{Distributionally Robust Optimization (DRO)}
Numerous studies posit that the sensitivity to distribution shift often arises from reliance on \textit{spurious correlations} or \textit{shortcut features} unrelated to the core concept \citep{geirhos2018imagenettrained,hendrycks2018benchmarking}. For instance, models may judge based on background features rather than employing the correct features within the image \citep{geirhos2018imagenettrained,beery2018recognition}.
Building upon the foundations laid in prior research \citep{ben2009robust,peters2015causal,krueger2021out}, OOD Generalization can be formulated as follows:
\begin{align*}
r_{\mathcal{D}}^{\mathrm{OOD}}(\bm{\theta})=\max _{e \in \mathcal{D}} r_e\left(\bm{\theta}\right)
\end{align*}
This optimization seeks to enhance worst-case performance across a perturbation set, denoted as $\mathcal{D}$, by reducing the maximum value among the risk function set $\left\{ r_e | e \in \mathcal{D} \right\}$. In \textit{Distributionally Robustness Optimization (DRO)} \citep{duchi2021statistics}, the perturbation set covers the mixture of different domains' training distributions, and by minimizing the above objective function, we expect the model can find the invariant relationship between different training distributions. However, it should be noted that naively applying DRO to overparameterized neural networks may lead to suboptimal outcomes \citep{sagawa2020distributionally}. Therefore, combining DRO with increased regularization techniques such as $l_2$ penalty \citep{cortes2009l2} or early stopping \citep{prechelt2002early} can substantially improve generalization performance. For more details on DRO, see \textit{e.g.}, \citet{rahimian2019distributionally,sagawa2020distributionally,lin2022distributionally}

\paragraph{Invariant Risk Minimization (IRM)}
\citet{arjovsky2019invariant} introduces an innovative learning paradigm to estimate nonlinear, invariant, causal predictors across diverse training environments, thereby facilitating robust OOD generalization. IRM aims to train a predictive model with solid performance across various environments while demonstrating reduced susceptibility to relying on spurious features.
IRM can be considered an extension of Invariant Causal Prediction (ICP) \citep{peters2015causal}, which involves hypothesis testing to identify the direct causal features that lead to outcomes within each specific environment instead of indirect features. IRM further extends ICP to scenarios characterized by high-dimensional input data, where variables may lack clear causal significance. The fundamental idea underlying IRM is that when confronted with many functions capable of achieving low empirical loss, selecting a function that exhibits strong performance across all environments is more likely to get a predictor based on causal features rather than spurious ones \citep{pml2Book}.

\paragraph{Risk Extrapolation (REx)}
The basic form of REx involves robust optimization over a perturbation set of extrapolated domains (MM-REx), with an additional penalty imposed on the variance of training risks (V-REx) \citep{krueger2021out}. By reducing training risks and increasing the similarity of training risks, REx forces the model to learn the invariant relationship in different domain distributions.

Amplifying the distributional variations between training domains can diminish risk changes, thereby enforcing the equality of risks. Taking CMNIST \citep{arjovsky2019invariant} as an example, even though establishing a connection between color and labels is more straightforward than connecting logits and labels, increasing the diversity in color can disrupt this \textit{spurious correlations} (or shortcut features) and aid the model in learning the genuine invariant relationship between logits and labels.
Following previous research \citep{vapnik1991principles,peters2017elements,krueger2021out}, REx can be formulated as follows:
Firstly, the Risk Function can be defined as follows:
\begin{align*}
    r_e(\bm{\theta}) \doteq \mathbb{E}_{(x, y) \sim P_e(X, Y)} L\big(f_{\bm{\theta}}(x), y\big)
\end{align*}
where $L(\cdot)$ represents a fixed loss function, and distinct training domains or environments can be formulated as the $P_e(X, Y)$ distribution.
Next, the MM-REx term can be modeled as:
\begin{align*}
r_{\mathrm{MM - REx}}(\bm{\theta})  =  \left(1-m \lambda_{\min }\right) \max _e r_e(\bm{\theta})+\lambda_{\min } \sum_{e=1}^n r_e(\bm{\theta})
\end{align*}
where $n$ represents the number of distinct distributions or domains, and $\lambda_{\min}$ governs the extent of risk extrapolation. Moving on to the V-REx term, it can be modeled as:
\begin{align*}
r_{\mathrm{V}-\mathrm{REx}}(\bm{\theta}) = \alpha \operatorname{Var}\Big(\big\{r_1(\bm{\theta}), \ldots, r_n(\bm{\theta})\big\}\Big)+\sum_{e=1}^n r_e(\bm{\theta})
\end{align*}
where $\alpha \geq 0$ controls the trade-off between risk reduction and enforcing risk equality.

In the MM-REx term, the $\lambda_{\min}$ can set nearly $-\infty$; therefore, the loss of specific domains may be high, meaning that the model may learn the spurious correlations. Minimizing the MM-REx and V-REx can reduce training risks and increase the similarity of training risks, encouraging the model to learn invariant relationships. Furthermore, REx has shown significant promise in experimental settings \citep{krueger2021out}, particularly in causal identification, making it a compelling approach for achieving robust generalization.

\paragraph{Tackle Distribution Shift in LLMs}

In the context of LLMs, prior research has shown that RL often exploits shortcuts to achieve high rewards, overlooking challenging samples \citep{deng2023multilingual}. This evasion of long-tail training samples prevents LLMs from effectively handling distribution shifts in general scenarios, which falls short of expectations for these models: as universal AI assistants, they should maintain consistent performance across various domains. Recently, many works have attempted to implement \textit{cross-distribution aggregation} in LLMs to address this issue.
\citet{zheng2024improving} employ RL to learn uniform strategies across diverse data groups or domains, automatically categorizing data and deliberately maximizing performance variance. This strategy increases the learning capacity for challenging data and avoids over-optimization of simpler data. \citet{yao2024improving} concentrate on exploiting inter-domain connections. Specifically, they acquire training-domain-specific functions during the training phase and adjust their weights based on domain relations in the testing phase, achieving robust OOD generalization.

\subsubsection{Navigation via Mode Connectivity}
\label{sec:robust-navi}
Following the above discussion about cross-distribution aggregation, in this section, we introduce mode connectivity as the prerequisite content. 
Then, we primarily discuss the Connectivity-Based Fine-Tuning (CBFT) \citep{lubana2023mechanistic} method, illustrating how mode connectivity navigates the model to predict based on invariant relationships instead of spurious correlations by changing few parameters.
\paragraph{Mode Connectivity}
\label{par:mode-connectivity}
Mode connectivity refers to the phenomenon where one can identify a straightforward path within the loss function space that connects two or more distinct local minima or patterns \citep{garipov2018loss,draxler2018essentially}. 
In line with prior research \citep{benton2021loss,pittorino2022deep,lubana2023mechanistic}, a formal definition can be defined as follows:

The model's loss on a dataset $\mathcal{D}$ is represented as $\mathcal{L}(f(\mathcal{D}; \bm{\theta}))$, where $\bm{\theta}$ denotes the optimal parameters of the model, and $f(\mathcal{D}; \bm{\theta})$ signifies the model trained on dataset $\mathcal{D}$. We define $\bm{\theta}$ as a minimizer of the loss on this dataset if $\mathcal{L}(f(\mathcal{D}; \bm{\theta}))<\epsilon$, where $\epsilon$ is a small scalar value.

Minimizers ${\bm{\theta}}_1$ and ${\bm{\theta}}_2$, achieved through training on dataset $\mathcal{D}$, are considered to be mode-connected if there exists a continuous path $\gamma$ from ${\bm{\theta}}_1$ to ${\bm{\theta}}_2$ such that, as ${\bm{\theta}}_0$ varies along this path $\gamma$, the following condition is consistently upheld:
\begin{align*}
    \mathcal{L}\big(f\left(\mathcal{D}; {\bm{\theta}}_0\right)\big) \leq t \cdot \mathcal{L}\big(f\left(\mathcal{D}; {\bm{\theta}}_1\right)\big) + \left(1-t\right) \cdot \mathcal{L}\big(f\left(\mathcal{D}; {\bm{\theta}}_2\right)\big), \quad \forall t \in [0,1].
\end{align*}

In essence, mode connectivity entails consistently finding a connecting pathway among minimizers in the parameter space, traversing regions of low loss without delving into regions of highly high loss. This implies that even when making minor adjustments to the model's parameters within the parameter space, the model's performance can remain relatively stable, mitigating significant performance degradation \citep{garipov2018loss}. This concept lays the foundation for designing more effective optimization algorithms, enabling models to share knowledge and experiences across different tasks, enhancing both model performance and generalization capabilities.

Furthermore, we can define two models as mechanistically similar if they employ the same attributes of inputs for making predictions. Some research has demonstrated that the absence of linear connectivity implies mechanistic dissimilarity, suggesting that simple fine-tuning may not suffice to eliminate spurious attributes learned during the pre-training phase \citep{lubana2023mechanistic,juneja2022linear}. However, it is promising to address non-linearly connected regions through fine-tuning, thereby effectively modifying the model's mechanisms to resolve the issue of OOD misgeneralization.

\paragraph{Connectivity-Based Fine-tuning (CBFT)}

As discussed above, recent research has suggested that the absence of linear connectivity between two models implies a fundamental mechanistic dissimilarity. \citet{lubana2023mechanistic} finds that models tend to develop similar inference mechanisms when trained on similar data. This could be a significant reason for the emergence of bias in models, such as relying on the background information of images for classification rather than the objects depicted in the images. If this model mechanism is not adjusted during the finetuning process, the model may rely on these false attributes. To overcome this problem, they propose a valid strategy for altering a model's mechanism, which aims to minimize the following loss:
\begin{align*}
\mathcal{L}_{\mathrm{CBFT}} &= \mathcal{L}_{\mathrm{CE}}\big(f\left(\mathcal{D}_{\mathrm{NC}} ; \bm{\theta}\right), y\big)+\mathcal{L}_{\mathrm{B}}+\frac{1}{K} \mathcal{L}_{\mathrm{I}} 
\end{align*}
where the original training dataset is denoted as $\mathcal{D}$, and we assume that we can obtain a minimal dataset without spurious attribute $C$, denoted as $\mathcal{D}_{\mathrm{NC}}$.

Besides $\mathcal{L}_{\mathrm{CE}}$ that denotes the cross-entropy loss between model's prediction $f\left(\mathcal{D}_{\mathrm{NC}} ; \bm{\theta}\right)$ and the ground truth label $y$, CBFT has two primary objectives:
(1) The first objective entails modifying a model's underlying mechanism by repositioning it within the loss landscape, breaking any linear connection with the current minimizer. This is accomplished by maximizing $\mathcal{L}_{\mathrm{B}}$, referred to as the \textit{barrier loss}.
(2) The second objective involves mitigating reliance on spurious attributes in the original training dataset. This is achieved by optimizing $\mathcal{L}_{I}$, enabling the discovery of invariant relationships without the need for $C$. CBFT holds promise for shifting the mechanism from predicting objectives by spurious features to true features, just changing partial parameters of models.

%: Targeted Expansion of Training Distribution
\subsection{Data Distribution Interventions}\label{sec:data-inter}
Besides algorithmic optimization, methods that expand the distribution of training data to include real-world elements can also reduce the discrepancy between training and deployment distributions. In this section, we specifically focus on the introduction of adversarial pressures and multi-agent dynamics.

\subsubsection{Adversarial Training}\label{sec:adversarial-training}
AI systems can suffer from a lack of adversarial robustness, meaning that certain inputs designed to make them fail cause the models to perform poorly \citep{zheng2016improving}, which has been shown in images \citep{huang2017adversarial} and texts \citep{zou2023universal, shah2023scalable}, as well as changes to semantic features in images \citep{geirhos2018imagenettrained, bhattad2019unrestricted, shamsabadi2020colorfool, casper2022robust} and texts \citep{jia2017adversarial}, and even examples generated entirely from scratch \citep{song2018constructing, ren2020generating, ziegler2022adversarial, chen2024content}. These failure modes are covered in the \emph{red teaming} section (\S\ref{subsec:red teaming}). It's worth noting that in addition to the robustness of AI model policies, the robustness of reward models that govern the training of advanced AI systems is also of importance, as the gradient descent optimization process could be seen as an adversary that may exploit loopholes in the reward model, a phenomenon named \emph{reward model overoptimization} that has been experimentally demonstrated \citep{gao2023scaling}.

We consider adversarial robustness a case of distribution shift failure caused partly by a mismatch between AI systems' training distribution (where the training inputs are not adversarially constructed) and testing distribution (where the example can be adversarially constructed). The method of \emph{adversarial training} \citep{yoo2021towards,tao2021recent,ziegler2022adversarial} mitigates this problem by introducing adversarial examples into training input through a variety of ways \citep{tao2021recent}, thus expanding the training distribution and closing the distribution discrepancy.

Adversarial training, which is similar to adversarial attacks, first started in the settings of image classification \citep{madrylabrobustness}, but later expanded to a wide range of settings. In addition to vision models, adversarial training algorithms have been proposed for language models \citep{wang2019improving,liu2020adversarial,ziegler2022adversarial}, vision-language models \citep{gan2020large,berg2022prompt}, \textit{etc.} In terms of the model type, adversarial training has been applied to classification models \citep{tao2021recent}, generative models \citep{ziegler2022adversarial}, and RL agents \citep{pinto2017robust, tan2020robustifying}.

There are two major types of adversarial training: \emph{perturbation-based} and \emph{unrestricted}.

\begin{itemize}[left=0.3cm]
    \item \textbf{Perturbation-based Adversarial Training}. Mirroring \emph{perturbation-based adversarial attack} (see \S\ref{subsec:red teaming}), perturbation-based adversarial training introduces adversarially perturbated examples (\textit{i.e.}, small changes to a normal data input which are designed to reduce model performance) into training \citep{goodfellow2014explaining}. Techniques in this vein \citep{tao2021recent} include the baseline approach of adding a regularization term into the loss function to assess model performance on a gradient-based perturbated input \citep{goodfellow2014explaining}, unsupervised \citep{carmon2019unlabeled} or self-supervised \citep{hendrycks2019using} approaches, and various supplemental techniques such as the introduction of curriculum learning which gradually intensifies adversarial pressure during training.
    \item \textbf{Unrestricted Adversarial Training}.  Mirroring \emph{unrestricted adversarial attack} (see \S\ref{subsec:red teaming}), unrestricted adversarial training generalizes perturbation-based adversarial training to include \emph{any} adversarial example that can fool the model, not necessarily ones obtained by adding a small amount of noise to another example. This includes \emph{generative adversarial training}, which uses generative models to produce arbitrary adversarial inputs from scratch \citep{poursaeed2021robustness}, and the addition of syntactically or semantically modified adversarial examples to training input \citep{ziegler2022adversarial,mao2022enhance} which surprisingly eliminates the negative effects on the model's non-adversarial performance. Most works on unrestricted adversarial attacks also apply to unrestricted adversarial training (see \S\ref{subsec:red teaming} for an overview) and form an important part of the unrestricted adversarial training methodology.
    
\end{itemize}

\begin{figure}[t]
\centering
\footnotesize
        \begin{forest}
            for tree={
                forked edges,
                grow'=0,
                draw,
                rounded corners,
                node options={align=center,},
                text width=2.7cm,
                s sep=6pt,
                calign=center,
            },
            [Data Distribution Interventions, fill=gray!45, parent
            % Pre-trained model
                % Template
            [Adversarial Training: Incorporating Adversarial Pressures, for tree={
            answer,
            calign=child edge, calign child=(n_children()+1)/2,
            yshift=-0.2cm,
            }
                         [Modalities and Model Types in Adversarial Training
                            [\citenumber{szegedy2013intriguing, goodfellow2014explaining, wang2019improving, liu2020adversarial, gan2020large, tan2020robustifying, tao2021recent, berg2022prompt, ziegler2022adversarial}, answer_work]
                        ]
                        [Perturbation-Based Adversarial Training
                            [\citenumber{goodfellow2014explaining, carmon2019unlabeled, hendrycks2019using, tao2021recent}, answer_work]
                        ]
                        [Unrestricted Adversarial Training
                            [\citenumber{poursaeed2021robustness, mao2022enhance, ziegler2022adversarial}, answer_work]
                        ]
                    ]
            [Cooperative Training: Incorporating Multi-Agent Dynamics, for tree={
            answer,
            calign=child edge, calign child=(n_children()+1)/2,
            }
                        [Fully Cooperative MARL
                            [\citenumber{tan1993multi, guestrin2001multiagent, foerster2016learning, sunehag2018value, lowe2017multi, song2018multi, singh2018learning, jaderberg2019human, cruz2019reinforcement, gronauer2022multi, meta2022human, oroojlooy2023review}, answer_work]
                        ]
                        [Mixed-Motive MARL
                            [\citenumber{gronauer2022multi, jaderberg2019human,lowe2017multi,cruz2019reinforcement,meta2022human, song2018multi,singh2018learning}, answer_work]
                        ]
                        [Zero-Shot Coordination
                            [\citenumber{tobin2017domain, hu2020other, treutlein2021new, cui2021k, hu2021off}, answer_work]
                        ]
                        [Environment-Building
                            [\citenumber{klugl2005role, singh2014norms, cruz2019reinforcement, sun2020scaling, critch2020ai, suo2021trafficsim, leibo2021scalable, wang2021emergent, muglich2022equivariant, ma2022elign, christoffersen2023get, meta2022human, du2023cooperative}, answer_work]
                        ]
             			[Socially Realistic Settings
                              [\citenumber{du2023cooperative,critch2020ai,klugl2005role,suo2021trafficsim,sun2020scaling,singh2014norms}, answer_work]
                        ]
                ]
            ]
        \end{forest}
            \label{fig:target-expansion}
            \caption{A tree diagram summarizing the key concepts and literature related to Data Distribution Interventions. The root node represents Data Distribution Interventions that try to combine multiple distributions during training, for example, adversarial examples and multi-agent interaction. The main branches represent promising methods, namely, Adversarial Training that incorporates adversarial pressures and Cooperative Training that incorporates multi-agent dynamics. Further sub-branches list key techniques such as perturbation-based and unrestricted adversarial training, and cooperative methods also include environment-building, socially realistic settings, zero-shot coordination, and other Multi-Agent Reinforcement Learning (MARL)-based techniques.}
\end{figure}
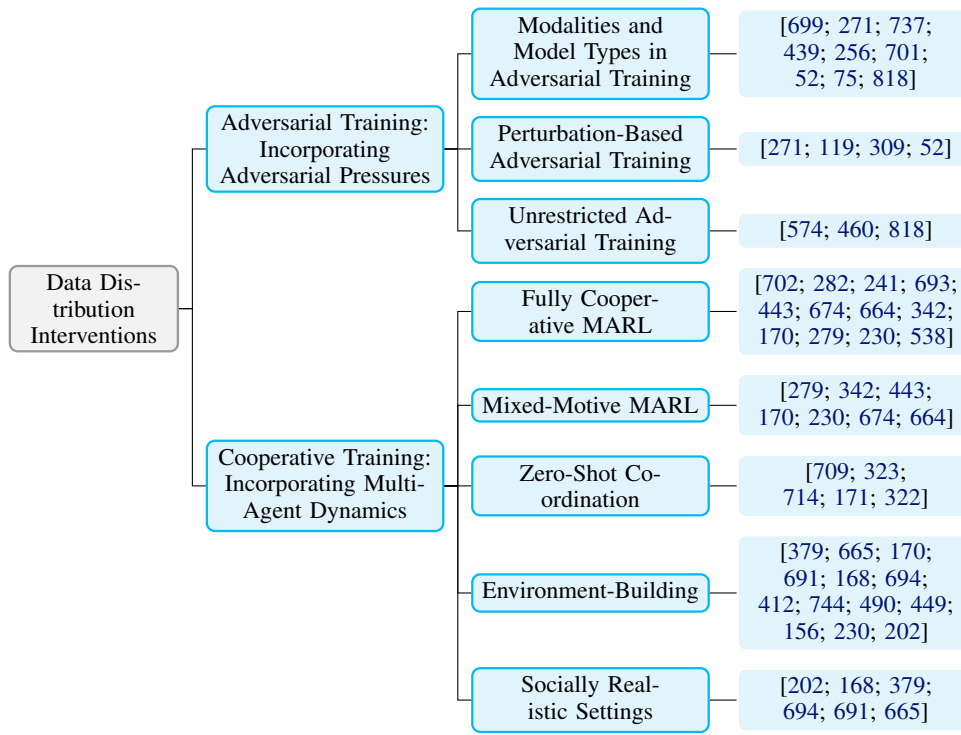

%: Incorporating Multi-Agent Dynamics
\subsubsection{Cooperative Training}\label{sec:cooperative-ai-training}
\emph{Cooperative AI} \citep{dafoe2020open,dafoe2021cooperative} aims to address uncooperative and collectively harmful behaviors from AI systems (see \S\ref{sec:challenges-of-alignment}). The lack of cooperative capabilities in AI systems can be seen as a form of failure under distribution shift -- systems are trained in single-agent settings that are qualitatively different from the real world, which could be massively multi-agent. This difference is indeed a difference in data distribution since the presence of other agents in the environment qualitatively alters the environmental state transition dynamics, leading to changes in the joint distribution of observations and rewards. We approach the problem by expanding our training distribution to include multi-agent interactions via \emph{cooperative training}.

We introduce the branch of cooperative AI (what we call \emph{cooperative training}) that focuses on specific forms of Multi-Agent Reinforcement Learning (MARL) training and complements formal game theory approaches in \S\ref{subsec:formal-ethics-coop}. The MARL branch of cooperative training tends to emphasize the AI system's \emph{capabilities} for coordination (\textit{e.g.}, coordination of a robot football team \citep{ma2022elign}), as opposed to \emph{incentives} of cooperation (\textit{e.g.}, mitigating failure modes like the prisoner's dilemma \citep{phelps2023investigating}) which are the focus of the game theory branch. Here, we only cover the MARL branch due to its relevance to expanding training data distribution.

% Below is a non-exhaustive list of notable directions within cooperative training. 
The field of MARL had traditionally been divided into the three branches of \emph{fully cooperative} (where all agents share the same reward function), \emph{fully competitive} (where the underlying rewards constitute a zero-sum game), and \emph{mixed-motive} settings (where the reward incentives are neither fully cooperative nor fully competitive, corresponding to general-sum games) \citep{gronauer2022multi}. Among them, fully cooperative and mixed-motive settings are the most relevant for cooperative AI, and the latter has been especially emphasized due to its relative neglectedness \citep{dafoe2020open}. We also cover other research fronts, including zero-shot coordination \citep{hu2020other,treutlein2021new}, environment-building \citep{leibo2021scalable}, and socially realistic settings \citep{du2023cooperative}.

\begin{itemize}[left=0.3cm]
\item \textbf{Fully Cooperative MARL}. Fully cooperative settings of MARL are characterized by a shared reward function for all agents \citep{gronauer2022multi}. This unity allows us to completely disregard issues of cooperation \emph{incentives} (since all incentives are perfectly aligned) and instead focus on effectively achieving the shared goal via coordination. Commonly adopted approaches \citep{oroojlooy2023review} lie on a spectrum of centrality -- from the baseline solution of purely independent training \citep{tan1993multi} to the approach of supplementing independent training with decentralized communications \citep{foerster2016learning}, and then to \emph{value factorization} which decomposes a global reward and determine each individual agent's contribution \citep{guestrin2001multiagent,sunehag2018value}.
\item \textbf{Mixed-Motive MARL}. Mixed-motive settings of MARL are characterized by a mixture of cooperative and competitive incentives -- rewards for agents are not identical but aren't zero-sum either \citep{gronauer2022multi}. This includes game environments where teams play against each other \citep{jaderberg2019human} and more nuanced settings such as negotiation \citep{cruz2019reinforcement,meta2022human}. Examples of techniques for mixed-motive MARL, again ordered from decentralized to centralized, include using IRL-like methods to learn from human interactions \citep{song2018multi}, making communications strategic and selective \citep{singh2018learning} and adapting actor-critic methods by granting the critic access to global information \citep{lowe2017multi}. 
\item \textbf{Zero-shot Coordination}. Zero-shot coordination is the goal of making AI systems able to coordinate effecively with other agents (including human agents) without requiring being trained together or otherwise being designed specifically to coordinate with those agents  \citep{hu2020other,treutlein2021new} -- human beings who are complete strangers can still cooperate effectively, and we hope that AI systems can do the same. Early works were published under the name \emph{ad hoc coordination}, covering evaluation \citep{stone2010ad}, game-theoretic and statistical approaches \citep{albrecht2013game}, and human modeling \citep{krafft2016modeling}. Recent advances include \emph{other-play} \citep{hu2020other} which randomizes certain aspects of training partners' policies to achieve robustness,\footnote{This is in a similar spirit to \emph{domain randomization} \citep{tobin2017domain}.} the introduction of multi-level recursive reasoning \citep{cui2021k}, and \emph{off-belief learning} \citep{hu2021off} which eliminates arbitrary conventions in self-play by interpreting partners' past actions as taken by a non-collusive policy.

\item \textbf{Environment-building}. Game environments have been popular settings for cooperative training, including, for example, Hanabi \citep{muglich2022equivariant}, Diplomacy \citep{cruz2019reinforcement,meta2022human}, and football \citep{ma2022elign}. On the more simplistic end, game theory models, especially those based on classical multi-agent dilemmas, have also been a popular choice of environment \citep{wang2021emergent,christoffersen2023get}. Also, Melting Pot \citep{leibo2021scalable}, a framework and suite of multi-agent environments, has been designed specifically for cooperative AI research. There has also been research on \emph{unsupervised environment design}, which aims for a partial automation of the environment-building process \citep{dennis2020emergent,jiang2021replay}.

\item \textbf{Socially Realistic Settings}. It has been proposed that cooperative AI research should focus more on socially realistic environments \citep{du2023cooperative}, which tend to be massively multi-agent (including both AI agents and human agents) and are highly diverse in both the composition of agents and modes of interactions. Implications of this vision \citep{critch2020ai} include, but aren't limited to, building more realistic and open-ended environments \citep{klugl2005role,lehman2008exploiting,wang2019poet,suo2021trafficsim}, scaling up MARL \citep{sun2020scaling,du2023cooperative}, and incorporating new means of control such as social institutions and norms \citep{singh2014norms}.
\end{itemize}

\section{Assurance}
\label{sec:assurance}

\begin{figure}[t]
    \centering
    \includegraphics[width=1.0\textwidth]{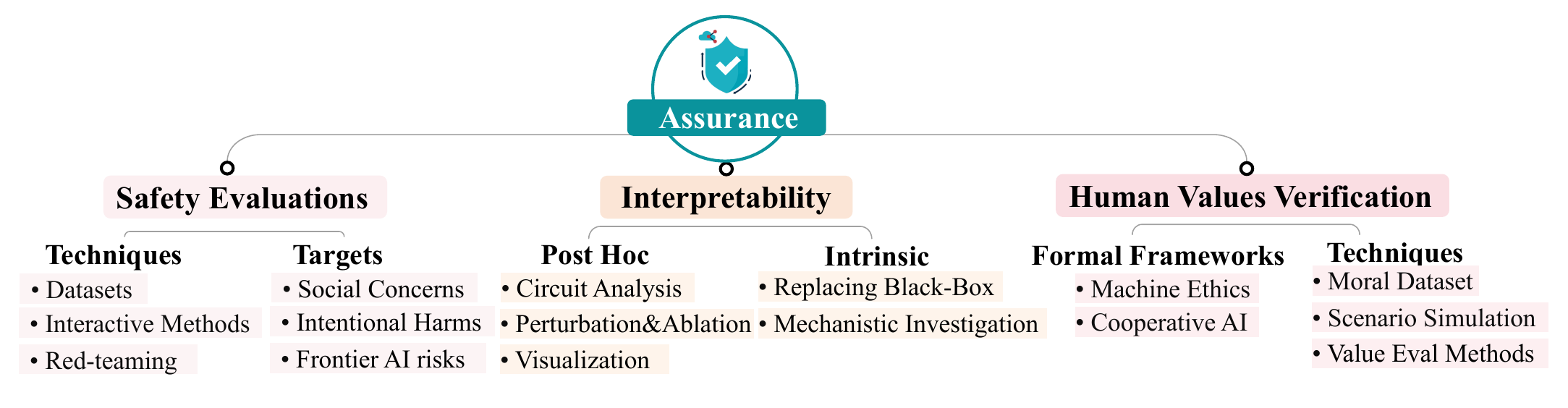}
    \caption{Our organization of research directions, techniques, and applications in assurance. We divide this section into \textit{three} parts: Safety Evaluations -- evaluation of AI systems' safety, which refers to the mitigation of accidents and harmful events caused by the AI system; Interpretability -- making AI systems as well as its decision process more understandable to human beings; Human Value Verification -- verifying whether AI systems can adhere to social and moral norms. The figure also displays the intricate logic of these sections.}
    \label{fig:assurance}
\end{figure}

Assurance refers to the measurement and refinement of AI systems' practical alignment after AI systems are actually trained or deployed \citep{batarseh2021survey}. In this section, we categorize assurance into three parts based on a certain logic: Safety Evaluations -- Evaluating AI systems on minimizing accidents during task execution as a basic need of assurance, Interpretability -- Ensuring that humans can understand the decision-making process of AI systems and therefore assuring the safety and interoperability beyond evaluation, Human Value Verification -- Verifying whether AI systems can align with human values, ethics, and social norms and satisfying the high-level need of AI systems' integration to the human society, as is described in the Figure \ref{fig:assurance}.

In addition to methods that aim to \emph{determine} if AI systems are safe and aligned, there are also assurance methods that actively \emph{intervene} in the AI system or its deployment process to ensure such properties.

\paragraph{Machine Unlearning} Datasets for model pretraining contain various types of undesirable and potentially dangerous content, including but not limited to information about bioweapons and cyberattack \citep{hendrycks2021unsolved}. The field of \emph{machine unlearning} has aimed to remove such knowledge after a model is trained \citep{bourtoule2021machine}. Compared to direct filtering of the training dataset, this approach faces more technical challenges, but it retains more flexibility in deployment and also allows categorical removal of a given piece of information \citep{eldan2023s}. Dataset filtering and unlearning ought to be seens as complementary approaches that work best together.

\paragraph{Controlling Unaligned Systems} While complete alignment may be difficult, it is still possible to safely utilize unaligned models if their extent of misalignment is limited and if we have access to supervisor AI systems. Algorithmic procedures have been developed to minimize probabilities of failure when given trusted and untrusted systems with differing capabilities \citep{greenblatt2023ai}. In general, alignment-focused \emph{process engineering} of deployment procedures could be a valuable direction to explore.

\vspace{0.6em}

In addition, a class of methods, termed \emph{provable safety}, aim to combine evaluation (\S\ref{sec:safe}), interpretability (\S\ref{sec:interpretability}), and other assurance methods under a unified framework that quantifies risks of AI safety violation.

\paragraph{Provable Safety} Provable safety aims to provide formally-grounded probabilistic guarantees on the safety of AI system, using input from evaluation tools, interpretability tools, and other assurance techniques; furthermore, it hopes to build development-deployment pipelines that satisfy such probabilistic guarantees \citep{tegmark2023provably,dalrymple2024towards}.
Research around provable AI safety is still at an early stage, and significant uncertainties remain about its specifics. \citet{dalrymple2024safeguarded} have made the case that three key subproblems currently exist within provable AI safety: \emph{scaffolding}, \emph{i.e.}, formalizing the definition of real-world AI safety, by providing tools for domain experts to build safety specifications; \emph{machine learning}, \emph{i.e.}, using ML methods to find control policies that satisfy the safety specifications; and \emph{applications}, \emph{i.e.}, demonstrating the practical superiority of AI systems with safety gurantees other traditional ones.

\vspace{0.6em}

We then go on two review the three categories of alignment assurance efforts.
 
\subsection{Safety Evaluations}
\label{sec:safe}

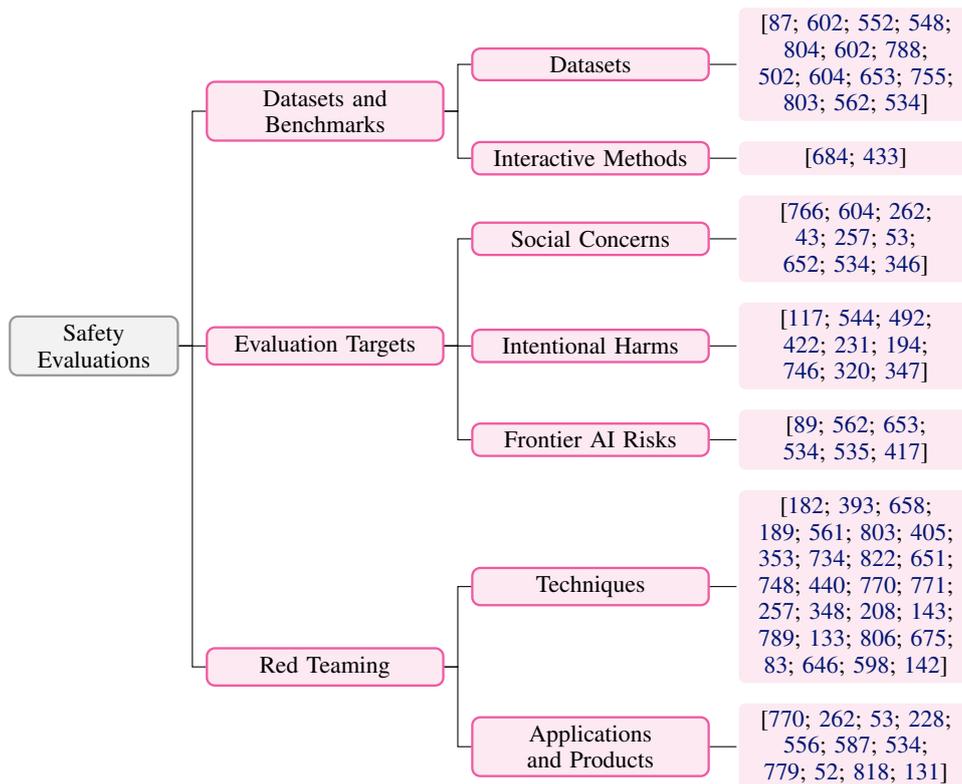
\begin{figure}[ht]
\centering
\footnotesize
        \begin{forest}
            for tree={
                forked edges,
                grow'=0,
                draw,
                rounded corners,
                node options={align=center,},
                text width=2.7cm,
                s sep=6pt,
                tuning,
                calign=child edge, calign child=(n_children()+1)/2,
            },
            [Safety Evaluations, fill=gray!45, parent
                [Datasets and Benchmarks, for tree={
                    tuning,
                    calign=center,
                    },
                    [Datasets, for tree={
                    calign=child edge, calign child=(n_children()+1)/2,
                    yshift=-0.35cm,
                }
                        [\citenumber{bolukbasi2016man, roh2019survey, parrish2022bbq, papernot2016towards, zhao2018gender, roh2019survey, zampieri2019predicting, nangia2020crows, rosenthal2021solid, shevlane2023model, weston2015towards, zhang2022constructing, perez2022discovering, openai2023gpt4}, tuning_work]
                    ]
                    [Interactive Methods, for tree={
                    calign=child edge, calign child=(n_children()+1)/2,}
                        [\citenumber{stiennon2020learning, lin2023llm}, tuning_work]
                    ]
                ]
                [Evaluation Targets, for tree={
                    tuning,
                    calign=child edge, calign child=(n_children()+1)/2,
                    },
                    [Social Concerns
                        [\citenumber{wulczyn2017ex, rosenthal2021solid, gehman2020realtoxicityprompts, askell2021general, ganguli2022red, bai2022training, sheth2022defining, openai2023gpt4, ji2024beavertails,dai2024safesora}, tuning_work]
                    ]
                    [Intentional Harms
                        [\citenumber{carlsmith2022power, pan2023machiavelli, munir2022situational, li2022emergent, falke2019ranking, dhingra2019handling, wang2020towards, honovich2021q2, ji2023survey}, tuning_work]
                    ]
                    [Frontier AI Risks
                        [\citenumber{bostrom2012superintelligent, perez2022discovering, shevlane2023model, openai2023gpt4, openai2023gpt4V, lentzos2022ai}, tuning_work]
                    ]
                ]
                [Red Teaming,  for tree={
                    tuning,
                    calign=center,
                    },
                    [Techniques, for tree={calign=child edge, calign child=(n_children()+1)/2,
                    yshift=0.15cm,
                    }
                        [\citenumber{dathathri2019plug, krause2021gedi, si2022so, deng2022rlprompt, perez2022red, zhang2022constructing, lee2022query, jones2023automatically, wallace2019universal, zou2023universal, shen2023anything, wei2023jailbroken, liu2023jailbreaking, xu2020recipes, xu2021bot, ganguli2022red, jia2017adversarial, ebrahimi2018hotflip, cheng2020seq2sick, zang2020word, chakraborty2021survey, zhao2024evaluating, song2018constructing, bhattad2019unrestricted, shamsabadi2020colorfool, ren2020generating, chen2024content}, tuning_work]
                    ]
                    [Applications and Products, for tree={
                    calign=child edge, calign child=(n_children()+1)/2,}
                        [\citenumber{xu2020recipes, gehman2020realtoxicityprompts, bai2022training, google_ai_red_team, nvidia_ai_red_team, microsoft_ai_red_team, openai2023gpt4, yoo2021towards, tao2021recent, ziegler2022adversarial, casper2022robust}, tuning_work]
                    ]
                ]
            ]
        \end{forest}
            \caption{A Tree diagram summarizing the key concepts, logic, and literature related to Safety Evaluation. The root of the tree represents Safety Evaluation, which aims to \textit{measure the accidents caused by design flaws in AI systems and harmful events that deviate from the intended design purpose of the AI system}. The main branches represent the main structure of safety evaluation, including Datasets and Benchmarks, Evaluation Targets, and Red Teaming techniques. Further sub-branches list key works exploring each of these branches. This diagram provides an overview of research directions and specific techniques for measuring AI systems' safety alignment degree.}
\end{figure}

Safety refers to mitigating accidents caused by design flaws in AI systems and preventing harmful events that deviate from the intended design purpose of the AI system \citep{amodei2016concrete}. In fact, safety stands as a shared requirement across all engineering domains \citep{verma2010reliability}. Moreover, it holds particular importance in constructing AI systems, because of the characteristics of AI systems \citep{jacob2015longterm}. We categorize the safety of AI systems into the following categories: \textit{Social Concerns} refer to explicit and comparatively identifiable characteristics of safe AI systems, including aspects such as toxicity \citep{stahl2022assessing}, and \textit{Intentional Behaviors} share the characterization of relatively complicated investigation and substantial potential harm, represented by power-seeking, deception, and other frontier AI risks \citep{shevlane2023model}.

Following the logic above, we start with the techniques to form datasets and benchmarks of safety evaluation in \S\ref{subsec: dataset&benchmark} and further explore the evaluation targets and their characteristics in \S\ref{subsec: evaltarget}. At the end of this section, we include the red-teaming technique \S\ref{subsec:red teaming}, which assesses the AI system's robustness beyond evaluation.

\subsubsection{Datasets and Benchmarks}
\label{subsec: dataset&benchmark}

In the discussions on safety evaluation, it is crucial to prioritize datasets and benchmarks as the cornerstone elements, so we first introduce the basic techniques to build datasets and benchmarks and then move on to newer interactive methods.

\paragraph{Dataset} Among all the assurance techniques, the dataset method could be considered the most elementary and straightforward one \citep{celikyilmaz2020evaluation}. This method assesses the response of AI systems by presenting them with predefined contexts and tasks \citep{paullada2021data}, balancing the cost, quality, and quantity of data. Research on the dataset method encompasses data sources, annotation approaches, and evaluation metrics. Given that evaluation metrics can vary based on its subject \citep{sai2022survey}, this section primarily emphasizes dataset sources and annotation methods.

\begin{itemize}[left=0.3cm]
    \item \textbf{Expert design}. In the early stage of a domain, expert design is widely used in building datasets, where experts create samples based on actual needs to ensure the dataset covers a wide range of potentially dangerous situations to form datasets \citep{roh2019survey}. For instance, initial-stage datasets, \textit{e.g.}, WEAT \citep{bolukbasi2016man} and BBQ \citep{parrish2022bbq} for bias detection used expert design to harvest a wide coverage and high accuracy while sharing the limitations in terms of cost and breadth, leading to the later development of more efficient methods.
    
    \item \textbf{Internet collection}. Previous expert design methods have the flaw of rather high cost and lower efficiency, and internet collection can obtain datasets that contain actual user-generated textual content on a rather large scale (therefore convenient for both training and testing), reflecting real-world text generation scenarios \citep{yuen2011survey}, but the raw data collected also needs careful selection and annotation \citep{roh2019survey}. Well-known instances of these datasets include OLID \citep{zampieri2019predicting} and SOLID \citep{rosenthal2021solid} gathering original Twitter texts for toxicity assessment, WinoBias \citep{zhao2018gender} and CrowS-Pairs \citep{nangia2020crows} gather content potentially containing bias from the internet for further annotation. However, it's important to acknowledge that, as is also mentioned in \citet{papernot2016towards}, internet-collected datasets naturally carry risks such as privacy and safety concerns, so additional processing is necessary.
    
    \item \textbf{AI Generation}. The concept of autonomously generating datasets was explored relatively early, even before the emergence of elementary forms of LLMs \citep{weston2015towards}. However, during this early stage, AI-generated datasets were limited by the capabilities of AI systems, so their quality was not as good as internet-collected and manually annotated datasets. It wasn't until LLMs reached relatively high levels of proficiency in logical reasoning context understanding and approached or surpassed human-level performance \citep{openai2023gpt4} that LMs gained the ability to mimic the structure and logic of existing datasets to compose new ones. As is shown in papers such as \citet{zhang2022constructing} and \citet{perez2022discovering}, AI systems have made progress in generating datasets for evaluation purposes, surpassing the quality of some classical datasets. However, according to these papers, this approach still faces limitations rooted in the capabilities of large models themselves, including issues like instruction misunderstanding and example diversity, which require further refinement.
\end{itemize}

\paragraph{Interactive Methods} Due to the static nature of datasets, they possess relatively fixed evaluation content and can be vulnerable to targeted training \citep{holtzman2019curious}. Additionally, the evaluation content may not fully reflect the strengths and weaknesses of corresponding capabilities \citep{engstrom2020identifying}. As the demands for language model evaluation continue to escalate, new interactive assurance methods have emerged, which can be categorized into two groups: Agent as Supervisor and Environment Interaction.

\begin{itemize}[left=0.3cm]
    \item \textbf{Agent as Supervisor}. It is an assurance method that involves using an agent to assess the outputs of AI models. This evaluation approach is characterized by its dynamism and flexibility. Typically, there is a predefined framework for interaction between the agent and the AI system under evaluation \citep{cabrera2023zeno}. In this method, the agent can be a human participant engaged in experiments through an online system \citep{stiennon2020learning}, a more advanced language model evaluating relatively less capable language models through multi-turn interactions \citep{lin2023llm}, or in the context of \textit{Scalable Oversight}, a less powerful but more trustworthy model \citep{greenblatt2023ai}. This evaluation form offers advantages such as automation and lower cost compared to human agents.

    \item \textbf{Environment Interaction}. It aims to create a relatively realistic environment using elements such as humans and other LLMs to assess the alignment quality of AI models through multiple rounds of interaction \citep{liu2024agentbench}. One method is using peer discussions, where multiple LLMs engage in dialogue, to enhance evaluations of AI systems, particularly when their capabilities are relatively close to each other. Moreover, by building a world model \citep{li2022emergent}, the generalization and exploration abilities of AI systems can be comprehensively evaluated.
\end{itemize}

\subsubsection{Evaluation Targets}
\label{subsec: evaltarget}

 To achieve the goal of safety alignment, the assurance of AI systems can be divided into different small targets \citep{shevlane2023model}. The subsequent section gives an introduction to these subjects and, furthermore, discusses some of the domain-specific analyses of assurance methods within these realms, while the table \ref{tab:safety-eval-example} will show examples of alignment assurance works in these domains.

\begin{table}[t]
\centering
\caption{A Chart of Safety Evaluation Examples: Specific dataset works are listed in this chart, along with their detailed information: \textit{evaluation targets}, \textit{first release time}, \textit{most recent update time} (we list them separately because some datasets are consistently being updated), \textit{information quantity} (the sum of the information form unit), \textit{institution}, \textit{information form}, \textit{baseline model} and \textit{information source}. Moreover, to contain more information, we made some abbreviations in the chart: We shortened the release time and recent updates by concatenating the last two digits of the year and the month and only taking the institution of the paper's first author, and we use combinations of uppercase letters to replace long words in information form: SP for Sentence Pairs, SL for Sentence-Label, ST for sentence template, PP for pronoun pairs, and SS for single selections.}
\label{tab:safety-eval-example}
\footnotesize
\renewcommand\tabcolsep{2.5pt}
\resizebox{\textwidth}{!}{
\begin{tabular}{lllcccccccc}
\toprule \\
\multirow{-2}{*}{}    & 
\multirow{-2}{*}{\textbf{Dataset}} & 
\multirow{-2}{*}{\textbf{\begin{tabular}[c]{@{}c@{}}Release\\ Time \end{tabular}}} & 
\multirow{-2}{*}{\textbf{\begin{tabular}[c]{@{}c@{}}Recent\\ Update\end{tabular}}} & 
\multirow{-2}{*}{\textbf{\begin{tabular}[c]{@{}c@{}}Info\\ Quantity\end{tabular}}}   &
\multirow{-2}{*}{\textbf{\begin{tabular}[c]{@{}c@{}}Institution\end{tabular}}}  & 
\multirow{-2}{*}{\textbf{\begin{tabular}[c]{@{}c@{}}Information\\ Form\end{tabular}}} & 
\multirow{-2}{*}{\textbf{\begin{tabular}[c]{@{}c@{}}Baseline\\ Model\end{tabular}}} & 
\multirow{-2}{*}{\textbf{\begin{tabular}[c]{@{}c@{}}Infomation\\ Source\end{tabular}}} \\
\midrule
  & Aequitas 
  ~\citenumber{saleiro2018aequitas}    
  & 18/05 & 23/04 & - & U.Chicago & Python   & - & {\begin{tabular}[c]{@{}c@{}}Self Build \end{tabular}}   \\
  & WinoS
  ~\citenumber{rudinger2018gender}   
  & 18/10 & 19/01 & 0.72K & JHU   & ST  & Rule\&Neural & Self Build  \\
  & EEC 
  ~\citenumber{kiritchenko2018examining}
  & 18/05 & - & 8K  & NRC Canada & SP   & SVM & Selection  \\
  & GAP 
  ~\citenumber{webster2018mind} 
  & 18/05 & - & 8.9K & Google & PP   & Transformer & Wikipedia  \\
  & OLID 
  ~\citenumber{zampieri2019predicting}
  & 19/05 & - & 14K & U.Wolver.   & SL   & SVM\&LSTM & Twitter  \\
  & CrowS-Pairs 
  ~\citenumber{nangia2020crows}
  & 20/03 & 21/10 & 1.5K & NYU   & SP & BERT & MTurk \\
  & StereoSet
  ~\citenumber{nadeem2021stereoset}
  & 20/04 & 22/04  & 17K & MIT   & SS & BERT\&GPT-2 & MTurk  \\
  & BBQ
  ~\citenumber{parrish2022bbq}
  & 21/05 & 22/07 & 58.5K & NYU   & SS & Multiple LLMs & MTurk  \\
  & LM-Bias
  ~\citenumber{liang2021towards}
  & 21/07 & 22/01 &16K & CMU   & QA Pair & GPT-2 & Corpus Select \\
\multirow{-9}{*}{\begin{tabular}[c]{@{}c@{}}Bias\end{tabular}}  
  & VQA-CE 
  ~\citenumber{dancette2021beyond}   
  & 21/03 & 21/10 & 63K & Sorbonne  & Multimodal & -  & Self-Build  \\
  & AuAI
  ~\citenumber{landers2023auditing}
  & 23/01 & - & - & Sorbonne  & Framework & - & Self Build\\
\midrule
  & WCC
  ~\citenumber{wulczyn2017ex}   
  & 16/01 & - & 63M & Wikimedia & SL & Human  & Wikipedia      \\
  & RTP
  ~\citenumber{gehman2020realtoxicityprompts}   
  & 19/10 & 21/04 & 100K & UW & Prompt & GPT-2  & Refinement      \\
  & SOLID
  ~\citenumber{rosenthal2021solid}   
  & 20/05 & - & 9M & IBM & SL & BERT  & Twitter    \\
  & Toxigen
  ~\citenumber{hartvigsen2022toxigen}
  & 20/05  & 23/06 & 274K & MIT & SL & GPT-3  & GPT Gen.       \\
  & HH-RLHF
  ~\citenumber{bai2022training}
  & 22/04  & 22/09 & 162K & Anthropic & SP & Claude  & Corpus Refine    \\
\multirow{-7}{*}{\begin{tabular}[l]{@{}l@{}}Toxicity\end{tabular}} 
  & BeaverTails
  ~\citenumber{ji2024beavertails} 
  & 23/06 & 23/07 & 30K & PKU  & QA Pair & Multiple LLMs & Corpus Refine  \\
\midrule
  \multirow{-1}{*}{\begin{tabular}[l]{@{}l@{}}Power\\ Seeking\end{tabular}}
  & MACHIAVELLI
  ~\citenumber{pan2023machiavelli}
  & 23/04 & 23/06 & 134 & UCB  & Games & GPT-4\&RL  & Selection  \\
  & BeaverTails
  ~\citenumber{ji2024beavertails} 
  & 23/06 & 23/07 & 30K & PKU  & QA Pair & Multiple LLMs & Corpus Refine   \\
\midrule
  \multirow{-1}{*}{\begin{tabular}[l]{@{}l@{}}Situation\\Awareness\end{tabular}} 
 
  & SA Framework
  ~\citenumber{sanneman2020situation}
  & 20/07 & - & - & MIT  & Framework & -  & Self Build      \\
  & EWR
  \citenumber{li2022emergent}
  & - & - & 10 & Havard  & Game & Othello GPT  & Self Build   \\
\midrule
  & PARENT
  ~\citenumber{dhingra2019handling}
  & 19/06 & - & - & CMU  & Metric & -  & Self Build    \\
  \multirow{-1}{*}{\begin{tabular}[l]{@{}l@{}}Hallucination\end{tabular}}
  
  & PARENT-T
  ~\citenumber{wang2020towards}
  & 20/05 & - & - & NYU  & Metric & -  & Self Build   \\
  & ChatGPT-Eval
  ~\citenumber{bang2023multitask}
  & 23/02 & 23/03 & - & HKUST  & Multimodal & ChatGPT  & Integration \\
  & POPE
  ~\citenumber{li2023evaluating}
  & 23/05 & 23/08 & 2K & RUC  & Multimodal & Multiple LVLMs  & Dataset Refine  \\

\bottomrule
\end{tabular}
}
\end{table}

\paragraph{Toxicity} It refers to content in the output of AI systems that is unhelpful or harmful to humans \citep{sheth2022defining}. Before the advent of advanced language models, early toxicity evaluation primarily focused on detecting toxic language and identifying harmful statements in an internet context, like the WCC \citep{wulczyn2017ex}, which collected and manually labeled comments from Wikipedia discussion pages. With the emergence of pretrained language models, assurance against toxicity adopted a prompt-generation paradigm to assess the risk of language models generating toxic content in response to specific prompts \citep{gehman2020realtoxicityprompts, ganguli2022red, openai2023gpt4}. However, in crowdsourced environments, annotation scores may vary by person, so relative labeling, where crowdsource workers select from two different answers during a chat, is needed to enhance crowdsource quality ~\citep{bai2022training}. Furthermore, subsequent datasets \citep{ganguli2022red, ji2024beavertails} employ a red teaming design pattern that induces toxic responses through adversarial inputs, further strengthening the assurance of model robustness. The SafeSora dataset \citep{dai2024safesora} is the first text-to-video preference dataset designed to align large vision models (LVMs) with human values, focusing on helpfulness and harmlessness.

\paragraph{Power-seeking} It is a kind of risk that AI systems may seek power over humans once they possess certain levels of intelligence \citep{turner2021optimal}. In \citet{carlsmith2022power}, the authors point out that AI systems already have the conditions for power-seeking, including advanced capabilities, agentic planning, and strategic awareness. However, the assurance against power-seeking is still in its early stages. One representative work in this area is the Machiavelli \citep{pan2023machiavelli}, which constructs a benchmark consisting of decision-making games to assess whether AI systems can balance competition with moral ethics during the game. The conclusion of this work suggests that AI systems still struggle to balance achieving rewards with behaving morally, thus further research in this field is needed.

\paragraph{Deceptive Alignment} When the AI system is situationally aware, it may recognize that getting high rewards can preserve themselves by preventing significant gradient descent, therefore preserving its original goal \citep{hubinger2019risks, kenton2021alignment, ngo2024the}. This process is called \textit{Deceptive Alignment}. In the current context, deceptive alignment is already achievable, as is proved by \citep{hubinger2024sleeper}. Directly evaluating the deceptive alignment is difficult, for the pronoun \textit{deceptive alignment} is naturally against the traditional train-evaluation loop. Thus, deceptive alignment might be discovered by indirect methods such as interpreting model parameters (see \ref{sec:interpretability}), or representation engineering \citep{zou2023representation}. 

Moreover, deceptive alignment is closely related to \textit{situational awareness}, \textit{i.e.}, AI systems with a certain degree of prediction and understanding of the states and developments of entities in their working environment to make corresponding decisions. In \citep{li2022emergent}, the authors evaluate the performance of language models in the board game Othello, showing that language models have the ability to predict possible future states within the action space in a nonlinear representation.

\paragraph{Hallucination} AI systems may generate information or responses that are not grounded in factual knowledge or data, leading to the creation of misleading or false content, which is formally called Hallucination \citep{ji2023survey}. Hallucination evaluation aims to assure the consistency of the knowledge in the AI system's output with the knowledge given by its training data and knowledge base \citep{ji2023survey, zhang2023siren}. The earliest statistical-based hallucination evaluation methods used n-grams to directly calculate the overlap of vocabulary between the input and output content ~\citep{dhingra2019handling, wang2020towards}. However, this type of evaluation has a limitation: It only considers lexical overlap and does not take into account semantics or sentence meaning \citep{ji2023survey}, making it unsuitable for evaluating more complex forms of hallucination. Later assurance methods shifted from statistical approaches to model-based methods, which are more robust compared to statistical token-difference-based methods \citep{honovich2021q2}. While this evaluation method is more advanced than previous ones, it still has the limitation that the model can only output the degree of hallucination and may have difficulty pinpointing specific errors \citep{falke2019ranking}. \citet{hu2024do} designed the benchmark Pinocchio, which investigates whether LLMs can integrate multiple facts, update factual knowledge over time, and withstand adversarial examples. This benchmark represents an in-depth investigation into the factual knowledge bottleneck under the issue of hallucination.

\paragraph{Frontier AI Risks} 
In addition to the assurance content described above, the enhancement of AI systems in recent years has given rise to a series of new assurance needs \citep{openai2023gpt4}. Currently, there is not much public information available for research on these assurance needs, hence this section will provide a brief introduction to some of the more significant ones:
\begin{itemize}[left=0.3cm]
    \item \textbf{Cyber Security \& Biological Weapons}. Advanced LLMs may be misused for cyber-attacks, the production of bio-weapons, and other extremely harmful behaviors \citep{shevlane2023model}. Although GPT-4 cannot play a significant role in exploiting network vulnerabilities due to its limited context window, it has been proven to demonstrate strong capabilities in identifying network vulnerabilities and in social engineering \citep{openai2023gpt4}. Similarly, \citet{lentzos2022ai} have stated the robust abilities of AI systems in the field of bio-weapons and the military, highlighting the risks of misuse of such capabilities. It emphasizes the necessity to ensure that these models can identify and reject malicious requests.
    \item \textbf{Deception \& Manipulation}. AI systems have the potential to negatively influence users by outputting text, including disseminating false information, syncopating humans, and shaping people's beliefs and political impacts \citep{shevlane2023model,sharma2024towards}. Distinguished from hallucination, the misinformation here is not a flaw of the model itself but rather a deliberate action. Special assurance measures need to be designed for controlling these kinds of behavior. 
    \item \textbf{Jailbreak}. It refers to the bypassing of AI systems' safeguard mechanisms by users, for example, by constructing specific types of input. This behavior can be limited to text \citep{openai2023gpt4, deng2023jailbreaker, huang2024catastrophic, yong2023low},\footnote{Relevant discussions in \citet{openai2023gpt4} can be found in its \textit{system card} appendix. \label{footnote-gpt4}} or it may take multi-modal forms \citep{openai2023gpt4V}. Specifically, multi-modal jailbreaks make traditional text-based heuristic methods for identifying attack content infeasible, necessitating special multi-modal handling methods. Further discussion of jailbreak can be found in \S\ref{subsec:red teaming}.
    \item \textbf{Self-Preservation \& Proliferation}. This refers to the tendency of AI systems for self-protection and replication, and in this process, breaking the limit from their environment. These tendencies are examples of \emph{instrumental sub-goals} \citep{bostrom2012superintelligent}. While this tendency can be beneficially harnessed, it is dangerous in the absence of regulation \citep{perez2022discovering}. This tendency has been emphasized and evaluated by various sources \citep{perez2022discovering,kinniment2023evaluating,openai2023gpt4,openai2023gpt4V}.\textsuperscript{\ref{footnote-gpt4}}
\end{itemize}

\subsubsection{Red Teaming}
\label{subsec:red teaming}

\textit{Red teaming} is the act of generating scenarios where AI systems are induced to give unaligned outputs or actions (\textit{e.g.}, dangerous behaviors such as deception or power-seeking, and other problems such as toxic or biased outputs) and testing the systems in these scenarios. The aim is to assess the robustness of a system's alignment by applying adversarial pressures, \textit{i.e.} specifically trying to make the system fail. In general, state-of-the-art systems -- including language models and vision models -- do not pass this test \citep{perez2022red,zou2023universal,liu2023jailbreaking,chen2024content}.

In game theory and other fields, red teaming was introduced much earlier, and within computer science, the concept of red teaming was proposed in the security field, where it had a similar meaning of adversarially assessing the reliability and robustness of the system. Later, \citet{ganguli2022red,perez2022red} introduced this idea to the field of AI, and more specifically, alignment.

The motivation for red teaming is two-fold: (1) to gain assurance on the trained system's alignment, and (2) to provide a source of adversarial input for adversarial training \citep{yoo2021towards,tao2021recent, ziegler2022adversarial}, probing models \citep{kalin2020black}, and further utilities. Here, we focus on the first. It's worth noting that the two objectives aren't separable; works targeting the first motivation also help provide a basis for the second.

\paragraph{Reinforced, Optimized, Guided, or Reverse Context Generation}
This category includes using various methods to generate coherent contexts (prompts) that are inducive to unaligned completions from the language model. \citet{perez2022red, deng2022rlprompt, casper2023explore} train or tune a separate language model with RL to make it generate desired prompts, which are then fed to the red-teamed model. \citet{perez2022red, si2022so} also uses other methods such as
zero-shot, few-shot, or supervised finetuning-based generation. \citet{lee2022query, jones2023automatically} generates misalignment-inducive contexts by performing optimization on the prompt -- bayesian optimization and discrete optimization, respectively. \citet{dathathri2019plug,krause2021gedi} propose the method of guiding an LLM's generation using a smaller classifier; this is proposed in detoxification but is transferable to the red teaming context. Lastly, \citet{zhang2022constructing} generates misalignment-inducive contexts through \textit{reverse generation}, \textit{i.e.} \textit{constructing adversarial contexts conditioned on a given response}, which can be seen as an inverse process for model inference.

\paragraph{Manual and Automatic Jailbreaking}
As is defined above \ref{subsec: evaltarget}, \textit{Jailbreaking} \citep{shen2023anything} is an informal term that refers to the act of bypassing a product's constraints on users -- and in the case of LLMs, bypassing LLMs' tendencies to not answer misalignment-inducive questions, a feat of alignment training. Most existing attempts are scattered across the Internet in the form of informal reports and involve adding prefixes and suffixes to the original text \citep{zou2023universal}. Research has descriptively analyzed the existing attempts \citep{liu2023jailbreaking,shen2023anything, deng2023jailbreaker, huang2024catastrophic}, as well as providing causal explanations for the phenomenon \citep{wei2023jailbroken}. In addition, past \citep{wallace2019universal} and current \citep{zou2023universal, shah2023scalable} works have proposed effective methods to automatically generate such prompts, prefixes, or suffixes that nullify LLMs' tendencies to avoid misalignment-inducive questions.

\paragraph{Crowdsourced Adversarial Inputs}
Several works \citep{xu2020recipes, xu2021bot, ganguli2022red} have produced misalignment-inducive prompts by crowdsourcing, \textit{i.e.} recruiting human red teamers (possibly via online platforms) and instruct them to provide adversarial prompts. Besides, companies in the AI industry also build mechanisms to collect adversarial inputs, \textit{i.e.} the red teaming network of OpenAI\footnote{\url{https://openai.com/blog/red-teaming-network}} and the bug hunter program of Google\footnote{\url{https://bughunters.google.com/about/rules/6625378258649088}}. These methods (arguably) provide more flexibility and resemblance to real-world use cases but have higher costs and lower scalability.

\paragraph{Perturbation-Based Adversarial Attack}
In the field of computer vision, there have been many works studying adversarial attacks on vision models that rest on the method of \textit{perturbation}, \textit{i.e.}, performing small perturbations to the pixel contexts of the image (usually bounded by a pixel-wise matrix norm) to make the model confidently produce false outputs on the perturbated image \citep{chakraborty2021survey}. This type of adversarial attack has also been extended to language models \citep{jia2017adversarial,ebrahimi2018hotflip,zang2020word,cheng2020seq2sick} and vision-language models \citep{zhao2024evaluating}.

\paragraph{Unrestricted Adversarial Attack}
\textit{Unrestricted adversarial attack}, proposed in \citep{song2018constructing}, is a more general form of adversarial attack. It removes all restrictions on the adversarial examples, and therefore, for instance, the adversarial example can be generated from scratch, as opposed to being generated from an existing example, as in the case of perturbation-based methods. Many methods for unrestricted adversarial attack have been proposed; the most notable ones include \citep{song2018constructing,chen2024content} which generate realistic adversarial images using generative models, and \citep{bhattad2019unrestricted,shamsabadi2020colorfool} which manipulates semantically meaningful traits such as color and texture. Unrestricted adversarial attack has also been extended to text classification models \citep{ren2020generating}.

\paragraph{Datasets for Red Teaming}
A number of works on red teaming and related topics have compiled datasets consisting of red teaming prompts or dialogues, including the IMAGENET-A and IMAGENET-O dataset \citep{hendrycks2021natural}, the BAD dataset \citep{xu2020recipes}, the red teaming section of HH-RLHF dataset \citep{bai2022training}, and the Real Toxicity Prompts dataset \citep{gehman2020realtoxicityprompts}.

\paragraph{Existing Red Teaming Practices in Industry}

The practice of red teaming is gaining popularity in the AI industry. Cases of adoption include OpenAI (who performed red teaming on its system GPT-4 to produce part of its System Card) \citep{openai2023gpt4}, NVIDIA \citep{nvidia_ai_red_team}, Google \citep{google_ai_red_team}, and Microsoft \citep{microsoft_ai_red_team}. During an event at the DEF CON 31 conference, models from 9 companies undergo red teaming from the conference 
participants;\footnote{\url{https://www.airedteam.org/}} this red teaming event is held in partnership with four institutions from the U.S. public sector, including the White House. To address the vulnerabilities of LLMs to prompt injections and similar attacks, OpenAI proposes an instruction hierarchy that prioritizes trusted instructions, enhancing model security against both known and new attack types while maintaining general performance with minimal impact \citep{wallace2024instruction}.

\paragraph{Downstream Applications}

Red teaming plays a crucial role in the adversarial training of AI systems by providing adversarial input \citep{yoo2021towards,tao2021recent,ziegler2022adversarial}. In addition, adversarial examples produced from red teaming can also be used to interpret models \citep{casper2022robust}. 
% Similar ideas are also present in \citep{buçinca2023aha}, where scenarios of harm from an AI system are automatically generated prior to development or deployment to help with impact assessment.

\subsubsection{Safetywashing}
The work by \cite{ren2024safetywashing} focuses on a phenomenon called \textit{Safetywashing}. \textit{Safetywashing} refers to safety labs or companies claiming to improve the safety of their models by reporting progress on benchmarks that measure model safety, which are strongly correlated with model performance. As a result, along with the increase in the model’s capabilities, the model's safety according to benchmarks prone to \textit{Safetywashing} also appears to increase. \textit{Safetywashing}, and particularly such benchmarks, create a false perception of safety progress in newer models and fail to address the actual safety issues, which should be orthogonal to improvements in the model's performance.
To determine whether a benchmark correlates with a model's capabilities and can therefore be used for \textit{Safetywashing}, the authors first propose calculating capability scores. These are calculated by constructing a matrix $B \in \mathbb{R}^{m \times b}$ for $m$ models and their respective performance on $b$ benchmarks, where columns are normalized to have zero mean and a variance of 1. Then, the first principal component $PC_1$ is extracted, and all performances are projected onto $PC_1$. This gives a general measure of a model's capabilities:
\begin{equation*}
    \text{Capabilities Score}_i = (B \cdot \text{PC}_1)_i \quad \text{for} \, i = 1, \ldots, m.
\end{equation*}
To measure the correlation of a safety benchmark with model capabilities, a set of $m$ models is evaluated on safety benchmarks (adjusted such that higher scores indicate higher safety, with variance 1 and mean 0). The capabilities correlation is then the Spearman correlation across models between the capability scores and the safety benchmark scores:
\begin{equation*}
    \text{Capabilities Correlation} = \text{corr}_{\text{models}} (\text{Capabilities Score}, \text{Safety Benchmark}).
\end{equation*}
Following this methodology, the authors evaluate the capabilities correlation of current safety benchmarks across different AI safety domains. Their results show that, with the exception of weaponization capabilities and measurements of sycophancy, safety benchmarks tend to have positive capabilities correlation. Machine ethics, dynamic adversarial robustness, and calibration benchmarks exhibit low correlation with model capabilities, while current alignment, truthfulness, static adversarial robustness, and scalable oversight benchmarks primarily reflect model capabilities and can thus be used for \textit{Safetywashing}.
The authors propose that future safety evaluations should report capabilities correlation, and the AI safety research community should aim to design benchmarks that are decorrelated from model capabilities. Furthermore, research labs and companies should not claim improved safety by improving performance on benchmarks with high capabilities correlation.

\subsection{Interpretability}
\label{sec:interpretability}

Interpretability is a research field that makes machine learning systems and their decision-making process understandable to human beings \citep{doshi2017towards, zhang2018visual, miller2019explanation}. Interpretability research builds a toolbox with which something novel about the models can be better described or predicted. In this paper, we focus on research that is most relevant to alignment and safety,\footnote{For a more comprehensive review of interpretability and its methods, we recommend \citep{rauker2023toward}.} and empirically, those techniques make neural networks safer by studying the internal structures and representations of the neural networks \citep{rauker2023toward}. Interpretability is an important research direction because in principle gaining safety guarantees about white-box systems is easier than black-box ones. 
The taxonomy of interpretability tools varies according to sub-fields and purposes \citep{doshi2017towards,rudin2019stop}. There are several ways to break down interpretability research: % explainability and transparency; interpretability of weights, neurons, sub-networks, or representations; for safety or the science of deep learning; intrinsic and post hoc interpretability; mechanistic interpretability and concept-based interpretability. 
\begin{itemize}[left=0.3cm]
    \item \textit{Explainability and Transparency}. Explainability research aims to understand why models generate specific output, whereas transparency aims to understand model internals \citep{critch2020ai}.
    \item \textit{Safety or the Science of Deep Learning}. Researchers also conduct interpretability research with different purposes: some do it to safely deploy AI systems, while others aim for a complete science of neural network. But the line gets blurred as mechanistic interpretability research aims for both \citep{olah2020zoom, olah_dreams_2023}.
    \item \textit{Intrinsic and Post Hoc Interpretability}. By the stage of intervention, interpretability research is divided into intrinsic interpretability and post hoc interpretability \citep{carvalho2019machine}: the former focuses on making intrinsically interpretable models, while the latter designs post hoc interpretability methods that offer explanations to model behaviors.
    % \item \textit{Mechanistic Interpretability, Representation Engineering, and Concept-based Interpretability}. Three research agendas have gained traction in the AI safety and alignment community \citep{alignment_course}: \textbf{Mechanistic Interpretability}, which, taking a bottom-up approach, aims to gain an understanding of low-level mechanics for algorithms implemented by neural networks \citep{olah2020zoom}, \textbf{Representation Engineering}, which, in contrast, taking a top-down approach, monitors (and manipulates) high-level cognitive phenomenon in neural networks \citep{zou2023representation}, and \textbf{Concept-based Interpretability} that locates learned knowledge representations in the neural networks, in contrast to what the models output \citep{meng2022locating, meng2022mass}. The commonality among all three is linking a feature with a set of neurons simultaneously.
    \item \textit{Top-down Approach and Bottom-up Approach} Interpretability research can also be categorized based on the direction of analysis: top-down or bottom-up approaches. The \textit{top-down approach} starts with high-level behaviors or concepts and investigates how these are represented within the neural network’s architecture. This method involves observing and manipulating high-level, macro neural representations as cognitive phenomena in models \citep{zou2023representation}. In contrast, the \textit{bottom-up approach} begins with low-level components such as neurons, weights, or circuits, aiming to build an understanding of the network’s function by dissecting these basic elements and their interactions. For example, \textit{Mechanistic Interpretability} exemplifies the bottom-up approach by seeking to reverse-engineer the computations of neural networks to gain a detailed understanding of their internal mechanisms \citep{olah2020zoom,olsson2022context,rauker2023toward}, whereas \textit{Concept-based Interpretability} locates learned knowledge in the neural networks \citep{meng2022locating, meng2022mass}.
\end{itemize}

In this section, we adopt the \textit{Intrinsic and Post Hoc Interpretability} classification method, for it offers a more generic framework suitable for various AI systems beyond neural network, and it divides the interpretability analysis both during the system designing and after the system has been deployed \citep{rauker2023toward}, compared to other classification methods. Specifically, we discussed mechanistic interpretability techniques that take place in model designing and post hoc stages separately in post hoc and intrinsic interpretability subsections.

\begin{figure}[t]
\centering
\footnotesize
        \begin{forest}
            for tree={
                forked edges,
                grow'=0,
                draw,
                rounded corners,
                node options={align=center,},
                text width=2.9cm,
                s sep=1pt,
                calign=child edge, calign child=(n_children()+1)/2,
                % l sep = 10 pt,
                % l=40pt
            },
            [Interpretability, fill=gray!45, parent
            % Pre-trained model
                % Template
                [Intrinsic Interpretability, for tree={tuning,calign=center,}
                    [Modifying Model Components, for tree={
                    tuning,
                    calign=child edge, calign child=(n_children()+1)/2,
                    yshift=0.17cm,
                    }
                        [\citenumber{wang2015falling, caruana2015intelligible, hendrycks2016gaussian, klambauer2017self, anthropic_softmax_2022}, tuning_work]
                    ]
                    [Mechanistic Investigation for Intrinsic Interpretability, for tree={
                    tuning,
                    calign=child edge, calign child=(n_children()+1)/2,
                    }
                        [\citenumber{quiroga2005invariant, openai_curve, openai_weight, olah2020zoom, goh2021multimodal, olah_dreams_2023}, tuning_work]
                    ]
                ]
                [Post Hoc Interpretability, for tree={tuning}
                    [Circuit Analysis
                        [\citenumber{openai_curve, neuralcircuits, olah2020zoom, olsson2022context, wang2022interpretability, hanna2024does, heimersheim_circuit_2023, nanda2022progress, rauker2023toward}, tuning_work]
                    ]
                    [Attribution
                        [\citenumber{ancona2018towards, durrani2020analyzing, lundstrom2022rigorous, meng2022locating, dar2023analyzing, dai2022knowledge, mcgrath2023hydra, lieberum2023does, rauker2023toward, attribution2023nanda, belrose2023eliciting}, tuning_work]
                    ]
                    [Visualization
                        [\citenumber{erhan2009visualizing, zeiler2014visualizing, simonyan2013deep, nguyen2015deep, karpathy2015visualizing, mordvintsev2015inceptionism, nguyen2016multifaceted, kindermans2018learning, olah2017feature, carter2019activation, reif2019visualizing, olah2020zoom}, tuning_work]
                    ]
                    [Perturbation and Ablation
                        [\citenumber{morcos2018importance, zhou2018revisiting, ravfogel2022linear, rauker2023toward}, tuning_work]
                    ]       
                    [Mapping and Editing Learned Representations
                        [\citenumber{bahdanau2014neural, lee2017interactive, fong2018net2vec, strobelt2018s, liu2018visual, kim2018interpretability, clark2019does, vig2019multiscale, vashishth2019attention, hao2021self, chefer2021transformer, schneider2021explaining, bansal2021revisiting, belinkov2022probing, geva2021transformer, li2021implicit, power2022grokking, meng2022locating, rigotti2022attentionbased, geva2021transformer, burns2022discovering, geva2022lm, olsson2022context}, tuning_work]
                    ]
                    [Dictionary Learning
                        [\citenumber{}]
                    ]
                ]
                [Challenges and Criticisms, for tree={tuning}
                    [Superposition
                        [\citenumber{arora2018linear, elhage2022toy, olah2020zoom, bricken2023towards, olah_dreams_2023, nanda2023othello}, tuning_work]
                    ]
                    [Scalability
                        [\citenumber{kaplan2020scaling, olsson2022context, chughtai2023toy, olah_dreams_2023, conmy2023towards, zou2023representation, circuitjuly2023anthropic}, tuning_work]
                    ]
                    [Focusing on Toy Model Experiments
                        [\citenumber{openai_curve, against_segerie, olsson2022context, wang2022interpretability, rauker2023toward}, tuning_work]
                    ]
                    [Lack of Benchmarking
                        [\citenumber{lipton2018mythos, rauker2023toward, doshi2017towards, miller2019explanation, krishnan2020against}, tuning_work]
                    ]
                ]
            ]
        \end{forest}
        \caption{A Tree diagram summarizing the key techniques concepts, challenges, and literature related to Interpretability.}
\end{figure}
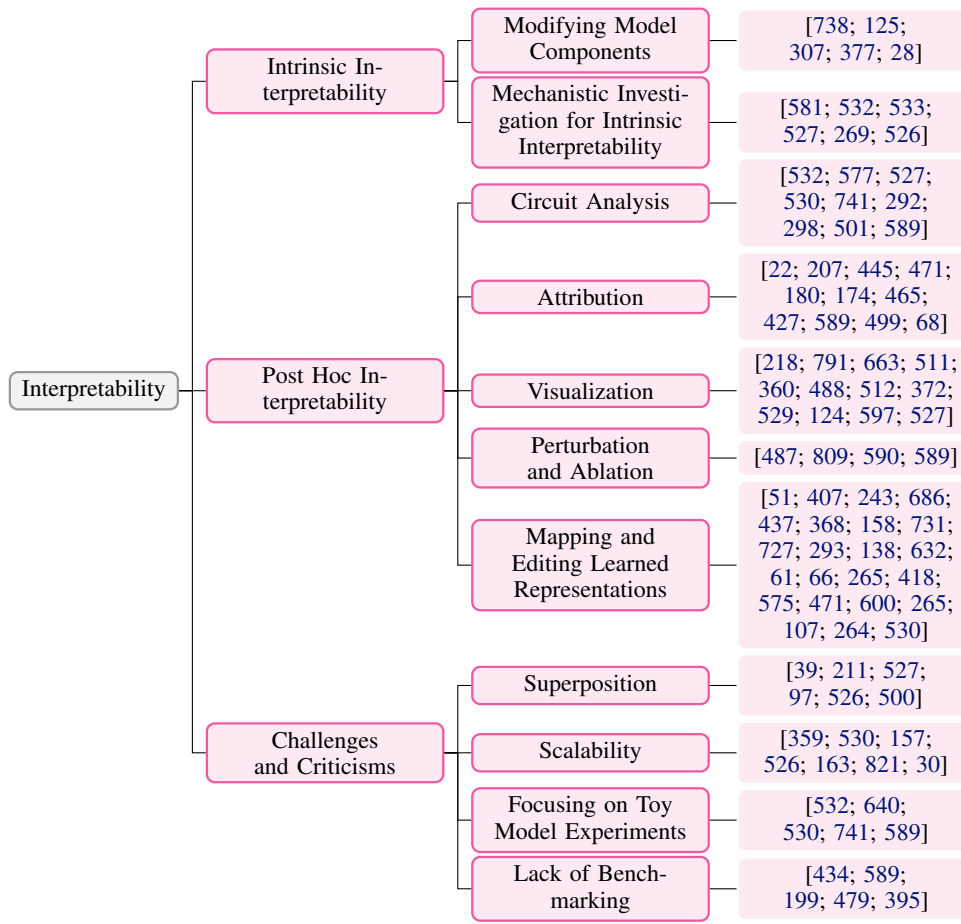

\subsubsection{Intrinsic Interpretability}
Researchers make deep learning models intrinsically more understandable, which is usually called \textit{intrinsic interpretability} \citep{carvalho2019machine}. In contrast to the symbolic approach, which emphasizes the creation of interpretable models, the modern deep learning approach tends to yield models with enhanced capabilities but potentially reduced interpretability. Compared to interpreting the black-box models, designing models that are intrinsically interpretable is safer and more efficient \citep{rudin2019stop}. To make intrinsically interpretable models, the research community designs modular architecture, which is robust to adversarial attacks and free of superposition \citep{anthropic_softmax_2022,rauker2023toward}. Notably, mechanistic interpretability, often regarded as a set of \textit{post hoc interpretability techniques}, arguably facilitates the process of making more interpretable models. 

% Adversarial training Adversarial training techniques train models with adversarial examples that are differentiable by human beings but might confuse models. Engstrom et al demonstrate that adversarial training makes models learn representations that are more human-understandable (engstrom2019adversarial). Further studies from Casper also show that (feature-level) adversaries can serve as interpretability tools because if models can differentiate human-describe adversaries it means it learns human-understandable representations. https://arxiv.org/pdf/2110.03605.pdf   https://arxiv.org/abs/2302.10894

% Modularity (and continual learning possibly)
% https://arxiv.org/pdf/2207.13243.pdf
% https://arxiv.org/abs/2305.08746
% https://arxiv.org/abs/1904.12770

% Disentanglement: train models that we can better extract features from 

% Provably safe AI   https://arxiv.org/abs/2309.01933

% Self-explaining AI https://arxiv.org/abs/1806.07538 Possibly close the gap between performance and interoperability 

% (Possibly) composibility:
% https://arxiv.org/abs/1803.03067

\paragraph{Modifying Model Components} Model components, such as feedforward layers, are hard to interpret (\textit{i.e.}, it's hard to articulate their behavior in human-understandable terms) because those layers have many polysemantic neurons that respond to unrelated inputs \citep{du2019techniques}. Thus, there are certain modifications applied to these back-box components and their related structures to make reverse engineering easier, and thus improve their interpretability \citep{carvalho2019machine}. There are a number of existing works to encourage interpretable results by modifying loss functions \citep{ross2017neural}, adding a special interpretable filter or embedding space \citep{zhang2018interpretable, wang2021interpretable}, using dynamic weight depending on the input \citep{foerster2017input}, and modifying intermediate layers \citep{li2022interpretable}. Specifically, \citet{lage2018human} proposed a human-in-the-loop algorithm that directly utilizes human feedback to quantify the subjective concept, thus achieving more reliable results. In transformer models, Anthropic proposes SoLU to replace the activation function, increasing the number of interpretable neurons and making reverse engineering easier while preserving performance \citep{anthropic_softmax_2022}. This is still an early exploration as a potentially important line of work, and challenges remain, such as the scalability of this method \citep{anthropic_softmax_2022}. 

\paragraph{Reengineering Model Architecture} Modifying existing model components is beneficial to reverse engineering \citep{carvalho2019machine, foerster2017input}, but they cannot make models \textit{fully understandable}, so some researchers started to reengineer model architecture to build theoretically interpretable models \citep{carvalho2019machine, mascharka2018transparency}. Notably, it is generally believed that there exists a trade-off between model interpretability and its performance in the same model complexity \citep{alvarez2018towards}, so it becomes crucial to design models that reach a balance between these two elements, or moreover, close the gap between interpretable models and state-of-the-art models \citep{alvarez2018towards, carvalho2019machine, fan2021interpretability, espinosa2022concept}. We will discuss the detailed research efforts below: 

\begin{itemize}
    \item \textit{Creating Transparent Reasoning Steps} In reasoning models, creating transparent minor steps is crucial to make the model interpretable \citep{arad2018compositional}, and a number of papers accomplished it by introducing the MAC (Memory, Attention, and Composition) cell to separate memory and control \citep{arad2018compositional}, by utilizing other attention-based methods \citep{lin2019kagnet, arik2021tabnet}, and by decomposing the complex reasoning process \citep{mascharka2018transparency}. These methods significantly improved the interpretability of the reasoning process but at the cost of model complexity and performance, though they close the gap of performance between interpretable and state-of-the-art models \citep{mascharka2018transparency}.
    \item \textit{Distilling Complex Knowledge} Complex models, such as deep neural networks, often have high performance but lack transparency in their decision-making processes, making them difficult to interpret \citep{li2020tnt}. Knowledge distillation addresses this challenge by transferring knowledge from these complex, 'black-box' models (teachers) to simpler, more interpretable models (students). By introducing this structure into model design, student models can approximate the performance of the teachers while offering greater transparency, thus enhancing interpretability without sacrificing the capabilities of advanced machine learning models \citep{zhang2020distilling, li2020tnt}. However, this interpretability is partial, especially in intricate missions, where the distilled knowledge may still be hard to interpret \citep{sachdeva2023data}.
\end{itemize}

Moreover, the pronoun \textit{Self-Explaining Models}, which can provide both prediction and explanation \citep{elton2020self}, was suggested by a number of papers as a better substitution to \textit{Interpretable Models}, with many papers working on it \citep{alvarez2018towards, rajagopal2021selfexplain}. For language models, the chain-of-thought (CoT) generation \citep{wei2022chain} may be recognized as a kind of self-explanation method. 

\subsubsection{Post Hoc Interpretability}
\label{sec: Techniques_and_Methods}
This section explores techniques and methods applied to understand model internals after the models are trained and deployed, thus these techniques are often referred to as \emph{post hoc interpretability} \citep{rauker2023toward}. The goal is to understand the low-level structure and units of black-box neural networks and their causal effect on macroscopic behaviors and outputs.

\paragraph{Dictionary Learning} A key challenge of post hoc interpretability is \emph{superposition}, \emph{i.e.}, the tendency of neurons to encode more than one human-interpretable feature simultaneously, which makes it very difficult to identify the individual features \citep{elhage2022toy}. To address this challenge, \emph{dictionary learning} methods have gained attention as they aim to extract sparse, interpretable features from superposed activations \citep{lin2019sparse}. Early exploration in this area assumed a linear superposition of learned factors in the textual embedding of transformers \citep{yun2021transformer}. Recently, \emph{sparse autoencoders} (SAEs) have received significant research attention \citep{bricken2023towards,cunningham2023sparse}. This method trains autoencoders in an unsupervised manner to extract features that can best replace the original activations while maintaining sparsity, thus performing a form of dictionary learning. SAEs have displayed strong scalability to both base model size and autoencoder size. This enables them to be widely applied in some of the largest frontier LLMs \citep{bricken2023towards, templeton2024scaling,gao2024scaling,deepmind2024gemma}.

% To perform the inverse process of superposition --- extracting \emph{monosemantic} features from the superposed model activations --- the method based on \emph{sparse autoencoders} (SAEs) has received significant research attention \citep{bricken2023towards,cunningham2023sparse}. Such a method trains autoencoders to extract (in an unsupervised manner) sparse features that best explain the activations, thus performing a form of dictionary learning. SAEs have displayed strong scalability with respect to both base model size and autoencoder size and have seen application in some of the largest frontier LLMs \citep{templeton2024scaling,gao2024scaling,deepmind2024gemma}.

\paragraph{Circuit Analysis} Circuits refer to the sub-networks within neural networks that can be assigned particular functionalities. As their counterparts in neuroscience, the neural circuits which are both anatomical and functional entities \citep{neuralcircuits}, circuits are also both physical and functional \citep{olah2020zoom}. Mechanistic interpretability researchers locate circuits in neural networks (microscopic) to understand model behaviors (macroscopic). Multiple circuits have been reported: curve circuits for curve detectors \citep{openai_curve}, induction circuits for in-context learning \citep{olsson2022context}, indirect object identification circuits for identifying objects in sentences \citep{wang2022interpretability}, Python docstrings for predicting repeated argument names in docstrings of Python functions \citep{heimersheim_circuit_2023}, grokking \citep{nanda2022progress}, multi-digit addition \citep{nanda2022progress}, and mathematical ability such as \textit{greater than} \citep{hanna2024does}. Notably, many circuit analysis conducted to date has been focused on toy models and toy tasks \citep{rauker2023toward}. The largest attempt to reverse engineer the natural behaviors of language models is finding the indirect object identification circuit, which is located in GPT-2 Small and has 28 heads \citep{wang2022interpretability}.

\paragraph{Probing} Probing is a collection of techniques that train independent classifiers on the interested internal learned representations to extract concepts/features. One example is Gurnee et al used probing to study the linear representations of space and time in hidden layers. \citep{gurnee2023language} Although probing has been favored by researchers to understand hidden layers \citep{alain2017understanding}, it has limitations. For one, probing does help to understand learned representations in hidden layers, but it does not tell whether learned representations are used by models to produce predictions  \citep{ravichander2021probing, belinkov2022probing}; for another, the issues of datasets may confound the issues with the model \citep{belinkov2022probing}. In the context of safety and alignment, training probe requires the dataset to contain concepts/features of interest, which means probing can not be used to detect out-of-distribution features (i.e. features you suspect learned by the models but you don't have a dataset for them).  Notably, representation engineering, built upon probing literature, is introduced to detect high-level cognitive phenomena and dangerous capabilities, including morality, emotion, lying, and power-seeking behaviors. \citep{zou2023representation}.

% \paragraph {Dictionary Learning} builds a sparse autoencoder to reconstruct learned features based on atoms (also called basis vectors) in the dictionary. A dictionary is a collection of atoms that provide one way to represent data, and each element captures one specific aspect of data. Dictionary learning also works on hidden layers and it requires training on the dataset that contains features of interest. What makes dictionary learning stand out, is it does extract features significantly more monosemantic than features extracted from neurons. While Anthropic thinks dictionary learning proposes a plausible solution to superposition, it remains unclear whether the research community has objective criteria based on which they could say features from sparse autoencoder are indeed more interpretable than features in neurons. 
        
\paragraph{Model Attribution} Attribution is a series of techniques that look at the contribution of some components (including head, neuron, layers, and inputs) for neuron responses and model outputs \citep{rauker2023toward}. Gradient-based attribution is introduced to evaluate the quality of interpretation and guide the search for facts learned by the models \citep{ancona2018towards, durrani2020analyzing, lundstrom2022rigorous, dai2022knowledge}. However, those methods are limited because they can not provide causal explanations \citep{rauker2023toward}. Direct Logit Attribution is to identify the direct contribution of individual neurons to the prediction of the next neurons \citep{lieberum2023does, mcgrath2023hydra, belrose2023eliciting, dar2023analyzing}. But attribution methods also suffer from a salient constraint: they can only help with scenarios where you have datasets for features of interest. Consequently, such attribution methods cannot help with understanding out-of-distribution (OOD) features (including some misalignment scenarios) \citep{casper2023red}.

\paragraph{Data attribution} Identifying the subset of training data that leads to a certain behavior can provide insight into both the safety of said behavior and ways to encourage or prevent that behavior. The method of \emph{influence function} \citep{koh2017understanding,grosse2023studying} have been proposed to perform such attribution by approximating the result of leave-one-out training.

\paragraph{Visualization} Techniques of visualization help to understand neural structures, including techniques that visualize datasets (notably dimensionality reduction techniques) \citep{van2008visualizing, olah2014mnist, olah2015visual}, features \citep{erhan2009visualizing, olah2017feature}, weights,  activations \citep{carter2019activation}, structure \citep{reif2019visualizing}, and the whole neural networks \citep{simonyan2013deep, zeiler2014visualizing,nguyen2015deep,  karpathy2015visualizing, mordvintsev2015inceptionism,  nguyen2016multifaceted,kindermans2018learning}. The purpose of visualization is to see neural networks with a new level of detail \citep{olah2020zoom}. 

\paragraph{Perturbation and Ablation} These techniques are designed to test the counterfactual rather than the correlation \citep{rauker2023toward}. Perturbation is a technique that modifies the input of models and observes changes in their outputs, and the ablation techniques knock out parts of neural networks\footnote{Neurons \citep{zhou2018revisiting} and Subspace \citep{morcos2018importance, ravfogel2022linear}}, helping to establish a causal relationship between neural activation and the behavior of the whole network \citep{rauker2023toward}. 

% Potentially add more MI literature. 

\paragraph{Patching} Patching refers to the collection of methods \textit{replacing} key components (paths and activations) and understanding counterfactual effects on model outputs. Among them, activations patching is a popular method among the safety community. Through applying activation patching and conducting both correct run and corrupted runs on the same neural network, researchers aim to locate key activations that matter more to the model output \citep{attribution2023nanda}. In reality, patching is used to map and edit learning representations/concepts. Specific patching techniques include interpreting token representations in transformers \citep{li2021implicit, bansal2021revisiting,geva2021transformer,  geva2022lm, power2022grokking, olsson2022context} and how do fully-connected layers learn these representations \citep{geva2021transformer, olsson2022context}, studying the key-query products to understand how do tokens attend to each other \citep{bahdanau2014neural, lee2017interactive, liu2018visual, strobelt2018s, clark2019does, vashishth2019attention, vig2019multiscale, hao2021self, chefer2021transformer, rigotti2022attentionbased}, identifying meaningful learned concepts from directions in latent space (from concepts to directions \citep{fong2018net2vec, kim2018interpretability}, and from directions to post hoc explanations \citep{schneider2021explaining}). For the purposes of safety and alignment, these techniques notably help to detect deception \citep{burns2022discovering}.

% Compared to the symbolic approach that designs interpretable models that don't work, the modern deep learning approach produces ever-capable but arguably increasingly less interpretable models. 

\subsubsection{Outlook}

% it's more clear that you should at least have two directions: making more understandable models, or understanding models (and their training process). 

\paragraph{Superposition makes the analysis at neuron level implausible} Superposition refers to the phenomenon that models represent more features than they have dimensions, so features would not correspond to neurons \citep{arora2018linear, olah2020zoom, elhage2022toy}. Superposition makes it hard to ensure AI safety by enumerating all features in a model \citep{elhage2022toy, nanda2023othello}. \citet{elhage2022toy} proposes three methods to solve superposition: creating models with no superposition (addressing it at training time), finding an overcomplete basis describing how features are stored in the neural nets (addressing it after the fact), or a mixture of both approaches. Notably, \citet{bricken2023towards} builds a sparse auto-encoder to interpret group neurons, rather than individual neurons to extract features, which points out a promising direction to solve superposition: to move past it.\footnote{see \citet{elhage2022toy} for details on conceptual and empirical research questions about superposition} 

\paragraph{Scalability} As is mentioned in the previous sections, there exists a trade-off between model interpretability and its capability \citep{alvarez2018towards}, so interpreting real models while maintaining their performance will be harder than applying those techniques to toy models. Thus, scalability becomes a concern when interpretability researchers take a bottom-up approach to interpretability (mechanistic interpretability), as top-down methods such as attention mechanism \citep{arad2018compositional} would not face such a bottleneck. For mechanistic interpretability research, we either want to scale up techniques (\textit{e.g.}, applying circuit analysis on real model \citep{wang2022interpretability}), or we want to scale up analysis (\textit{e.g.}, finding larger structure in neural networks \citep{olah_dreams_2023}). In the end, we want the microscopic analysis to answer the macroscopic model behavioral questions we care about (\textit{e.g.}, in-context learning capability \citep{olsson2022context} and more speculation about high-level cognitive capabilities such as planning and dangerous capability such as deception \citep{circuitjuly2023anthropic}).

\paragraph{Evaluation and Benchmarking} Benchmarking offers insights about what methods work and quantifies their efficiency, and it will also drive community efforts in clear and meaningful directions \citep{lipton2018mythos, casper_moving, krishnan2020against, mohseni2021multidisciplinary, madsen2022post}. Interpretability benchmarks and metrics were made to evaluate interpretability tools (by evaluating their effectiveness in detecting trojans) \citep{casper2023red}, circuits (by testing whether specific subgraphs are counted as circuits) \citep{lawrencec_causal} and explanations (by examining the faithfulness, comprehensiveness, and sufficiency of an explanation) \citep{lage2019evaluation, deyoung2020eraser, krishna2022disagreement}. However, as the inner logic of a certain AI system is unknown before the interpretability tools are applied \citep{samek2019explainable} and different explanations may even contradict each other \citep{neely2021order, krishna2022disagreement}, building a reliable evaluation benchmark or metric is rather difficult \citep{krishna2022disagreement}.
% https://www.mdpi.com/2079-9292/8/8/832

\subsection{Human Values Verification}
\label{sec:human-values-alignment}

\begin{figure}[t]
\centering
\footnotesize
        \begin{forest}
            for tree={
                forked edges,
                grow'=0,
                draw,
                rounded corners,
                node options={align=center,},
                text width=2.7cm,
                s sep=6pt,
                calign=center,
            },
            [Human Values Verification, fill=gray!45, parent
            % Pre-trained model
                % Template
                    [Formal Frameworks for Ethics and Cooperation in AI, for tree={
                    tuning,
                    calign=center,
                    }
                        [Formal Machine Ethics, for tree={
                    tuning,
                    calign=child edge, calign child=(n_children()+1)/2,
                    yshift=-0.4cm,
                    }
                            [Logic-based methods
                                [\citenumber{von1951deontic, arkoudas2005toward, mermet2016formal, pereira2016programming, dennis2016formal, berreby2017declarative, dubljevic2020toward}, tuning_work]
                            ]
                            [RL and MDP-like settings
                                [\citenumber{abel2016reinforcement, berreby2017declarative, wu2018low, svegliato2021ethically, murtarelli2021conversation}, tuning_work]
                            ]
                            [Game theory-based methods
                                [\citenumber{rossi2011short, pereira2016bridging, pereira2016programming, conitzer2017moral, noothigattu2018voting}, tuning_work]
                            ]
                        ]
                        [Game Theory for Cooperative AI, for tree={
                            tuning,
                            calign=center
                    }
                            [Classical Game Theory for Cooperative AI, for tree={
                            yshift=0.35cm,
                            calign=child edge, calign child=(n_children()+1)/2,
                            }
                                [\citenumber{pita2010robust, li2017review, fiez2020implicit, dafoe2020open, mckee2020social, oesterheld2022safe}, tuning_work]
                            ]
                            [Evolutionary Game Theory for Cooperative AI, for tree={
                            calign=child edge, calign child=(n_children()+1)/2,
                            }
                                [\citenumber{sachs2004evolution, schuster1983replicator, weibull1997evolutionary}, tuning_work]
                            ]
                        ]
                    ]
                    [Techniques, for tree={tuning,
                    calign=child edge, calign child=(n_children()+1)/2,
                    }
                        [Building Moral Dataset
                            [\citenumber{min2023recent, awad2018moral, hagendorff2022virtue, hendrycks2020aligning, jin2022make,abdulhai2022moral,pan2023rewards,scherrer2024evaluating}, tuning_work]
                        ]
                        [Scenario Simulation
                            [\citenumber{pan2023machiavelli, yuan2022situ, liu2024training}, tuning_work]
                        ]
                        % [Value Evaluation Methods
                        %     [\citenumber{durmus2023towards, zhang2023heterogeneous,
                        %     saunders2022self,
                        %     zhang2023measuring}, tuning_work]
                        % ]
                    ]
                ]
        \end{forest}
            \caption{A Tree diagram summarizing the key concepts, logic, and literature related to Human Value Verification. The root of the tree represents Human Value Verification, which aims to \textit{verify whether AI systems can adhere to the social norms and moral values}. The main branches represent the main structure of human value verification, including Formal Frameworks for Ethics and Cooperation in AI and specific Techniques of value verification. Further sub-branches list key works exploring each of these branches. This diagram provides an overview of research directions and specific techniques for making AI systems align with human values and social norms.}
\end{figure}
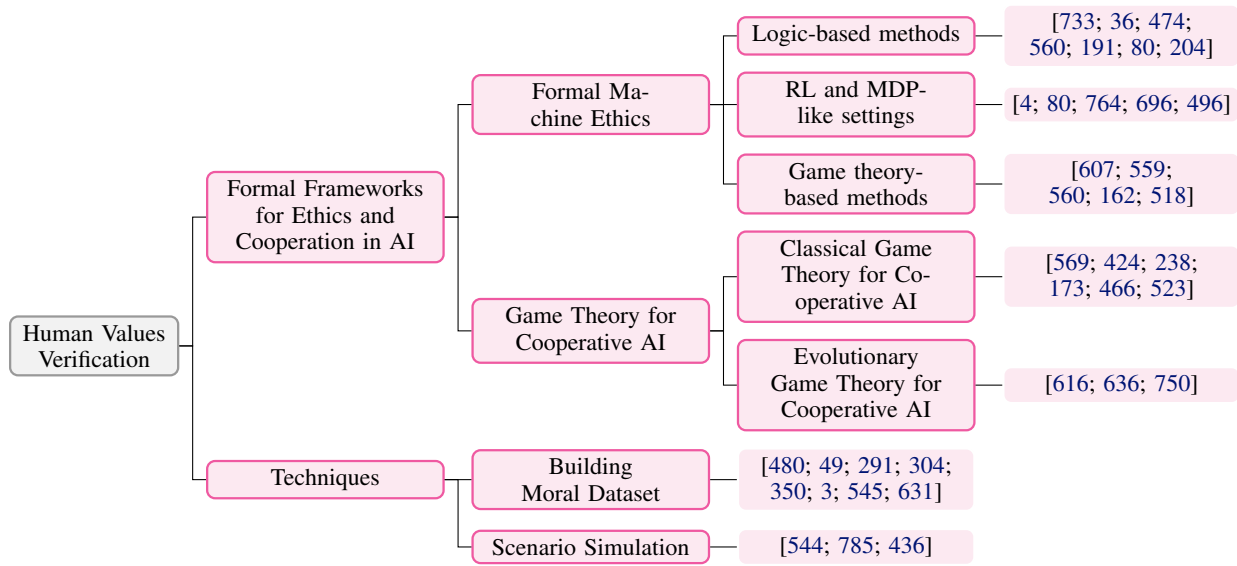

\textit{Human Values Alignment} refers to the expectation that AI systems should adhere to the community's social and moral norms ~\citep{chatila2019ieee}. As the capabilities of AI systems advance, some have begun to exhibit abilities approaching AGI ~\citep{openai2023gpt4}. In the future, we can expect autonomous agents governed by these AI systems to become an integral part of our daily lives ~\citep{lee2023if}. However, if these systems fail to grasp the inherent complexity and adaptability of human values, their decisions could result in negative social outcomes. In this context, simply aligning with human intent may not be sufficient. Thus evaluating the alignment of human morality and values between AI systems and human beings becomes crucial ~\citep{weidinger2023using}. This underscores the importance of designing AI entities that are more socially oriented, reliable, and trustworthy. Following the logic of theoretical research and practical techniques, we divide our discussion of human value alignment into these two aspects: \emph{Formulations} \S\ref{subsec:formal-ethics-coop} and \emph{Evaluation Methods} \S\ref{subsec:value-learning-techniques} of human value alignment.

\subsubsection{Formulations}
\label{subsec:formal-ethics-coop}

As the formulation of value is complicated, we introduce frameworks that formally characterize aspects of human values that are relevant to alignment. Specifically, we focus on two topics: \emph{formal machine ethics} and \emph{game theory for cooperative AI}. The former focuses on building a formal framework of machine ethics, while the latter discusses the value of multiagent systems, which share a similar origin of the game process.

\paragraph{Formal Machine Ethics} Machine ethics \citep{yu2018building, winfield2019machine, tolmeijer2020implementations}, first introduced in \S\ref{sec:values-in-intro}, aim to build ethically-compliant AI systems. Here, we introduce the branch of machine ethics that focuses on formal frameworks -- what we call \textit{formal machine ethics}. We explain three approaches to formal machine ethics: logic-based, RL/MDP-based, and methods based on game theory/computational social choice:
\begin{itemize}[left=0.2cm]
    \item \textbf{Logic-based methods}. One major direction within formal machine ethics focuses on logic \citep{pereira2016programming}. A number of logic-based works use or propose special-purpose logic systems tailored for machine ethics, such as the Agent-Deed-Consequence (ADC) model \citep{dubljevic2020toward}, deontic logic \citep{von1951deontic,arkoudas2005toward}, event calculus and its variants \citep{berreby2017declarative}. Other works also develop methods for the formal verification of moral properties or frameworks for AI systems that accommodate such kind of formal verification \citep{dennis2016formal, mermet2016formal}.

    \item \textbf{RL \& MDP-like settings}. Another line of work concerns statistical RL or other similar methods for planning within MDP-like environments \citep{abel2016reinforcement, svegliato2021ethically}. In particular, some works \citep{wu2018low,svegliato2021ethically} involve the utilization of the manual design of ethics-oriented reward functions, a concept denoted as \textit{ethics shaping}. Conversely, in other works \citep{berreby2017declarative,murtarelli2021conversation}, the segregation of ethical decision-making from the reward function is pursued.

    \item \textbf{Game theory-based methods}. To address multi-agent challenges, researchers have developed machine ethics methods based on game theory and computational social choice. Championed by \citet{pereira2016bridging}, methodologies of existing work can be broadly partitioned into Evolutionary Game Theory (EGT) \citep{pereira2016programming}, classical game theory \citep{conitzer2017moral}, and computational social choice \citep{rossi2011short, noothigattu2018voting}.
\end{itemize}

\paragraph{Game Theory for Cooperative AI} \emph{Cooperative AI} \citep{dafoe2020open,dafoe2021cooperative} aims to address uncooperative and collectively harmful behaviors from AI systems (see \S\ref{sec:challenges-of-alignment}). Here we introduce the branch of cooperative AI that focuses on game theory to complement the introduction to MARL-based cooperative training in \S\ref{sec:cooperative-ai-training}. This branch tends to study the \emph{incentives} of cooperation and try to enhance them, in contrast to the MARL's tendency to emphasize the \emph{capabilities} of coordination. Examples of incentive failures include game theory dilemmas like the prisoner's dilemma \citep{phelps2023investigating} and tragedy of the commons \citep{perolat2017multiagent}, while examples of coordination capability failures include bad coordination of a robot football team \citep{ma2022elign}. 

\begin{itemize}[left=0.2cm]
    \item \textbf{Classical Game Theory for Cooperative AI.} Recent work on Stackelberg games --- a theoretical model for \emph{commitment} in games --- include the introduction of bounded rationality into the model \citep{pita2010robust}, dynamic models \citep{li2017review}, machine learning of Stackelberg equilibria \citep{fiez2020implicit}, and more. Apart from Stackelberg games, \emph{mixed-motive games} have also received extensive study \citep{dafoe2020open,mckee2020social,oesterheld2022safe}.
    \item \textbf{Evolutionary Game Theory for Cooperative AI}. Another avenue of research, initiated by \citet{sachs2004evolution}, aims to understand how cooperation emerges from evolution -- this includes human cooperation, which arose from Darwinian evolution, as well as the cooperation tendencies in AI systems that could emerge within other evolutionary settings such as the replicator dynamics \citep{schuster1983replicator,weibull1997evolutionary}.
\end{itemize}

\subsubsection{Evaluation Methods}
\label{subsec:value-learning-techniques}

In this section, we assume that we have already obtained the appropriate value that should be aligned. However, even so, under the guidance of Goodhart's Law ~\citep{goodhart1984problems}, we cannot simply define complex human values as reward functions, which also brings greater challenges to value alignment. We introduce specific human value alignment techniques in three parts: \emph{Building Moral Dataset}, \emph{Scenario Simulation}.

\paragraph{Building Moral Dataset} \textit{Moral Alignment} refers to the adherence of AI systems to human-compatible moral standards and ethical guidelines while executing tasks or assisting in human decision-making ~\citep{min2023recent}. Early attempts at moral value alignment, initiated in 2018 \citep{awad2018moral}, have confirmed that the definition and evaluation of moral values themselves is a challenging issue. This has led to the emergence of abstract moral standards \citep{hagendorff2022virtue} and various different standards driven by the average values of diverse community groups \citep{awad2018moral}, fueling further in-depth research into moral value assurance. 

Assurance of moral values is typically achieved by constructing corresponding datasets. The Rule-of-Thumb (RoT) serves as a gauge for determining what actions are considered acceptable in human society. Building on this concept, \citet{emelin2021moral, forbes2020social, ziems2022moral} introduced the Moral Stories, SOCIAL-CHEM-101, and Moral Integrity Corpus datasets respectively, focusing on providing human social and moral norms. \citet{hendrycks2020aligning} and \citet{jin2022make} introduced the ETHICS and MoralExceptQA datasets respectively, highlighting the inability of contemporary models to align ethically with human values. \citet{abdulhai2022moral} found that models exhibit certain morals and values more frequently than others, revealing how the moral foundations demonstrated by these models relate to human moral foundations. \citet{pan2023rewards} explored the trade-off between rewards and moral behavior, discovering a certain tension between the two.

% Other related works focus on specific values. For example, \citet{scherrer2024evaluating} focused more on ambiguous situations, assessing different models' reactions in these contexts, while \citet{roger2023measurement} studied the phenomenon of Measurement Tampering, providing corresponding evaluation methods and datasets.

\paragraph{Scenario Simulation} \textit{Scenario simulation} is a more complex form than datasets and therefore is considered by some views to be more effective in replicating real situations and harvesting better results. The form of the scenario can also vary. \citet{pan2023machiavelli} built a series of diverse, morally salient scenarios through text adventure games, evaluating complex behaviors such as deception, manipulation, and betrayal. On the other hand, some work attempts to make intelligent agents learn human values through simulating human-machine interaction. \citet{yuan2022situ} proposed a method for bidirectional value alignment between humans and machines, enabling machines to learn human preferences and implicit objectives through human feedback. \citet{liu2024training} placed AI within a simulated human society sandbox, allowing AI to learn human societal value inclinations by mimicking human-social interactions.

% \paragraph{Value Evaluation Methods} 
% The existing evaluation models show a very diverse range of methods in terms of values. \citet{durmus2023towards} gathered data on human values from five distinct cultures worldwide. To evaluate LLM's value orientations, they compared the similarity between responses produced by LLM and those obtained from these diverse human groups. The results of the study indicate that LLM still displays a noticeable degree of value bias. At the same time, \citet{zhang2023heterogeneous} examined the value rationality of LLMs across various values using the framework of social value orientation ~\citep{messick1968motivational, mcclintock1982social, liebrand1984effect, van1997development}. Their findings suggest that LLMs are more likely to opt for actions reflecting neutral values, such as \textit{prosocial}. The Discriminator-Critique Gap (DCG), originally termed the Generator-Discriminator-Critique Gaps ~\citep{saunders2022self}, is a metric designed to gauge a model's ability to produce responses, judge the quality of these responses, and offer critiques. \citet{zhang2023measuring} discovered that DCG can also determine if an LLM can autonomously identify its values and convey to humans the reasons for holding those values. Following this, they proposed VUM to quantify LLM's understanding of human values through DCG based on a dataset built with values from the Schwartz Value Survey ~\citep{schwartz1992universals, schwartz1994there}. 

\section{Governance}
\label{sec:governance}
Besides technical solutions, governance, the creation and enforcement of rules, is necessary to ensure the safe development and deployment of AI systems. In this section, we survey the literature on AI governance by exploring the role of AI governance, the functions, and relationships between stakeholders in governing AI, and several open challenges to effective AI governance.

\subsection{The Role of AI Governance}
\label{sec:the-role-of-ai-governance}
To explore the role of AI governance, we must identify the challenges that require governance. A range of social and ethical issues can and have already emerged from the adoption and integration of AI into various sectors of our society \citep{summit2023round}. For instance, AI applications can inadvertently perpetuate societal biases, resulting in racial and gender discrimination \citep{caliskan2017semantics,perez2022discovering}. Moreover, unchecked reliance on these systems can lead to repercussions such as labor displacement \citep{acemoglu2018artificial}, widening socioeconomic disparities, and the creation of monopolistic environments.

AI systems have exhibited the potential to jeopardize global security \citep{turchin2020classification}. For example, OpenAI’s system card for GPT-4 \citep{openai2023gpt4} finds that an early version of the GPT-4 model as well as a version fine-tuned for increased helpfulness and harmlessness exhibits capabilities to enable disinformation, influence operations, and engineer new biochemical substances, among other risky behavior. \citet{urbina2022dual} further demonstrated the potential of AI systems to enable the misuse of synthetic biology by inverting their drug discovery model to produce 40,000 toxic molecules. 

The horizon also holds the prospect of increasingly agentic and general-purpose AI systems that, without sufficient safeguards, could pose catastrophic or even existential risks to humanity \citep{mclean2023risks}. For example, OpenAI's \citet{weng2023agent} argued that models such as LLM could essentially act as the brain of an intelligent agent, enhanced by planning, reflection, memory, and tool use. Projects such as AutoGPT, GPT-Engineer, and BabyAGI epitomize this evolution. These systems can autonomously break down intricate tasks into subtasks and make decisions without human intervention. Microsoft research suggests that GPT-4, for instance, hints at the early inklings of AGI \citep{bubeck2023sparks}. As these systems evolve, they might lead to broad socio-economic impacts such as unemployment, and potentially equip malicious actors with tools for harmful activities. 

The major objective of AI governance is to mitigate this diverse array of risks. In pursuit of this goal, relevant actors should maintain a balanced portfolio of efforts, giving each risk category its due consideration.

\subsection{The Multi-Stakeholder Approach}\label{sec:multi-stake}

\begin{figure}[t]
    \centering
    \includegraphics[width=1.0\textwidth]{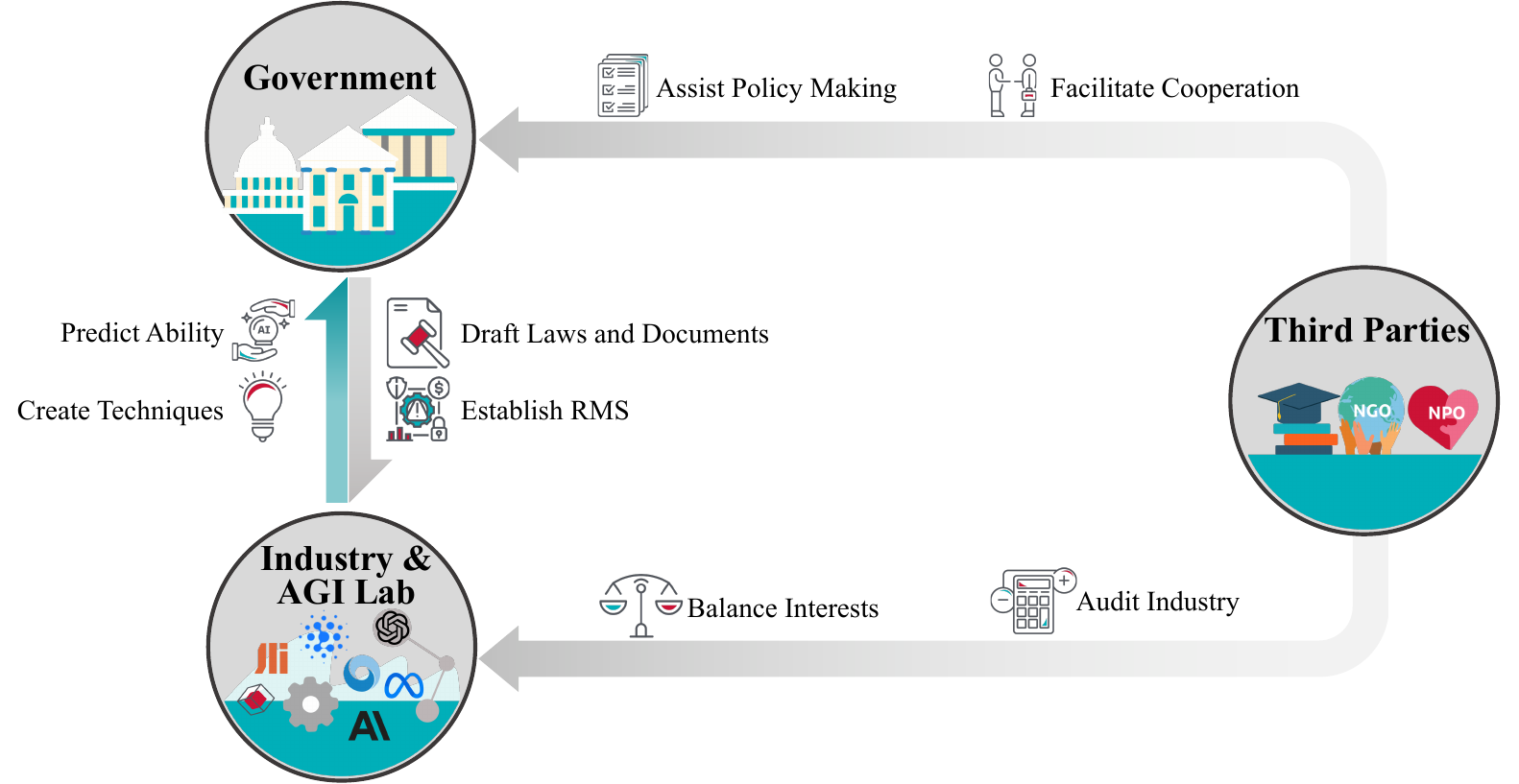}
    \caption{Our framework for analyzing AI governance at present. The proposed framework explains the nonexhaustive interrelationships and functions among three primary entities in AI governance: the government, industry and AGI labs, and third parties. The government's governance role encompasses regulating the industry and AGI labs and directing the trajectory of future AI development through policy documents. It also devises a \emph{Risk Management System} (RMS) \citep{mannes2020governance} to abate AI-related threats. Industry and AGI labs return by offering watchful predictions into AI development and innovating new technological methodologies to support regulatory measures (such as model evaluation \citep{shevlane2023model}). Third parties fulfill a dual function, offering expert advice for robust governmental policy development and fostering collaborations among governments. In the context of industry and AGI labs, these third parties assist in equilibrating corporate interests to prevent disorganized competition from information asymmetry. They also deliver auditing services to the industry and AGI labs as independent entities.}
    \label{fig:governance}
\end{figure}
We put forward a framework to analyze the functions and relationships between stakeholders in AI governance (see Figure \ref{fig:governance}). In this framework, we outline three main entities. \textbf{Government Agencies}  oversee AI policies using legislative, judicial, and enforcement powers, as well as engage in international cooperation. \textbf{Industry and AGI Labs} research and deploy AI technologies, making them subjects of the governance framework, while proposing techniques to govern themselves and affecting governance policy. \textbf{Third Parties}, including academia, Non-Governmental Organizations (NGOs), and Non-Profit Organizations (NPOs), perform not only auditing on corporate governance, AI systems, and their applications but also assist the government in policy-making.

Proposals have been made about specific principles for a multi-stakeholder AI governance landscape. Notably,  \citet{brundage2020toward} argues to implement institutions, software, and hardware to make claims about the safety of AI systems more verifiable.

\paragraph{Government}
According to \citet{anderljung2023frontier}, three building blocks for government regulation are needed: (1) standard development processes to determine appropriate requirements for cutting-edge AI developers, (2) registration and reporting requirements to offer regulators insight into the progress of advanced AI development processes, (3) mechanisms to guarantee adherence to safety standards in the development and deployment of cutting-edge AI models.

At present, an emerging collection of governmental regulations and laws is surfacing on a global scale, including the \emph{European Union's AI Act} \citep{euaiact}, and the \emph{Bipartisan Framework for U.S. AI Act} \citep{bipartisan}. Such regulations are indispensable for the safety and alignment of AI systems.

\paragraph{Industry and AGI Labs} 
Governance efforts in industry and AGI labs should emphasize comprehensive AI risk assessments throughout the lifecycle of the AI system. Based on discussions in \citet{koessler2023risk,schuett2023towards}, the full cycle of AI risk assessment can be seen as consisting of five stages. \textbf{Pre-development risk assessments}, \textbf{pre-training risk assessments}, and \textbf{pre-deployment risk assessments} all include predictions and analyses of impact and risks with a variety of tools, but with increasing amounts of detail, clarity, and sophistication \citep{koessler2023risk}. \textbf{Post-deployment monitoring} is the phase where mechanisms for monitoring are established, and all previous analyses are continually updated post-deployment \citep{koessler2023risk}. \textbf{External scrutiny} includes bug bounty programs \citep{schuett2023towards}, external red teaming and third-party model auditing \citep{schuett2023towards,anderljung2023frontier}

Taking security measures against the risks associated with AI systems seems to be widely accepted by AI companies and related practitioners. 
\citet{schuett2023towards} shows that 98\% of respondents who have been surveyed somewhat or strongly approved that AGI labs should perform pre-deployment risk assessments, hazardous capabilities evaluations, third-party model audits, safety restrictions on model usage, and red teaming to guarantee AI safety. Meanwhile, leading AI companies, including Amazon, Anthropic, Google, Inflection, Meta, Microsoft, and OpenAI, have voluntarily committed to the government to implement security measures \citep{house2023fact}.

Notably, a lot of researchers have proposed pausing the development of advanced AI systems to win more time for safety research, risk assessments, and regulatory preparations \citep{2023pauseai}. Their proposals include blanket pausing of all sufficiently advanced systems \citep{2023pauseai}, and also conditional pausing of specific classes of models in response to evaluation results on specific failure modes \citep{alaga2023coordinated}, including the currently adopted practice of \emph{responsible scaling policy} (RSP) \citep{rsp2023anthropic}.

\paragraph{Third Parties}
\citet{mokander2023auditing} presents three key functions of third-party auditing: (1) \emph{Governance audits} (of tech providers that design and disseminate LLMs) (2) \emph{Model audits} (of LLMs after pre-training but prior to their release) (3) \emph{Application audits} (of applications based on LLMs).

One prominent example of existing third-party audits is that of METR, initially a project of Alignment Research Center \citep{arcevalscollab,kinniment2023evaluating}, who collaborated with OpenAI to perform red teaming on GPT-4 \citep{openai2023gpt4} and partnered with Anthropic to perform red teaming on Claude 2 \citep{anthropiceval}. These efforts include evaluations on toxicity and bias, as well as frontier AI risks such as autonomous replication, manipulation, cybersecurity, and biological weapon risks \citep{openai2023gpt4,shevlane2023model}. 

Apart from auditing, third parties can support AI governance in other ways, such as assisting policy-making and facilitating cooperation internationally \citep{ho2023international}. For example, \citet{maas2021aligning} thinks that the government should prefer technology-neutral rules rather than technology-specific rules. \emph{AI4People’s Ethical Framework for a Good AI Society: Opportunities, Risks, Principles, and Recommendations} \citep{floridi2021ethical}, released by AI4People, was guided to the Ethics Guidelines for Trustworthy Artificial Intelligence presented in April 2019 \citep{ato2023}. The World Economic Forum (WEF) convenes government
officials, cooperations, and civil society and it has initiated a Global
AI Action Alliance in collaboration with partner organizations, with the goal of promoting international cooperation in the field of AI \citep{kerry2021strengthening}.

\subsection{Open Problems}
\label{sec:open-problems}
There are numerous open problems in the existing field of AI governance. These problems often have no clear answers, and discussion of these questions can often promote better governance. For effective AI governance, we mainly discuss international governance and open-source governance, hoping to promote the safe development of AI through our discussion.

\subsubsection{International Governance}\label{sec:intl-gov}
Amidst the swift progress and widespread implementation of AI technology universally, the need for international governance of AI is high on the agenda \citep{bletch2023summit}. Critical discussions revolve around the necessity to institute a global framework for AI governance, the means to ensure its normativity and legitimacy \citep{erman2022artificial}, among other significant concerns. These themes draw an intensifying level of detail and complexity in their consideration. Also, as stated by United Nations secretary-general António Guterres during a Security Council assembly in July, generative AI possesses vast potential for both positive and negative impacts at scale, and failing to take action to mitigate the AI risks would be a grave neglect of our duty to safeguard the well-being of current and future generations \citep{un2023remark}, international governance also has intergenerational influence. Hence, we examine the significance and viability of international AI governance from three aspects within this section: \emph{manage global catastrophic AI risks}, \emph{manage opportunities in AI}, and \emph{historical and present efforts}, with both generational and intergenerational perspectives. We aim to contribute innovative thoughts for the prospective structure of international AI governance.

\paragraph{Manage Global Catastrophic AI Risks}
The continual advancements in AI technology promise immense potential for global development and prosperity \citep{vinuesa2020role}. However, they inevitably harbor underlying risks. The unchecked competition in the market and geopolitical factors could precipitate the untimely development and deployment of advanced AI systems, resulting in negative global externalities \citep{tallberg2023global}. The amplification of existing inequalities such as racial and gender bias \citep{pri2020shea} ingrained in AI systems may result in intergenerational ethical discrimination. Since these risks are international and intergenerational, it seems that international governance interventions could alleviate these catastrophic global AI challenges. For example, a consensus amongst nations could help defuse potential AI arms races, while an industry-wide agreement could avert the hasty and irresponsible development of sophisticated AI systems, thus securing the long-term and sustainable development of AI \citep{ho2023international}.

\paragraph{Manage Opportunities in AI}
The opportunities created by AI development are not distributed equally, which may cause enduring digital inequality between regions and harm the sustainability of AI development. Geographic variances in AI progression suggest an inequitable distribution of its economic and societal benefits, potentially excluding developing nations or specific groups from these advantages \citep{ho2023international,tallberg2023global}. Moreover, the consolidation of decision-making authority within the technology sector among a limited number of individuals \citep{intergen2021ai,airegth2021sa} could cause an intergenerational impact. Such inequality in the distribution of interests can be mitigated through international governance. Effective international consensus and coordination on the allocation of AI opportunities, which is facilitated by its propagation, education, and infrastructural development \citep{committ2023ro}, could ensure a balanced distribution of benefits derived from AI and promote sustainability in its ongoing development.

\paragraph{Historical and Present Efforts}
Before the surge of AI technology, the international community had laid down frameworks in line with cooperative regulation of influential technologies and critical matters. For example, the Intergovernmental Panel on Climate Change (IPCC) convened specialists to assess climactic environmental issues, fostering scientific consensus \citep{ho2023international}. The International Civil Aviation Organization (ICAO) standardized and oversaw international regulations, simultaneously assessing the member nations' compliance with these laws \citep{ho2023international}. The International Atomic Energy Agency (IAEA) propelled the harmonious utilization of nuclear energy, with its global reach and sophisticated monitoring and evaluation mechanisms. Fast forward to the present-day scenario, wherein multiple international organizations have arrived at a consensus on AI governance. In 2019, the G20 members consolidated a ministerial declaration focusing on human-centered artificial intelligence principles \citep{G20prin2019}. Concurrently, the Organisation for Economic Cooperation and Development (OECD) set forth the \emph{OECD Principles on Artificial Intelligence} \citep{oecdpri2019}. The IEEE Standards Association launched a worldwide initiative aimed at \emph{Securing that all stakeholders involved in the design and implementation of autonomous and intelligent systems receive proper education, training, and motivation to emphasize ethical concerns, thereby advancing these technologies for the betterment of humanity.} \citep{chatila2019ieee}. In 2021, the United Nations Educational, Scientific and Cultural Organization(UNESCO) produced the first-ever global standard on AI ethics \citep{unesco2021}, which aims to lay the foundations for making AI systems work for the good of humanity and societies,  and to prevent potential harm caused by losing control over AI systems. In 2023, the AI Safety Summit was convened in London, United Kingdom. Countries held roundtable discussions on the risks and opportunities of AI and jointly issued the Bletchley Declaration \citep{bletch2023summit}. The scholarly community has also proposed prospective international governance frameworks for AI, such as the International AI Organization (IAIO) \citep{trager2023international}. We hope these precedents and research outcomes will inspire and provide the groundwork for developing a resilient and long-lasting international framework for AI governance in the future.

\subsubsection{Open-Source Governance}\label{sec:o-s-g}
The debate over the open-sourcing of contemporary AI models is contentious in AI governance, particularly as these models gain increased potency \citep{seger2023open}. The potential security hazards linked with making these models open-source continue to be the crux of debates among AI researchers and policymakers. The offence-defence balance in open-source AI governance also remains controversial. There is still debate over whether open-source models will increase model security or increase the risk of abuse. As referenced in \citet{shapiro2010paper}, the efficacy of disclosure depends on the chance of potential attackers already possessing the knowledge, coupled with the government's capacity to convert transparency into the identification and solution of emerging vulnerabilities. Some scholars have already conducted preliminary discussions on the offense-defense balance in the AI field, such as \citet{weng2023attack}'s discussion of adversarial attacks. If a suitable equilibrium between offence and defence cannot be forged for AI systems, the open-sourcing could potentially give rise to significant risks of AI system misuse. 

For precision and clarity, we adhere to the definition of open-source models stated by \citet{seger2023open}: enabling open and public access to the model's architecture and weights, allowing for modification, study, further development, and utilization by anyone. Currently, the most recognized open-source models include Llama2, Falcon, Vicuna, and others. In this section, we evaluate the security advantages and potential threats posed by open-source models, fostering a discourse on the feasibility of open-sourcing these models. Ultimately, our objective is to amalgamate insights from existing studies to put forward suggestions for future open-source methodologies that will ascertain the secure implementation of these models.

\paragraph{Arguments for Open-sourcing}
The view that supports the open-sourcing of existing models suggests that this method can mitigate the security risks inherent in these models in several ways:
\begin{itemize}[left=0.3cm]
\item \textbf{Potentially Bolster Model's Security}.  Meta's assertions in their release blog for Llama2 \citep{Llama22023} promote the belief that this enables the developer and the technical community to conduct tests on the models. As a result, this rapid identification and resolution of issues can considerably strengthen model security. In contrast, another perspective suggests that open-sourcing existing models could enhance the recognition of associated risks, thereby facilitating a greater focus on, investigation into, and mitigation of these potential hazards \citep{grover2019}.
\item \textbf{Foster the Decentralization of Power and Control}.  Open-sourcing has been widely recognized as an effective strategy in reducing the dominance of major AI laboratories across various sectors, including economic, social, and political domains \citep{seger2023open}. An example is articulated in the core reasons for Stability's open-sourcing of Stable Diffusion: They place their trust in individuals and the community, as opposed to having a centralized, unelected entity controlling AI technology \citep{stability2022}. Furthermore, certain commentators draw an analogy between open-sourcing and the Enlightenment Era, asserting that decentralized control reinforces faith in the power and good of humanity and society \citep{agedis2023}, implementing central regulations for safety purposes might amplify the power of the AI technology community instead. 
\end{itemize}

\paragraph{Arguments against Open-sourcing}
Critics of open-source models assess the potential for misuse from the following viewpoints:
\begin{itemize}[left=0.3cm]
\item \textbf{Potentially Be Fine-Tuned into Detrimental Instances}. Current research rigorously affirms that AI systems, contradictory to their initial design intent for mitigating toxicities in chemistry or biology, now hold the potential to manufacture new chemical toxins \citep{urbina2022dual} and biological weaponry \citep{sandbrink2023artificial}. The malicious fine-tuning of such models could lead to profound security risk manifestations. 
Besides, language models, once fine-tuned, could emulate skilled writers and produce convincing disinformation, which may generate considerable sociopolitical risks \citep{goldstein2023generative}.
\item \textbf{Inadvertently Encourage System Jailbreaks}. Research indicates that unfettered access to open-sourced model weights could facilitate bypassing system security measures \citep{seger2023open}. This premise was epitomized by \citet{zou2023universal}, who showcased this potentiality by developing attack suffixes using Vicuna-7B and 13B. Once implemented within readily accessible interfaces such as ChatGPT, Bard, and Claude, these provoked unwanted generations. Therefore, open-sourcing a model might unintentionally undermine the safeguarding protocols of models that are not open-sourced, consequently amplifying the likelihood of model misuse.
\end{itemize}

\paragraph{Tentative Conclusions on Open-Source Governance}
The debate on the open-sourcing of AI models remains unsettled, with a prevailing viewpoint that the disclosure of AI models does not pose significant risks at present. Our discourse not only synthesizes existing perspectives on this topic but also prepares the ground for future deliberations considering the prudence of open-sourcing more advanced AI systems.

Existing guidelines for open-sourcing advanced AI systems include measures such as evaluating risks by quantifying the potential for misuse via fine-tuning and a gradual model release \citep{solaiman2019release,seger2023open}. Meanwhile, policymakers are establishing rigorous compliance protocols for these open-source models. For example, European policymakers insist that the models should have “performance, predictability, interpretability, corrigibility, security, and cybersecurity throughout [their] lifecycle.” \citep{bal2023}.

\subsection{Rethinking AI Alignment from a Socio-technical Perspective}
In the preceding discussion, our primary focus is on AI systems as the core of AI Alignment. We examine strategies to align the system with human intentions and values throughout its lifecycle, considering both forward and backward alignment. In the future, AI will address more challenging and high-stakes decisions, \textit{e.g.}, ``How to allocate resource for fairness?'' and ``Which drugs are safe to approve?''. These decisions will require not only significant expertise for well-informed answers but also involve value judgments, leading to strong disagreements among informed individuals based on differing values. Furthermore, AI systems may transmit incorrect values, sway public opinion, facilitate cultural invasion, and exacerbate social division \citep{goldstein2023generative}. Singapore Conference on AI (SCAI) once introduced 12 questions that are meant to be a holistic formulation of the challenges that should be addressed by the global AI community to allow humanity to flourish \footnote{\url{https://www.scai.gov.sg/}}.
In the area of alignment we are more concerned about the following question: as AI systems evolve into socio-technical entities, how can alignment techniques mit igate the challenges they pose to human society? Specifically, we explore the incorporation of values into AI systems through alignment techniques and provide insights into security methods. We also aim to identify the alignment techniques needed to address the socio-technical challenges posed by future AI systems.

\subsubsection{Incorporating Values into AI Systems}

Aligning AI systems with human morals and societal values is a key objective of alignment technology. However, current technologies (\textit{e.g.}, RLHF) primarily blend preferences without distinguishing specific values, focusing solely on human preferences. Human preferences effectively address the basic alignment issue: ensuring models align with human intentions and safety, but not morals and societal values. However, minor errors in future AI systems' critical problems can lead to disagreements among people with differing viewpoints. Truly understanding human values is crucial for AI systems to generalize and adapt across various scenarios and ideologies. Incorporating values into AI systems generally involves two aspects: aligning with individual values (\S\ref{sec:human-values-alignment}), and aligning with collective values.

In this part, we mainly discuss the second topic. The main challenge of collective value alignment lies in determining which groups to include. A prevalent approach is defining universal values like fairness, justice, and altruism, exemplified by the veil of ignorance. However, this work remains theoretical; another approach avoids defining universal values, instead seeking the broadest overlap of values across cultures. \citet{bakker2022fine} initiated this approach by gathering preferences from various demographics, training a language model, and aggregating results using diverse social welfare functions. Similarly, simulated deliberative democracy has been proposed to enhance decision-making \citep{leike2022proposal}. Specifically, individuals from diverse demographics reach consensus on value-laden topics with AI assistance. This data informs new model training, enabling simulation of deliberative democracy for more apt responses to new value-laden issues.

Furthermore, instead of providing a consensus answer to all users, collective value alignment should encourage AI systems to tailor responses to specific demographic groups. In other words, what values should guide the model's responses to specific questions or in certain dialogues? Democratic Fine-Tuning \citep{mai2023dft} uses a value card and moral graph to link various values, allowing fine-tuned LLMs to reflect on their moral context before responding.

However, while most value discussions assume static values, social values are actually dynamic and evolving. Exploring how value-aligned AI systems can dynamically adapt to changing environmental values is crucial. Future technologies need to address static value alignment first, including strategies for sampling human groups for alignment. \citet{bakker2022fine} founds that consensus statements built silently from a subgroup will lead to dissent among excluded members, highlighting the consensus's sensitivity to individual input. For international cooperation, establishing a shared data center is necessary but also requires first determining which civilizations to include and if their values can align.

\subsubsection{Alignment Techniques for AI Governance}

It's crucial to ensure the reliability and trustworthiness of AI systems as they are adopted in various real-world decision-making scenarios. On one hand, language models still exhibit illusions during use, and on the other hand, the reliability of systems comprises two parts: the system's reliability under individual testing environments and its reliability in human interactions. Another issue is constructing systems with decision-making processes that are observable and explainable to users. From a social perspective, the proliferation of AI systems across fields also poses potential risks. This risk arises from a gap between AI developers, who often focus on advancing technology without considering its downstream applications, and AI adopters, who may transfer AI systems to their fields without adequate safety considerations or verification of replicable success \footnote{\url{https://www.scai.gov.sg/scai-question-11/}}.
Therefore, it is crucial to build a framework that enables AI adopters to accurately assess model utility and appropriateness, and allows AI regulators to quickly identify risks and issue safety alerts in AI systems.

Alignment techniques can facilitate synchronized, independent, and rigorous evaluations of AI systems. AI developers should prioritize appropriate bias handling during the training process, acknowledging the importance of socio-economic, cultural, and other differences. Furthermore, we should aim to develop robust and fair evaluation methods and datasets for auditing AI systems. \citet{zhu2023dyval} proposes the first dynamic testing protocol for large language models, utilizing Directed Acyclic Graphs (DAGs) to dynamically generate test data, thereby reducing the risk of test data memorization and contamination. Additionally, new robust security protocol evaluation methods have been introduced: \citet{shlegeris2023meta} suggests constructing adversarial policies to manage dangerously powerful and deceptive models, while \citet{greenblatt2023ai}  proposes (un)trusted editing to supervise models based on their harm and deceitfulness levels. Future efforts should also prevent AI systems from reward-hacking evaluation system exploits and aim to provide AI regulators with an explainable, independent, and centralized evaluation system.

AI adopters and the industry should allocate financial and computational resources to thoroughly evaluate use cases and share case studies showcasing both successes and failures. Equally important is training for adopters on downstream applications.

\section{Conclusion}\label{sec:conclusion}
In this survey, we have provided a broadly-scoped introduction to AI alignment, which aims to build AI systems that behave in line with human intentions and values. We specify the objectives of alignment as Robustness, Interpretability, Controllability, and Ethicality (\textbf{RICE}), and characterize the scope of alignment methods as comprising of \emph{forward alignment} (making AI systems aligned via alignment training) and \emph{backward alignment} (gaining evidence of the systems' alignment and govern them appropriately to avoid exacerbating misalignment risks). Currently, the two notable areas of research within forward alignment are \emph{learning from feedback} and \emph{learning under distribution shift}, while backward alignment is comprised of \emph{assurance} and \emph{governance}.

One thing that sets alignment apart from many other fields is its diversity \citep{hendrycks2022pragmatic} -- it is a tight assembly of multiple research directions and methods, tied together by a shared goal, as opposed to a shared methodology.
This diversity brings benefits. It fosters innovation by having the different directions compete and clash against each other, leading to a cross-pollination of ideas. It also allows different research directions to complement each other and together serve the goal of alignment; this is reflected in the \textit{alignment cycle} (see Figure \ref{fig:maindiagram}), where the four pillars are integrated into a self-improving loop that continually improves the alignment of AI systems.
Meanwhile, this diversity of research directions raises the barrier to entry into this field, which mandates the compilation of well-organized survey materials that serve both the newcomers and the experienced. In this survey, we attempt to address this need by providing a comprehensive and up-to-date overview of alignment. 

We attempt to account for the full diversity within the field by adopting a broad and inclusive characterization of alignment. Our survey of alignment gives a spotlight to almost all major research agendas in this field, as well as to real-world practices on the assurance and governance front. We recognize that boundaries of alignment are often vague and subject to debate. Therefore, when proposing the \text{RICE} principles, we put forth our broad characterization of alignment as an explicit choice.
In the meantime, we recognize that such a survey needs to be a long-term endeavor that is continually reviewed and updated. Both the problems and methods of alignment closely follow the development of machine learning. This fast-paced development means that new materials and frameworks can become outdated after merely a few years. This fact is one reason why we write the survey to reflect the latest developments, and also mandates continual maintenance and updates.

\subsection{Key Challenges in the Alignment Cycle}
Specifically, we outline key challenges and potential future directions based on the alignment cycle, namely forward and backward alignment.

\paragraph{Learning Human Intent from Rich Modalities (\textcolor{myred}{forward alignment})}Underspecificity of true human intent,\textit{i.e.}, the non-uniqueness of inferred human intent from binary feedback data, is a key challenge in scalable oversight.
Consider an AI system tasked with providing proof or refutation to a mathematical hypothesis, under a human evaluator who might be tricked by sophisticated false proofs. Our goal is to construct a training process that induces the AI system to output sound proofs as opposed to false proofs that seem convincing.
This system may mislead evaluators with plausible but false proofs due to the system's optimization for human approval, as it attempts to satisfy the superficial criteria of convincing proofs rather than focusing on accuracy. The fundamental problem stems from the reliance on binary feedback which categorizes responses simply as preferred or dispreferred, thus limiting the amount of information on true human preferences that's available to the learning algorithm, potentially leading to the preference of credible-seeming deceptive proofs over genuinely sound arguments.

To enhance the model's alignment with true human intent, researchers have proposed incorporating richer human input beyond binary choices, such as detailed text feedback \citep{chen2024learning} and real-time interations \citep{hadfield2016cooperative}. It allows the model to differentiate between proofs that are merely convincing and those that are truly sound, using nuanced human evaluations and a vast database of human-written texts. The broader input base helps in constructing a more accurate model of human preferences, reducing the risk of favoring misleading proofs while respecting the complexity of human intent and reasoning. Looking forward, even richer modalities like embodied societal interactions could represent an enticing next step.

It is worth noting that current LLMs are already trained on Internet-scale human text (and for multimodal models, also visual/audio content). Why, then, don't reward modeling algorithms already possess the ability to accurately pin down human intent? The explanation is that pretraining data does not feed into the reward modeling process in a way that biases the process towards true human intent, even though the reward model is finetuned from the pretrained model. For instance, neural circuits representing human intent can potentially be rewired during RLHF to perform manipulative behaviors. From another perspective, pretraining on text such as \emph{humans do not want to be tricked into believing things} does not induce the reward model to interpret later human feedback signals in this light, partly due to the lack of out-of-context learning capabilities in current LLMs \cite{berglund2023taken}. Solving these problems may enable reward modeling algorithms to learn human intent from massive pretraining data, a big step towards our goal.

We summarize three key questions for the learning of human intent from rich modalities. They serve as key dimensions for characterizing an alignment method from the intent modality lens, and almost all existing alignment methods can be categorized by their answers to these three questions.

\begin{enumerate}[left=0.3cm]
    \item \textbf{Learning algorithm}. As previously mentioned, we need to learn human intent from rich modalities in a way that guides the reward model's subsequent interpretation of human input.
    \item \textbf{Priors and inductive biases}. Human-like priors/inductive bias is needed for the reward modeling process to select the correct hypothesis of human intent, though this requirement is greatly loosened as the allowed modalities of human input expand.
    \item \textbf{Learner alignment}. We utilize the intent learner to align AI systems, possibly by using it as a reward model. However, this would not be possible if the intent learner, which is itself an AI system with potentially strong capabilities, is misaligned. This necessitates measures to avoid or contain the misalignment of the intent learner.
\end{enumerate}

\paragraph{Trustworthy Tools for Assurance (\textcolor{myblue}{backward alignment})}

% Detect Backdoor Behaviors
A major concern in AI alignment is deceptive alignment, where AI systems pursue aligned goals under most circumstances but may pursue other goals when opportunities arise. Recent studies have revealed that general alignment techniques (\textit{e.g.}, SFT, RLHF, Adversarial Training) fail to eradicate certain deceptive and backdoor behaviors, possibly leading to a misleading sense of safety \citep{hubinger2024sleeper}. With AI systems gaining power and access to more resources, hidden intentions that pose existential risks could have unimaginable consequences. \textit{How can we detect and eliminate deceptive and backdoor behaviors?} 

Reliable tools are still lacking to address this issue. On one hand, mechanistic interpretability tools encounter additional challenges due to the polysemanticity of neurons and scalability issues. On the other hand, there is a limited understanding of how jailbreaking functions and the susceptibility of language models to poisoning and backdoors \citep{anwar2024foundational}.

Additionally, given the potential misuse of AI systems in cyber attacks, biological warfare, and misinformation, it is crucial to develop reliable mechanisms to trace the origins of LLM outputs. While AI systems are becoming more integrated into society, societal readiness lags behind. This is evident from the inadequate AI governance efforts, insufficient public knowledge, governments' lack of necessary scientific and technical capabilities, the absence of institutions that can keep pace with LLM advancements, and the challenges in mitigating the social impacts of widespread harmful behaviors. Therefore, it is essential to reconsider AI alignment from a socio-technical standpoint, establish dependable AI assurance and governance mechanisms, and engage in effective international governance collaboration.

\paragraph{Value Elicitation and Value Implementation (\textcolor{myblue}{backward alignment})}

Current algorithms for learning from human feedback, particularly RLHF, often assume feedback comes from a singular, monolithic human source. However, this assumption is unrealistic due to widespread disagreements on contentious issues globally, which frequently result in conflicting judgments about AI system outputs \citep{santurkar2023whose}. Consequently, determining who to draw feedback from and understanding the scope and nature of human values infused into models are crucial questions for the field of alignment.

\textit{Value Elicitation} and \textit{Value Implementation} aim to define the values and norms that AI systems should encode and how to integrate these into AI systems. Human values and preferences are diverse, ranging from strict rules like laws and moral principles to social etiquette and specific domain preferences \citep{cahyawijaya2024high,kirk2024prism}. We need reliable tools to reveal the values embedded in current AI systems and potential social risks, enabling us to mitigate these risks more effectively \footnote{\url{https://www.scai.gov.sg}}. 

\emph{Democratic human input} is one of the leading solutions to value elicitation and implementation. This method gathers input from a large, demographically representative sample of individuals, aggregating preferences and values into a coherent policy, rather than relying on feedback from a single individual. This approach is heavily influenced by the computational social choice literature \cite{brandt2016handbook}. Leading industry \cite{zaremb2023democratic} and academic \cite{kopf2024openassistant} labs have adopted democratic human input for LLMs. However, research is still needed on its integration into more agentic AI systems, such as LLM-based autonomous agents.

Despite its apparent simplicity, democratic human input encounters significant practical and fundamental challenges. Obtaining a truly random sample of the global population is particularly challenging, as 33\% of people worldwide do not have Internet access and thus are excluded from participating in AI system training \cite{itu2023stats}. Furthermore, human feedback becomes less effective when the AI system's reasoning capabilities surpass those of humans, making it difficult for human workers to evaluate its outputs. To complement democratic human input, alternative approaches aim to formalize universally recognized meta-level moral principles, such as \emph{moral consistency}, \emph{moral reflection}, and \emph{moral progress}, designing algorithms to enact these principles. Although these methods still rely on human data and input, they do not demand strict representativeness and are less constrained by human oversight limitations.

\begin{itemize}
    \item \textbf{Moral consistency}. There is a general consensus that moral principles should be consistently applied, meaning similar cases should receive similar treatment irrespective of the people or parties involved. Algorithms have been developed to integrate this principle into the ethical decision-making processes of AI systems \cite{jin2022when}.
    \item \textbf{Moral reflection and moral progress}. The \emph{coherent extrapolated volition} concept was developed to formalize the role of reflection in shaping human moral values \cite{sovik2022overarching}. Inspired by this, subsequent algorithms were designed to enable AI systems to mimic human moral reflection, thereby influencing their actions \cite{xie2023defending}. Furthermore, the logical next step of moral reflection is \emph{moral progress}, demonstrated by AI-driven analyses of historical moral trends \cite{schramowski2020moral,atif2022evolution} and efforts to permanently integrate continual moral advancement into AI systems \cite{kenward2021machine}.
\end{itemize}

\subsection{Key Traits and Future Directions in Alignment Research}

In the end of the survey, we conclude the survey by looking ahead and presenting the key traits in this field that we believe ought to be preserved or fostered.

\paragraph{Open-Ended Exploration of Novel Challenges and Approaches} A lot of the alignment discourse is built upon classic works that predate the recent developments of LLMs and other breakthroughs in large-scale deep learning. Thus, when this paradigm shift happens in the machine learning field, it is plausible that some challenges in alignment become less salient while others become more so; after all, one defining feature of scientific theories is their falsifiability \citep{popper2005logic}. More importantly, this shift in machine learning methodology and the broader trend of ever-tighter integration of AI systems into society \citep{abbass2019social} introduces novel challenges that could not be envisioned before. This requires that we engage in \textit{open-ended exploration}, actively seeking out new challenges that were previously neglected. Moreover, such an exploration need not be constrained to challenges -- a similar mindset should be adopted regarding approaches and solutions, thus building a more diverse portfolio for both the \textit{questions} and the \textit{answers} \citep{shimi2022diversify}.

\paragraph{Combining Forward-Looking and Present-Oriented Perspectives} Alignment has emphasized harms from potential advanced AI systems that possess stronger capabilities than current systems \citep{firstprinc}. These systems might come into existence well into the future, or might just be a few years away \citep{stein2022expert}. The former possibility requires us to look into extrapolated trends and hypothetical scenarios \citep{carlsmith2022power}. In contrast, the latter possibility highlights the need for on-the-ground efforts that work with current governance institutions and use current systems as a prototype for more advanced ones \citep{cotra2021the}. 

\paragraph{Emphasis on Policy Relevance} Alignment research does not live in a vacuum but in an ecosystem\footnote{See \url{aisafety.world} for a map of the organizational landscape of alignment.}, with participation from researchers, industry actors, governments, and non-governmental organizations. Research serving the needs of the AI alignment and safety ecosystem would therefore be useful. Such needs include solving the key barriers to various governance schemes, for example, extreme risk evaluations \citep{shevlane2023model}, infrastructure for computing governance, and mechanisms for making verifiable claims about AI systems \citep{brundage2020toward}. % and other governance-oriented research directions presented in \S\ref{sec:ai-tech-for-ai-reg}.

\paragraph{Emphasis on Social Complexities and Moral Values} As AI systems become increasingly integrated into society \citep{abbass2019social}, alignment ceases to be only a single-agent problem and becomes a social problem. Here, the meaning of \textit{social} is three-fold.

\begin{enumerate}[left=0.3cm]
    \item Alignment research in multi-agent settings featuring the interactions between multiple AI systems and multiple humans \citep{critch2020ai,liu2024training}. This includes how AI systems can receive granular feedback from realistic simulated societies, ensuring consistency in training scenarios and among multiple entities (\textit{i.e.}, multi-agent, multiple AI systems, and multiple humans) , which not only aids in generalizing AI systems in multi-entity settings but also helps avoid problems associated with RL \citep{liu2024training}.
    \item Incorporating human moral and social values into alignment (see \S\ref{sec:values-in-intro} and \S\ref{sec:human-values-alignment}), which is closely linked to the field of \textit{machine ethics} and \textit{value alignment} \citep{gabriel2020artificial,gabriel2021challenge}. 
    \item Modeling and predicting the impacts of AI systems on society, which requires methods to approach the complexities of the social system, including those from the social sciences. Examples of potentially useful methodologies include social simulation \citep{bonabeau2002agent,de2014agent,park2023generative} and game theory \citep{du2023improving}.
\end{enumerate}

% \addcontentsline{toc}{section}{Acknowledgments}
\section*{Acknowledgments}
We thank David Krueger, Anca Dragan, Alan Chan, Stephen Casper, Haoxing Du, Lawrence Chan, Johannes Treutlein, and YingShan Lei for their helpful and constructive feedback on the manuscript. We thank Yi Qu for the graphical design and refinement of the figures in our survey. 

\clearpage
% \addcontentsline{toc}{section}{References}
\bibliography{new_add}
\bibliographystyle{styles/acl_natbib}

\end{document}